\documentclass{article} 
\usepackage{iclr2024_conference,times}

\usepackage[utf8]{inputenc} 
\usepackage[T1]{fontenc}    
\usepackage{hyperref}       
\usepackage{url}            
\usepackage{booktabs}       
\usepackage{amsfonts}       
\usepackage{nicefrac}       
\usepackage{microtype}      
\usepackage{xcolor}         

\usepackage[pdftex]{graphicx}
\usepackage{caption}
\usepackage{subcaption}

\usepackage{amsmath,amsfonts,bm}

\def\eqref#1{equation~\ref{#1}}

\def\1{\bm{1}}

\DeclareMathAlphabet{\mathsfit}{\encodingdefault}{\sfdefault}{m}{sl}
\SetMathAlphabet{\mathsfit}{bold}{\encodingdefault}{\sfdefault}{bx}{n}

\def\gD{{\mathcal{D}}}

\usepackage{amsthm}
\usepackage{amssymb}
\usepackage{url}
\usepackage{subcaption}
\usepackage{graphics}
\usepackage{graphicx}
\graphicspath{ {./figures/} }
\usepackage{graphicx,wrapfig,lipsum}
\usepackage{multirow}
\usepackage{boldline}
\usepackage{comment}

\usepackage{comment}
\usepackage{multirow}
\usepackage{makecell}

\usepackage{algpseudocode}
\usepackage{algorithm}
\usepackage{algorithmicx}
\usepackage{adjustbox}

\usepackage{thmtools}
\usepackage{thm-restate}
\newtheorem{theorem}{Theorem}

\theoremstyle{definition}
\newtheorem{definition}{Definition}

\theoremstyle{remark}

\usepackage[capitalize]{cleveref}

\title{Efficient Episodic Memory Utilization of Cooperative Multi-Agent Reinforcement Learning}

\author{Hyungho Na\textsuperscript{\textmd{1}}, Yunkyeong Seo\textsuperscript{\textmd{1}} \& Il-Chul Moon\textsuperscript{\textmd{1,2}}\\
	\textsuperscript{\textmd{1}}Korea Advanced Institute of Science and Technology (KAIST), \textsuperscript{\textmd{2}}summary.ai\\	 
	\texttt{\{gudgh723\}@gmail.com,\{tjdbsrud,icmoon\}@kaist.ac.kr} \\
}

\iclrfinalcopy 
\begin{document}
	
	\maketitle
	
	\begin{abstract}
		In cooperative multi-agent reinforcement learning (MARL), agents aim to achieve a common goal, such as defeating enemies or scoring a goal. Existing MARL algorithms are effective but still require significant learning time and often get trapped in local optima by complex tasks, subsequently failing to discover a goal-reaching policy. To address this, we introduce Efficient episodic Memory Utilization (EMU) for MARL, with two primary objectives: (a) accelerating reinforcement learning by leveraging semantically coherent memory from an episodic buffer and (b) selectively promoting desirable transitions to prevent local convergence. To achieve (a), EMU incorporates a trainable encoder/decoder structure alongside MARL, creating coherent memory embeddings that facilitate exploratory memory recall. To achieve (b), EMU introduces a novel reward structure called episodic incentive based on the desirability of states. This reward improves the TD target in Q-learning and acts as an additional incentive for desirable transitions. We provide theoretical support for the proposed incentive and demonstrate the effectiveness of EMU compared to conventional episodic control. The proposed method is evaluated in StarCraft II and Google Research Football, and empirical results indicate further performance improvement over state-of-the-art methods. Our code is available at: \texttt{https://github.com/HyunghoNa/EMU}. 
	\end{abstract}
	
	\section{Introduction}
	\label{introduction}
	
	Recently, cooperative MARL has been adopted to many applications, including traffic control~\citep{wiering2000multi}, resource allocation~\citep{dandanov2017dynamic}, robot path planning~\citep{wang2020mobile}, and production systems ~\citep{dittrich2020cooperative}, etc. In spite of these successful applications, cooperative MARL still faces challenges in learning proper coordination among multiple agents because of the partial observability and the interaction between agents during training.
	
	To address these challenges, the framework of centralized training and decentralized execution (CTDE) \citep{oliehoek2008optimal,oliehoek2016concise,gupta2017cooperative} has been proposed. CTDE enables a decentralized execution while fully utilizing global information during centralized training, so CTDE improves policy learning by accessing to global states at the training phase. 
	Especially, value factorization approaches \citep{sunehag2017value,rashid2018qmix,son2019qtran,yang2020qatten,rashid2020weighted,wang2020qplex} maintain the consistency between individual and joint action selection, achieving the state-of-the-art performance on difficult multi-agent tasks, such as StarCraft II Multi-agent Challenge (SMAC)~\citep{samvelyan2019starcraft}.
	However, learning optimal policy in MARL still requires a long convergence time due to the interaction between agents, and the trained models often fall into local optima, particularly when agents perform complex tasks \citep{mahajan2019maven}. 
	Hence, researchers present a committed exploration mechanism under this CTDE training practice~\citep{mahajan2019maven,yang2019hierarchical,wang2019influence,liu2021cooperative} with the expectation to find episodes escaping from the local optima.
	
	Despite the required exploration in MARL with CTDE, recent works on episodic control emphasize the exploitation of episodic memory to expedite reinforcement learning. Episodic control \citep{lengyel2007hippocampal,blundell2016model,lin2018episodic,pritzel2017neural} memorizes explored states and their best returns from experience in the episodic memory, to converge on the best policy. Recently, this episodic control has been adopted to MARL \citep{zheng2021episodic}, and this episodic control case shows faster convergence than the learning without such memory. Whereas there are merits from episodic memory and control from its utilization, there exists a problem of determining which memories to recall and how to use them, to efficiently explore from the memory. According to \cite{blundell2016model,lin2018episodic,zheng2021episodic}, the previous episodic control generally utilizes a random projection to embed global states, but this random projection hardly makes the semantically similar states close to one another in the embedding space. In this case, exploration will be limited to a narrow distance threshold. However, this small threshold leads to inefficient memory utilization because the recall of episodic memory under such small thresholds retrieves only the same state without consideration of semantic similarity from the perspective of goal achievement. Additionally, the naive utilization of episodic control on complex tasks involves the risk of converging to local optima by repeatedly revisiting previously explored states, favoring exploitation over exploration. 

\textbf{Contribution.} This paper presents an \textbf{E}fficient episodic \textbf{M}emory \textbf{U}tilization for multi-agent reinforcement learning (EMU), a framework to selectively encourage desirable transitions with semantic memory embeddings.

\begin{itemize}
	\vspace{-0.05in}
	\item \textbf{Efficient memory embedding:} 
	When generating features of a global state for episodic memory (Figure \ref{fig:overview}(b)), we adopt an encoder/decoder structure where 1) an encoder embeds a global state conditioned on timestep into a low-dimensional feature and 2) a decoder takes this feature as an input conditioned on the timestep to predict the return of the global state. In addition, to ensure smoother embedding space, we also consider the reconstruction of the global state when training the decoder to predict its return. To this end, we develop \textbf{d}eterministic \textbf{C}onditional \textbf{A}uto\textbf{E}ncoder (dCAE) (Figure \ref{fig:overview}(c)). With this structure, important features for overall return can be captured in the embedding space. The proposed embedding contains semantic meaning and thus guarantees a gradual change of feature space, making the further exploration on memory space near the given state, i.e., efficient memory utilization.
	
	\item \textbf{Episodic incentive generation:} 
	While the semantic embedding provides a space to explore, we still need to identify promising state transitions to explore. Therefore, we define a {\bf\textit{desirable trajectory}} representing the highest return path, such as destroying all enemies in SMAC or scoring a goal in Google Research Football (GRF) \citep{kurach2020google}. States on this trajectory are marked as desirable in episodic memory, so we could incentivize the exploration on such states according to their desirability. We name this incentive structure as an episodic incentive (Figure \ref{fig:overview}(d)), encouraging desirable transitions and preventing convergence to unsatisfactory local optima. We provide theoretical analyses demonstrating that this episodic incentive yields a better gradient signal compared to conventional episodic control.
\end{itemize}
\vspace{-0.05in}
We evaluate EMU on SMAC and GRF, and empirical results demonstrate that the proposed method achieves further performance improvement compared to the state-of-art baseline methods. Ablation studies and qualitative analyses validate the propositions made by this paper.

\section{Preliminary}
\label{preliminary}	
\vspace{-0.05in}
\subsection{Decentralized POMDP}
\label{Decentralized POMDP}	
A fully cooperative multi-agent task can be formalized by following the Decentralized Partially Observable Markov Decision Process (Dec-POMDP) \citep{oliehoek2016concise}, $G = \left\langle {I,S,A,P,R,\Omega,O,n,\gamma} \right\rangle $, where $I$ is the finite set of $n$ agents; $s \in S$ is the true state of the environment; ${a_i} \in A$ is the $i$-th agent's action forming the joint action $\bm{a} \in {A^n}$; $P(s'|s,\bm{a})$ is the state transition function; $R$ is a reward function $r = R(s,\bm{a},s') \in \mathbb{R}$; $\Omega $ is the observation space; $O$ is the observation function generating an observation for each agent $o_i \in \Omega$; and finally, $\gamma \in [0,1)$ is a discount factor. At each timestep, an agent has its own local observation $o_i$, and the agent selects an action ${a_i} \in A$. The current state $s$ and the joint action of all agents $\bm{a}$ lead to a next state $s'$ according to $P(s'|s,\bm{a})$. The joint variable of $s$, $\bm{a}$, and $s'$ will determine the identical reward $r$ across the multi-agent group. In addition, similar to \cite{hausknecht2015deep,rashid2018qmix}, each agent utilizes a local action-observation history ${\tau_i} \in T \equiv (\Omega \times A)$ for its policy ${\pi_i}(a|{\tau_i})$, where ${\pi}:T \times {A} \to [0,1]$. 

\subsection{Desirability and Desirable Trajectory}
\label{Desirability and Desirable Trajectory}	
\begin{definition}
	\label{def: desirable trajectory}
	(Desirability and Desirable Trajectory) For a given threshold return $R_{thr}$ and a trajectory $\mathcal{T}:=\left\{s_0,\bm{a_0},r_0,s_1,\bm{a_1},r_1,...,s_T\right\}$, $\mathcal{T}$ is considered as a desirable trajectory, denoted as $\mathcal{T}_{\xi}$, when an episodic return is $R_{t=0}=\Sigma_{t'=t}^{T-1} r_{t'} \geq R_{\textrm{thr}}$. A binary indicator $\xi(\cdot)$ denotes the desirability of state $s_t$ as $\xi(s_t)=1$ when $s_t \in \forall \mathcal{T}_{\xi}$.
\end{definition}

In cooperative MARL tasks, such as SMAC and GRF, the total amount of rewards from the environment within an episode is often limited as $R_{\textrm{max}}$, which is only given when cooperative agents achieve a common goal. In such a case, we can set $R_{thr}=R_{\textrm{max}}$. For further description of cooperative MARL, please see Appendix \ref{related_works}.

\subsection{Episodic control in MARL}\label{Episodic_control}
Episodic control was introduced from the analogy of a brain's hippocampus for memory utilization \citep{lengyel2007hippocampal}. After the introduction of deep Q-network, \citet{blundell2016model} adopted this idea of episodic control to the model-free setting by storing the highest return of a given state, to efficiently estimate the Q-values of the state. This recalling of the high-reward experiences helps to increase sample efficiency and thus expedites the overall learning process \citep{blundell2016model, pritzel2017neural, lin2018episodic}. Please see Appendix \ref{related_works} for related works and further discussions.

At timestep $t$, let us define a global state as $s_t$. When utilizing episodic control, instead of directly using $s_t$, researchers adopt a state embedding function ${f_\phi(s): S \rightarrow \mathbb{R}^k}$ to project states toward a $k$-dimensional vector space. With this projection, a representation of global state ${s_t}$ becomes $x_t = {f_\phi }({s_t})$. The episodic control memorizes $H(f_\phi(s_t))$, i.e., the highest return of a given global state $s_t$, in episodic buffer $\gD_E$~\citep{pritzel2017neural,lin2018episodic,zheng2021episodic}. Here, ${x_t}$ is used as a key to the highest return, $H({x_t})$; as a key-value pair in $\gD_E$. The episodic control in \citet{lin2018episodic} updates $H({x_t})$ with the following rules.
\begin{equation}\label{Eq:Episodic_update}
	\begin{aligned}
		H({x_t}) = \left\{ {\begin{array}{*{20}{c}}		
				\max \{ H({{\hat x}_t}),{R_t}({s_t},\bm{a_t})\} ,{\rm{ }} &\textrm{if} \: ||{{\hat x}_t} - {x_t}||_2 < \delta \\
				{\rm{              }}{R_t}({s_t},\bm{a_t}),\; &{\textrm{otherwise    }}, \\
		\end{array}} \right.
	\end{aligned}
\end{equation}
where ${R_t}({s_t},\bm{a_t})$ is the return of a given $({s_t},{\bm{a_t}})$; $\delta$ is a threshold value of state-embedding difference; and $\hat x_t=f_{\phi}({\hat s_t})$ is $x_t=f_{\phi}({s_t})$'s nearest neighbor in $\gD_E$. If there is no similar projected state ${\hat x_t}$ such that $ ||{{\hat x}_t} - {x_t}||_2 < \delta$ in the memory, then $H({x_t})$ keeps the current ${R_t}({s_t},\bm{a_t})$. Leveraging the episodic memory, EMC \citep{zheng2021episodic} presents the one-step TD memory target $Q_{EC}(f_{\phi}(s_t),\bm{a_t})$ as
\begin{equation}\label{Eq:Episodic_target}
	\begin{aligned}
		Q_{EC}(f_{\phi}(s_t),\bm{a_t}) = r_t(s_t,\bm{a_t})+\gamma H(f_{\phi}(s_{t+1})).
	\end{aligned}
\end{equation}
Then, the loss function $L_\theta ^{EC} $ for training can be expressed as the weighted sum of one-step TD error and one-step TD memory error, i.e., Monte Carlo (MC) inference error, based on $Q_{EC}(f_{\phi}(s_t),\bm{a_t})$.
\begin{equation}\label{Eq:Episodic_loss}
	L_\theta ^{EC} = {( y(s,\bm{a}) - {Q_{tot}}(s,\bm{a};\theta))^2} + \lambda {({Q_{EC}}(f_{\phi}(s),\bm{a}) - {Q_{tot} }(s,\bm{a};\theta))^2},
\end{equation}
where $y(s,\bm{a})$ is one-step TD target; $Q_{tot}$ is the joint Q-value function parameterized by $\theta$; and $\lambda$ is a scale factor.	

\paragraph{Problem of the conventional episodic control with random projection}
\vspace{-0.1in}
Random projection is useful for dimensionality reduction as it preserves distance relationships, as demonstrated by the Johnson-Lindenstrauss lemma \citep{dasgupta2003elementary}. However, a random projection adopted for $f_\phi(s)$ hardly has a semantic meaning in its embedding $x_t$, as it puts random weights on the state features without considering the patterns of determining the state returns. Additionally, when recalling the memory from $\gD_E$, the projected state $x_t$ can abruptly change even with a small change of $s_t$ because the embedding is not being regulated by the return. This results in a sparse selection of semantically similar memories, i.e. similar states with similar or better rewards. As a result, conventional episodic control using random projection only recalls identical states and relies on its own Monte-Carlo (MC) return to regulate the one-step TD target inference, limiting exploration of nearby states on the embedding space.    

The problem intensifies when the high-return states in the early training phase are indeed local optima. In such cases, the naive utilization of episodic control is prone to converge on local minima. As a result, for the super hard tasks of SMAC, EMC~\citep{zheng2021episodic} had to decrease the magnitude of this regularization to almost zero, i.e., not considering episodic memories anymore. 

\section{Methodology}
\label{Methodology}
\begin{figure*}[!htb]
	\begin{center}
		\includegraphics[width=\linewidth]{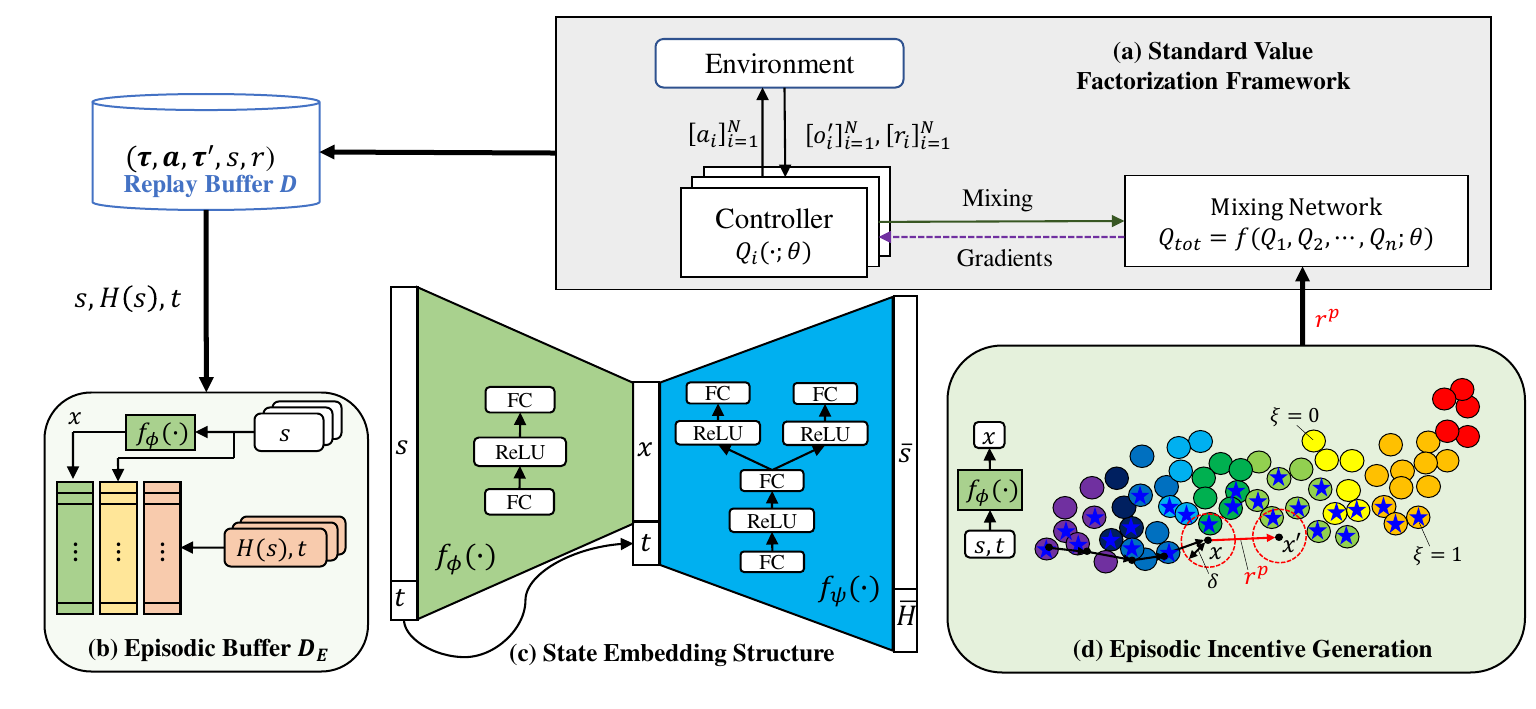}
		\vskip -0.1in
		\caption{Overview of EMU framework.}
		\label{fig:overview}
	\end{center}
	\vskip -0.25in
\end{figure*}

This section introduces \textbf{E}fficient episodic \textbf{M}emory \textbf{U}tilization (EMU) (Figure \ref{fig:overview}). 
We begin by explaining how to construct \textbf{(1) semantic memory embeddings} to better utilize the episodic memory, which enables memory recall of similar, more promising states. To further improve memory utilization, as an alternative to the conventional episodic control, we propose \textbf{(2) episodic incentive} that selectively encourages desirable transitions while preventing local convergence towards undesirable trajectories. 

\subsection{Semantic Memory Embedding}
\label{Efficient_Memory_embedding}
\vskip -0.05in
\textbf{Episodic Memory Construction} To address the problems of a random projection adopted in episodic control, we propose a trainable embedding function $f_\phi(s)$ to learn the state embedding patterns affected by the highest return. The problem of a learnable embedding network $f_{\phi}$ is that the match between $H(f_{\phi}(s_t))$ and $s_t$ breaks whenever $f_{\phi}$ is updated. Hence, we save the global state $s_t$ as well as a pair of $H_t$ and $x_t$ in $\gD_E$, so that we can update $x=f_{\phi}(s)$ whenever $f_{\phi}$ is updated. In addition, we store the desirability $\xi$ of $s_t$ according to Definition \ref{def: desirable trajectory}. Appendix \ref{memory_contruction} illustrates the details of memory construction proposed by this paper.

\textbf{Learning framework for State Embedding}
When training $f_{\phi}(s_t)$, it is critical to extract important features of a global state that affect its value, i.e., the highest return. Thus, we additionally adopt a decoder structure $\bar{H_t}=f_{\psi}(x_t)$ to predict the highest return $H_t$ of $s_t$. We call this embedding function as {\bf EmbNet}, and its learning objective of $f_{\phi}$ and $f_{\psi}$ can be written as
\begin{equation}\label{Eq:Loss_embedder}
	\mathcal{L}(\bm{\phi},\bm{\psi})=\left( H_t-f_{\psi}(f_{\phi}(s_t)) \right)^2.
\end{equation}  
When constructing the embedding space, we found that an additional consideration of reconstruction of state $s$ conditioned on timestep $t$ improves the quality of feature extraction and constitutes a smoother embedding space. To this end, we develop the deterministic conditional autoencoder ({\bf dCAE}), and the corresponding loss function can be expressed as
\begin{equation}\label{Eq:Loss_embedder_dCAE}
	\mathcal{L}(\bm{\phi},\bm{\psi})= \left( H_t-f_{\psi}^H(f_{\phi}(s_t|t)|t) \right)^2 + \lambda_{rcon} ||s_t-f_{\psi}^s(f_{\phi}(s_t|t)|t)||_2^2,
\end{equation}	
where $f_{\psi}^H$ predicts the highest return; $f_{\psi}^s$ reconstructs $s_t$; $\lambda_{rcon}$ is a scale factor. Here, $f_{\psi}^H$ and $f_{\psi}^s$ share the lower part of networks as illustrated in Figure \ref{fig:overview}(c). Appendix \ref{implementation_details} presents the details of network structure of $f_{\phi}$ and $f_{\psi}$, and Algorithm \ref{alg:embedder_training} in Appendix \ref{implementation_details} presents the learning framework for $f_{\phi}$ and $f_{\psi}$. This training is conducted periodically in parallel to the RL policy learning on $Q_{tot}(\cdot;\theta)$.

\begin{figure}[t]
	\begin{center}
		\begin{subfigure}{0.32\linewidth} 
			\includegraphics[width=\linewidth]{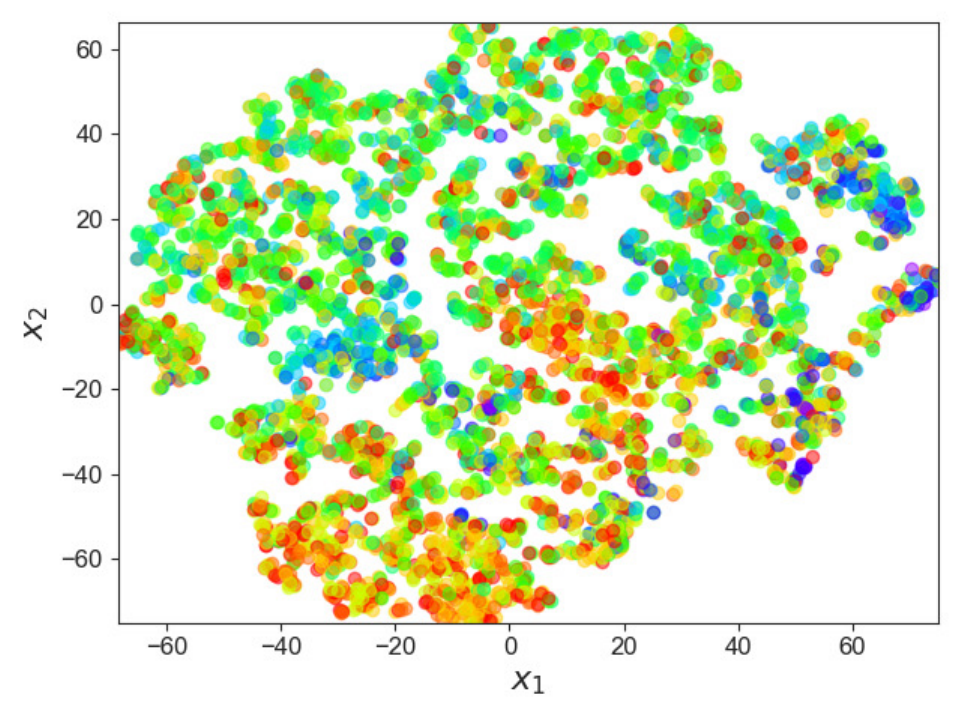}
			\caption{Random Projection}
		\end{subfigure}\hfill
		\begin{subfigure}{0.32\linewidth}
			\includegraphics[width=\linewidth]{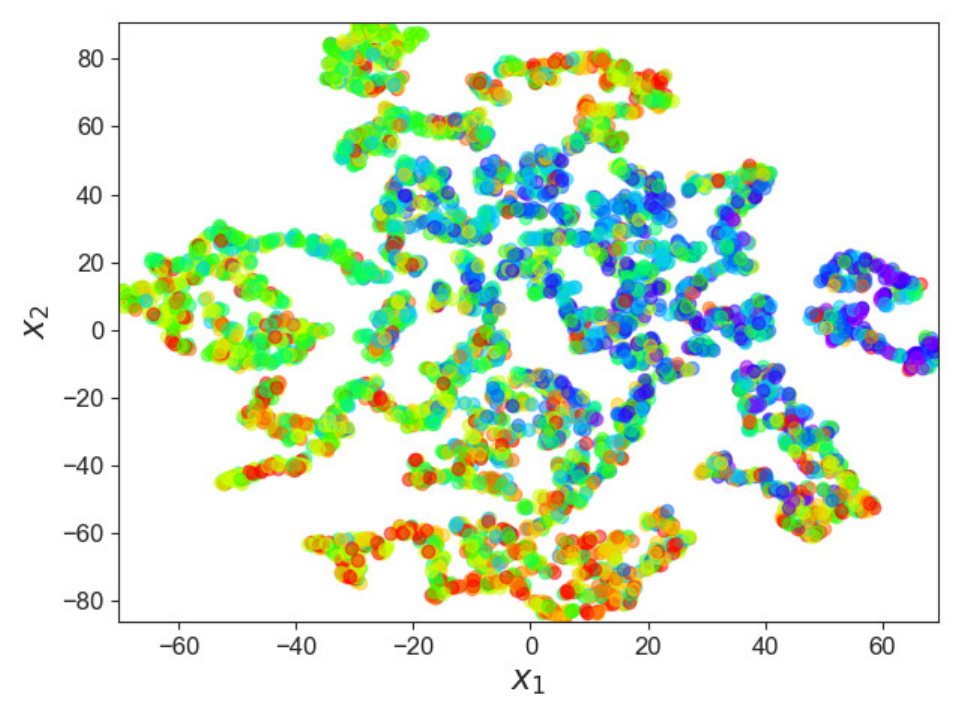}
			\caption{EmbNet}
		\end{subfigure}\hfill
		\begin{subfigure}{0.32\linewidth}
			\includegraphics[width=\linewidth]{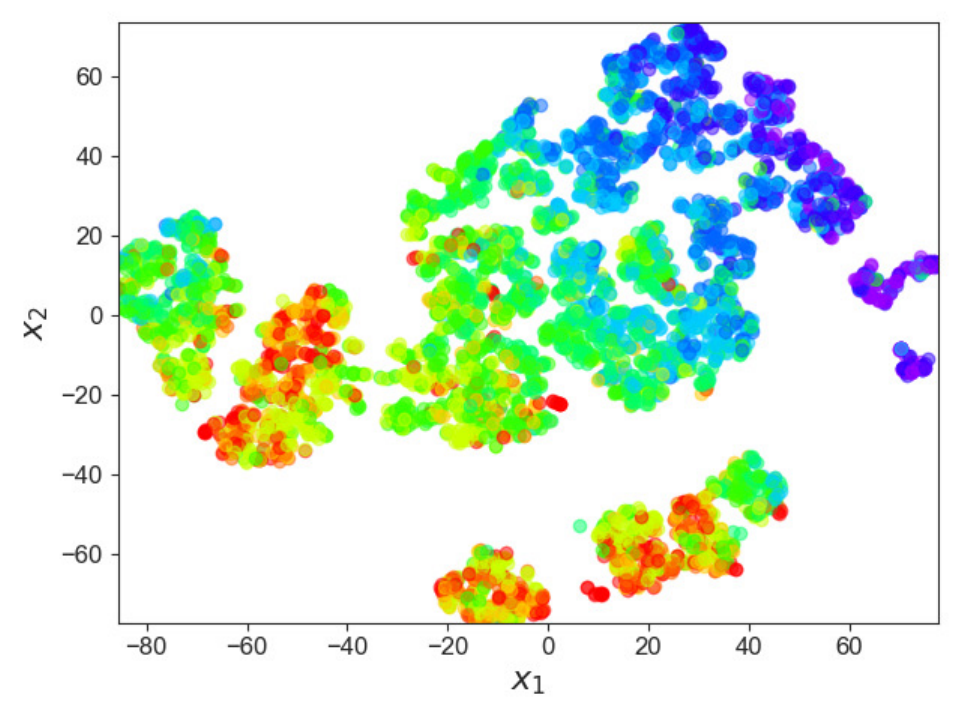}
			\caption{dCAE}
		\end{subfigure}\hfill
	\end{center}
	\vskip -0.10in	
	\caption{t-SNE of sampled embedding $x \in \gD_{E}$. Colors from red to purple (rainbow) represent from low return to high return.}
	\label{fig:t-SNE}
	\vskip -0.15in
\end{figure}

Figure \ref{fig:t-SNE} illustrates the result of t-SNE \citep{van2008visualizing} of 50K samples of $x \in \gD_E$ out of 1M memory data in training for \texttt{3s\_vs\_5z} task of SMAC. Unlike supervised learning with label data, there is no label for each $x_t$. Thus, we mark $x_t$ with its pair of the highest return $H_t$. Compared to a random projection in Figure \ref{fig:t-SNE}(a), $x_t$ via $f_\phi$ is well-clustered, according to the similarity of the embedded state and its return.
This clustering of $x_t$ enables us to safely select episodic memories around the key state $s_t$, which constitutes efficient memory utilization. This memory utilization expedites learning speed as well as encourages exploration to a more promising state $\hat{s}_t$ near $s_t$. 
Appendix \ref{memory_utilization} illustrates how to determine $\delta$ of Eq. \ref{Eq:Episodic_update} in a memory-efficient way.

\subsection{Episodic Incentive}
\label{Episodic_incentive}
\vskip -0.05in
With the learnable memory embedding for an efficient memory recall, how to use the selected memories still remains a challenge because a naive utilization of episodic memory is prone to converge on local minima.
To solve this issue, we propose a new reward structure called \textbf{episodic incentive} $r^p$ by leveraging the desirability $\xi$ of states in $\gD_{E}$. Before deriving the episodic incentive $r^p$, we first need to understand the characteristics of episodic control. In this section, we denote the joint Q-function $Q_{tot}(\cdot;\theta)$ simply as $Q_{\theta}$ for conciseness.

\begin{theorem}\label{thm1}
	Given a transition $(s,\bm{a},r,s')$ and $H(x')$, let $L_{\theta}$ be the Q-learning loss with additional transition reward, i.e., $L_{\theta}:= {(y(s,\bm{a})+{r^{EC}}(s,\bm{a},s') - {Q_{tot}}(s,\bm{a};\theta))^2}$ where ${r^{EC}}(s,\bm{a},s') := \lambda (r(s,\bm{a}) + \gamma H(x') - {Q_\theta }(s,\bm{a}))$, then ${\nabla _\theta }{L_\theta }={\nabla _\theta }L_\theta ^{EC}$. (Proof in Appendix \ref{proof_thm1}) 
\end{theorem}

As Theorem \ref{thm1} suggests, we can generate the same gradient signal as the episodic control by leveraging the additional transition reward ${r^{EC}}(s,\bm{a},s')$. However, ${r^{EC}}(s,\bm{a},s')$ accompanies a risk of local convergence as discussed in Section \ref{Episodic_control}. Therefore, instead of applying $r^{EC}(s,\bm{a},s')$, we propose the episodic incentive $r^p := \gamma \hat \eta (s')$ that provides an additional reward for the desirable transition $(s,\bm{a},r,s')$, such that $\xi (s') = 1$. Here, $\hat \eta (s')$  estimates $\eta^{*} (s')$, which represents the difference between the true value ${V^*}(s')$ of $s'$ and the predicted value via target network ${\max _{\bm{a}'}}{Q_{{\theta ^ {-} }}}(s',\bm{a}')$, defined as
\begin{equation}\label{Eq:eta}
	{\eta ^*}(s') := {V^*}(s') - {\max _{\bm{a}'}}{Q_{{\theta ^ - }}}(s',\bm{a}').
\end{equation}
Note that we do not know ${V^*}(s')$ and subsequently ${\eta ^*}(s')$. 
To accurately estimate $\eta^{*} (s')$ with ${\hat \eta}(s')$, we use the expected value considering the current policy ${\pi _\theta }$ as $\hat \eta(s')  := {\mathbb{E}_{{\pi _\theta }}}[\eta(s') ]$ where $\eta  \in [0,{\eta _{\max }(s')}]$ for $s' \sim P(s'|s,\bm{a} \sim \pi _\theta)$. Here, $\eta _{\max }(s')$ can be reasonably approximated by using $H(f_{\phi}(s'))$ in $\gD_{E}$. Then, with the count-based estimation $\hat\eta (s')$, episodic incentive $r^p$ can be expressed as 
\begin{equation}\label{Eq:exp_eta}
	\begin{aligned}
		r^p=\gamma\hat \eta(s') = \gamma{\mathbb{E}_{{\pi _\theta }}}[\eta(s') ] \simeq \gamma\frac{{{N_\xi }(s')}}{{{N_{call}}(s')}}{\eta _{\max }}(s')
		= \gamma\frac{{{N_\xi }(s')}}{{{N_{call}}(s')}}(H(f_{\phi}(s')) - {\max _{a'}}{Q_{{\theta ^ {-} }}}(s',a')),
	\end{aligned}
\end{equation}
where $N_{call}(s')$ is the number of visits on $\hat x' = \operatorname{NN}(f_{\phi}(s')) \in \gD_{E}$; and $N_{\xi}$ is the number of desirable transition from $\hat x'$. Here, $\operatorname{NN}(\cdot)$ represents a function for selecting the nearest neighbor.
From Theorem \ref{thm1}, the loss function adopting episodic control with an alternative transition reward $r^p$ instead of $r^{EC}$ can be expressed as
\begin{equation}\label{Eq:Loss_with_rp}
	{L_\theta ^p} = {(r(s,\bm{a}) + r^{p} + \gamma {\max _{\bm{a}'}}{Q_{{\theta ^ - }}}(s',\bm{a}') - {Q_\theta }(s,\bm{a}))^2}.
\end{equation}  
Then, the gradient signal of the one-step TD inference loss ${\nabla _\theta }L_\theta ^p$ with the episodic reward $r^p=\gamma \hat \eta(s')$ can be written as
\begin{equation}\label{Eq:grad_Lp}
	\begin{aligned}
		{\nabla _\theta }L_\theta ^p =  - 2{\nabla _\theta }{Q_\theta }(s,a)(\Delta {\varepsilon _{TD}} + {r^p})=- 2{\nabla _\theta }{Q_\theta }(s,a)(\Delta {\varepsilon _{TD}} + \gamma \frac{{{N_\xi }(s')}}{{{N_{call}}(s')}}{\eta _{\max }}(s')),
	\end{aligned}
\end{equation}
where $\Delta {\varepsilon _{TD}} = r(s,a) + \gamma {\max _{a'}}{Q_{{\theta ^ - }}}(s',a') - {Q_\theta }(s,a)$ is one-step inference TD error. Here, the gradient signal ${\nabla _\theta }L_\theta ^p$ with the proposed episodic reward $r^p$ can accurately estimate the optimal gradient signal as follows.
\begin{theorem}\label{thm2}
	Let ${\nabla _\theta }L_\theta ^* =  - 2{\nabla _\theta }{Q_\theta }(s,a)(\Delta \varepsilon _{TD}^*)$ be the optimal gradient signal with the true one step TD error $\Delta \varepsilon _{TD}^* = r(s,a) + \gamma {V^*}(s') - {Q_\theta }(s,a)$. Then, the gradient signal ${\nabla _\theta }L_\theta ^p$ with the episodic incentive $r^p$ converges to the optimal gradient signal as the policy converges to the optimal policy $\pi_{\theta}^*$, i.e., ${\nabla _\theta }L_\theta ^p \to {\nabla _\theta }L_\theta ^*$ as ${\pi _\theta } \to \pi _\theta ^*$. (Proof in Appendix \ref{proof_thm2}) 
\end{theorem}	
\begin{wrapfigure}{R}{0.61\linewidth} 
	\vspace{-0.15in}
	\begin{minipage}{\linewidth} 
		\begin{center} 
			\begin{subfigure}{0.5\linewidth}
				\includegraphics[width=\linewidth]{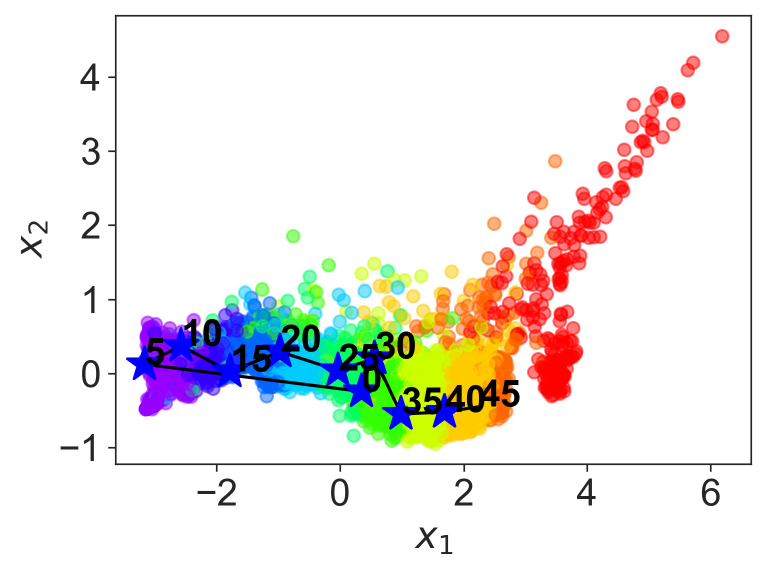}
				\vspace{-0.1in}
				\caption{\texttt{3s5z} }
			\end{subfigure}\hfill
			\begin{subfigure}{0.5\linewidth}
				\includegraphics[width=\linewidth]{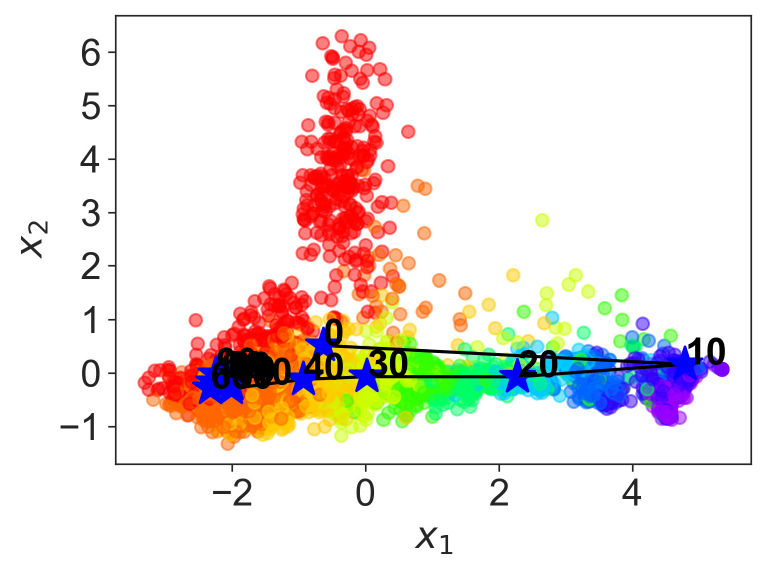}
				\vspace{-0.1in}
				\caption{\texttt{MMM2} }
			\end{subfigure}
		\end{center}	
		\vspace{-0.125in}
		\caption{Episodic incentive. Test trajectories are plotted on the embedded space with sampled memories in $\gD_{E}$, denoted with dotted markers. Star markers and numbers represent the desirability of state and timestep in the episode, respectively. Color represents the same semantics as Figure \ref{fig:t-SNE}.}
		\label{fig:optimality_incentive}
	\end{minipage}
\end{wrapfigure}

Theorem \ref{thm2} also implies that there exists a certain bias in ${\nabla _\theta }L_\theta ^{EC}$ as described in Appendix \ref{proof_thm2}. Besides the property of convergence to the optimal gradient signal presented in Theorem \ref{thm2}, the episodic incentive has the following additional characteristics. (1) The episodic incentive is only applied to the desirable transition. We can simply see that ${r^p} = \gamma \hat \eta  = \gamma {\mathbb{E}_{{\pi _\theta }}}[\eta ] \simeq \gamma \eta_{max} {N_\xi }/{N_{call}}$ and if $\xi (s') = 0$ then ${N_\xi } = 0$, yielding ${r^p} \to 0$. Subsequently, (2) there is no need to adjust a scale factor by the task complexity. (3) The episodic incentive can reduce the risk of overestimation by considering the expected value of ${\mathbb{E}_{{\pi _\theta }}}[\eta ]$. Instead of considering the optimistic $\eta_{max}$, the count-based estimation ${r^p} = \gamma \hat \eta  = \gamma {\mathbb{E}_{{\pi _\theta }}}[\eta ]$ can consider the randomness of the policy $\pi _\theta$.
Figure \ref{fig:optimality_incentive} illustrates how the episodic incentive works with the desirability stored in $\gD_{E}$ constructed by Algorithm \ref{alg:memory_construction} presented in Appendix \ref{memory_contruction}. In Figure \ref{fig:optimality_incentive} as we intended, high-value states (at small timesteps) are clustered close to the purple zone, while low-value states (at large timesteps) are located in the red zone. 

\subsection{Overall Learning Objective}
To construct the joint Q-function $Q_{tot}$ from individual ${Q_i}$ of the agent $i$, any form of mixer can be used. In this paper, we mainly adopt the mixer presented in QPLEX \citep{wang2020qplex} similar to \citet{zheng2021episodic}, which guarantees the complete Individual-Global-Max (IGM) condition \citep{son2019qtran,wang2020qplex}. Considering any intrinsic reward ${r}^c$ encouraging an exploration \citep{zheng2021episodic} or diversity \citep{chenghao2021celebrating}, the final loss function for the action policy learning from Eq. \ref{Eq:Loss_with_rp} can be extended as
\begin{equation}\label{Eq:action_policy_loss}
	\mathcal{L}_{\theta}^p = \left( r(s,\bm{a}) + r^p + \beta_{c} {r}^c + \gamma {{\max }_{{\bm{a}'}}}{Q_{tot}}({s'},{\bm{a}'};\theta^{-}) - {Q_{tot}}(s,{\bm{a}};\theta) \right)^2,
\end{equation}

where $\beta_{c}$ is a scale factor. Note that the episodic incentive $r^p$ can be used in conjunction with any form of intrinsic reward $r^c$ being properly annealed throughout the training. Again, $\theta$ denotes the parameters of networks related to action policy $Q_i$ and the corresponding mixer network to generate $Q_{tot}$. For the action selection via $Q$, we adopt a GRU to encode a local action-observation history $\tau$ presented in \ref{Decentralized POMDP} similar to \citet{sunehag2017value,rashid2018qmix,wang2020qplex}; but in Eq. \ref{Eq:action_policy_loss}, we denote equations with $s$ instead of $\tau$ for the coherence with derivation in the previous section. Appendix \ref{overall_training_algorithm} presents the overall training algorithm.
\vspace{-0.1in}    
\section{Experiments}\label{Experiments}	
\vskip -0.05in
In this part, we have formulated our experiments with the intention of addressing the following inquiries denoted as Q1-3.	
\begin{itemize}
	\item Q1. How does EMU compare to the state-of-the-art MARL frameworks?
	\item Q2. How does the proposed state embedding change the embedding space 
	and improve the performance?
	\item Q3. How does the episodic incentive improve performance?
\end{itemize}	

We conduct experiments on complex multi-agent tasks such as SMAC \citep{samvelyan2019starcraft} and GRF \citep{kurach2020google}. The experiments compare EMU against EMC adopting episodic control \citep{zheng2021episodic}. Also, we include notable baselines, such as value-based MARL methods QMIX \citep{rashid2018qmix}, QPLEX \citep{wang2020qplex}, CDS encouraging individual diversity \citep{chenghao2021celebrating}. Particularly, we emphasize that EMU can be combined with any MARL framework, so we present two versions of EMU implemented on original QPLEX and CDS, denoted as EMU (QPLEX) and EMU (CDS), respectively. Appendix \ref{Implement_experiment_details} provides further details of experiment settings and implementations, and Appendix \ref{additional_applicability} provides the applicability of EMU to single-agent tasks, including pixel-based high-dimensional tasks.

\subsection{Q1. Comparative Evaluation on StarCraft II (SMAC)}\label{performance_on_SMAC}	
\begin{figure}[!h]
	\vskip -0.1in
	\begin{center}
		\centerline{\includegraphics[width=0.85\linewidth]{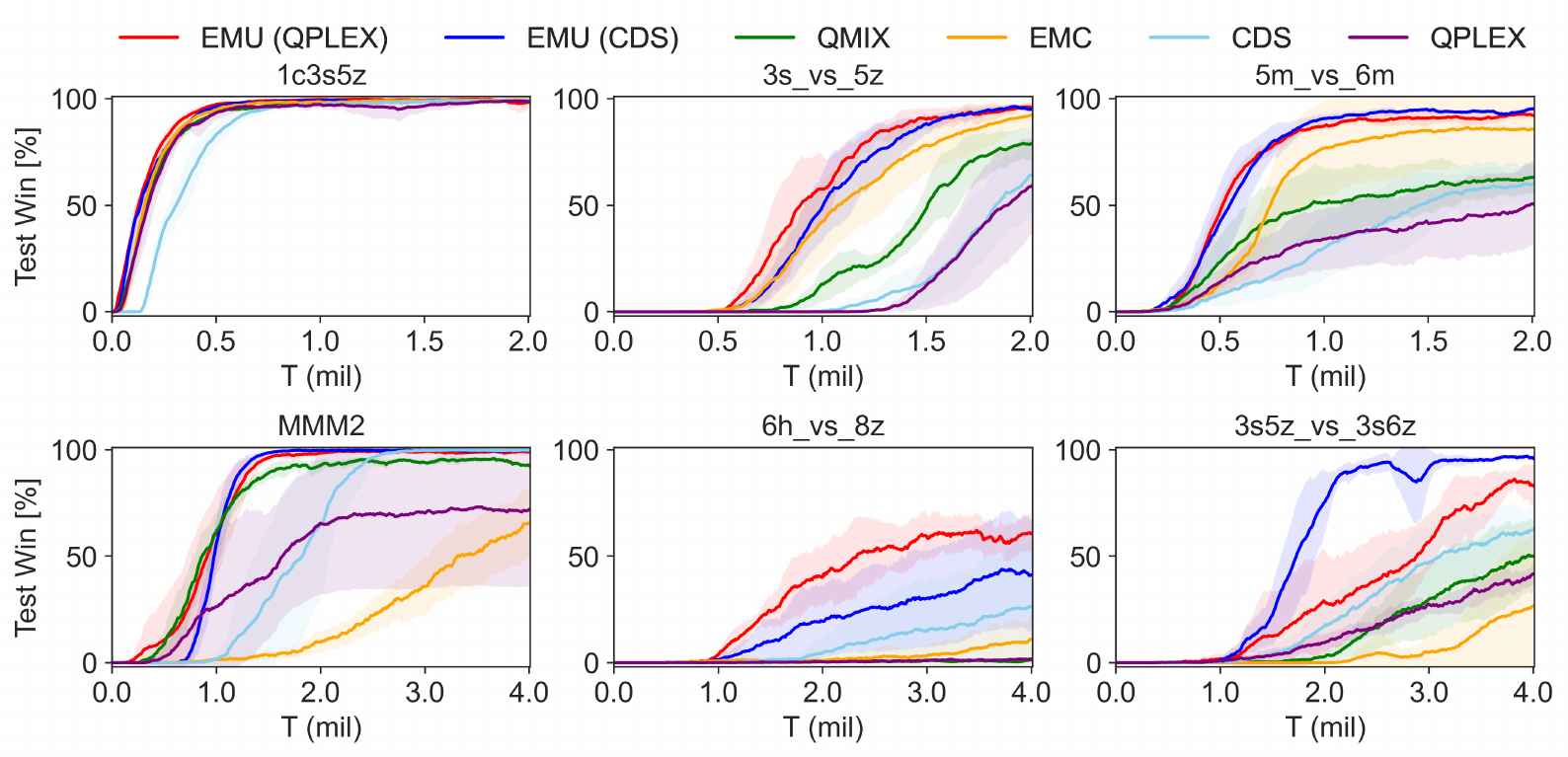}}
		\vskip -0.3in
	\end{center}
	\caption{Performance comparison of EMU against baseline algorithms on three {\bf easy and hard} SMAC maps: \texttt{1c3s5z}, \texttt{3s\_vs\_5z}, and \texttt{5m\_vs\_6m}, and three {\bf super hard} SMAC maps: \texttt{MMM2}, \texttt{6h\_vs\_8z}, and \texttt{3s5z\_vs\_3s6z}.}
	\label{fig:smac_performance}
	\vskip -0.1in
\end{figure}	
Figure \ref{fig:smac_performance} illustrates the overall performance of EMU on various SMAC maps. The map categorization regarding the level of difficulty follows the practice of \citet{samvelyan2019starcraft}. Thanks to the efficient memory utilization and episodic incentive, both EMU (QPLEX) and EMU (CDS) show significant performance improvement compared to their original methodologies. Especially, in {\bf super hard} SMAC maps, the proposed method markedly expedites convergence on optimal policy.

\subsection{Q1. Comparative Evaluation on Google Research Football (GRF)}\label{performance_on_GRF}	
Here, we conduct experiments on GRF to further compare the performance of EMU with other baseline algorithms. In our GRF task, CDS and EMU (CDS) do not utilize the agent's index on observation as they contain the prediction networks while other baselines (QMIX, EMC, QPLEX) use information of the agent's identity in observations. In addition, we do not utilize any additional algorithm, such as prioritized experience replay \citep{schaul2015prioritized}, for all baselines and our method, to expedite learning efficiency. From the experiments, adopting EMU achieves significant performance improvement, and EMU quickly finds the winning or scoring policy at the early learning phase by utilizing semantically similar memory.
\begin{figure}[!h]
\vspace{-0.15in}
\begin{center}			
	\centerline{\includegraphics[width=0.85\linewidth]{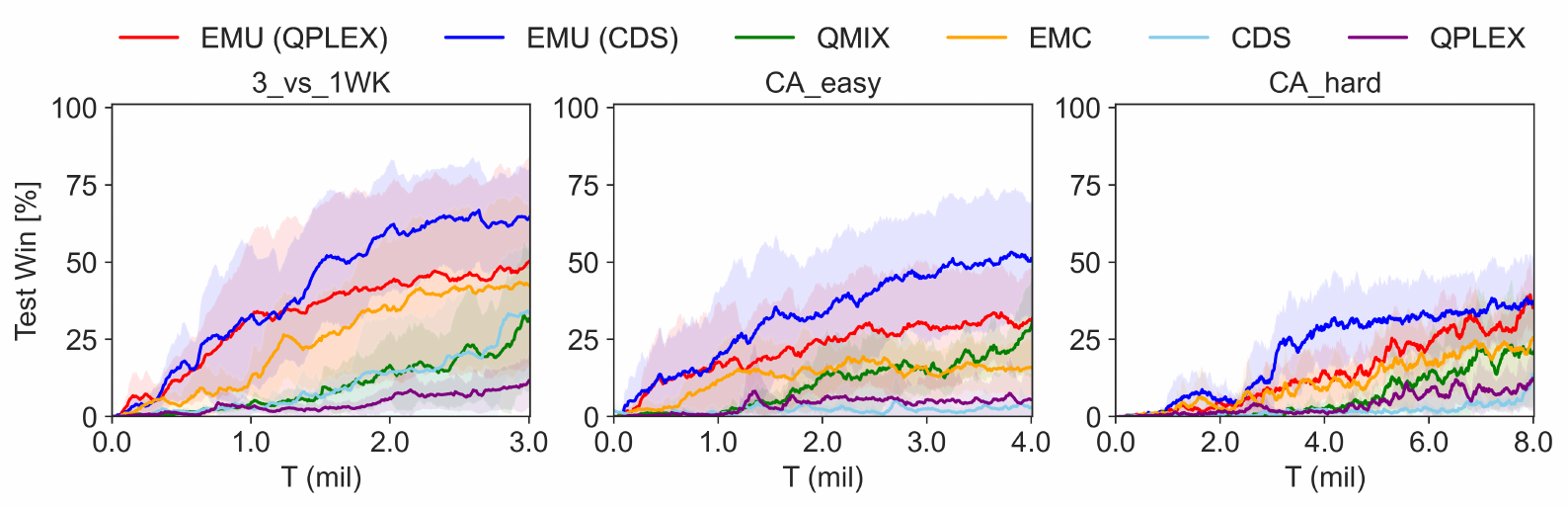}}
\end{center}
\vspace{-0.3in}
\caption{Performance comparison of EMU against baseline algorithms on Google Research Football.}
\label{fig:grf_performance}
\vspace{-0.10in}
\end{figure}

\subsection{Q2. Parametric and Ablation Study}\label{parametric_ablation_study}
\vskip -0.05in
In this section, we examine how the key hyperparameter $\delta$ and the choice of design for $f_{\phi}$ affect the performance. To compare the learning quality and performance more quantitatively, we propose a new performance index called \textit{overall win-rate}, $\bar \mu_{w}$. The purpose of $\bar \mu_{w}$ is to consider both training efficiency (speed) and quality (win-rate) for different seed cases (see Appendix \ref{performance_index} for details).
We conduct experiments on selected SMAC maps to measure $\bar \mu _w$ according to $\delta$ and design choice for $f_{\phi}$ such as (1) random projection, (2) {\bf EmbNet} with Eq. \ref{Eq:Loss_embedder} and (3) {\bf dCAE} with Eq. \ref{Eq:Loss_embedder_dCAE}.

\begin{figure}[!h] 
\vspace{-0.08in}	
\begin{minipage}[t]{0.45\linewidth}
	\begin{center}
		\subfloat[\centering \texttt{3s\_vs\_5z}]{{\includegraphics[width=3.25cm]{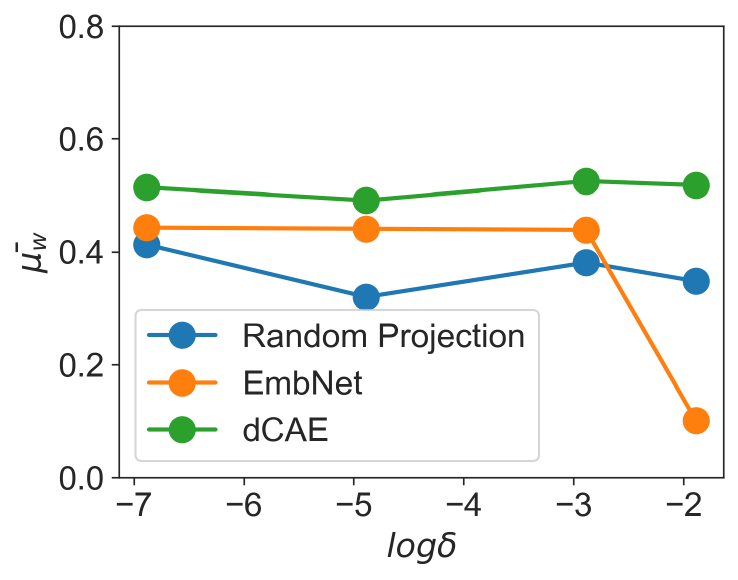} }}%
		\subfloat[\centering \texttt{5m\_vs\_6m}]{{\includegraphics[width=3.25cm]{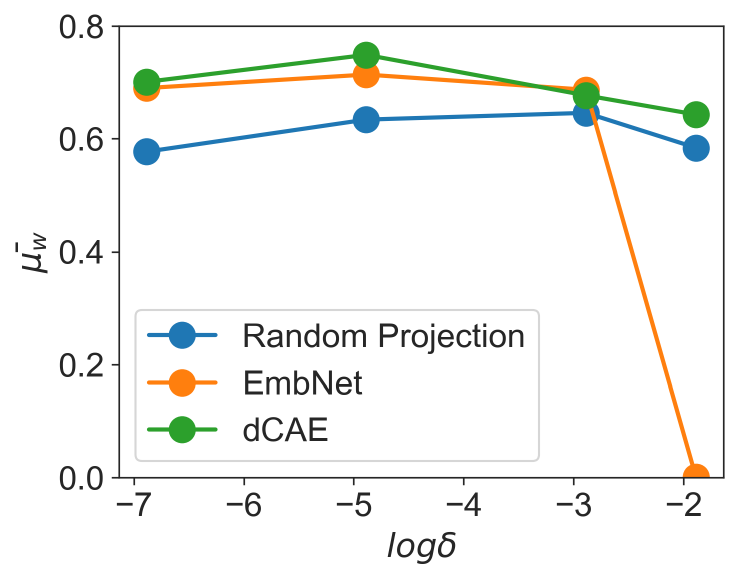} }}%
		\vspace{-0.1in}	
	\end{center}
	\caption{$\bar \mu_{w}$ according to $\delta$ and various design choices for $f_{\phi}$ on SMAC maps.}%
	\label{fig:delta_param}%
\end{minipage}\qquad
\begin{minipage}[t]{0.45\linewidth}
	\begin{center}
		\subfloat[\centering \texttt{3s\_vs\_5z}]{{\includegraphics[width=3.25cm]{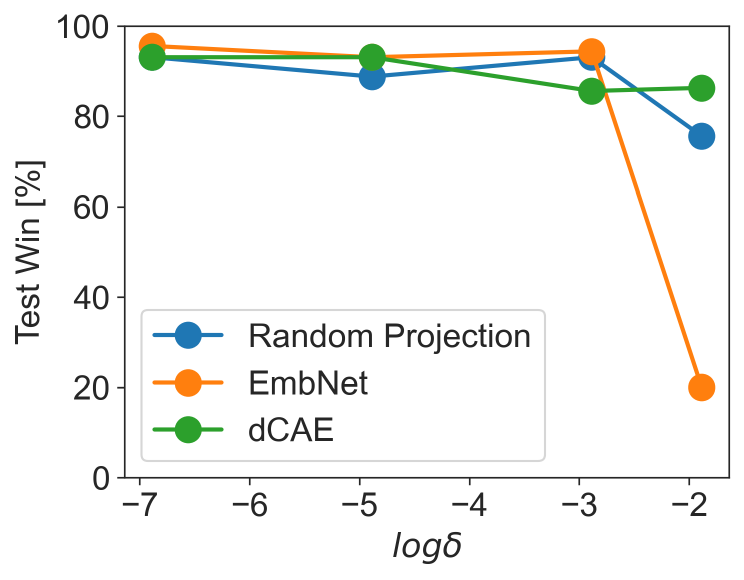} }}%
		\subfloat[\centering \texttt{5m\_vs\_6m}]{{\includegraphics[width=3.25cm]{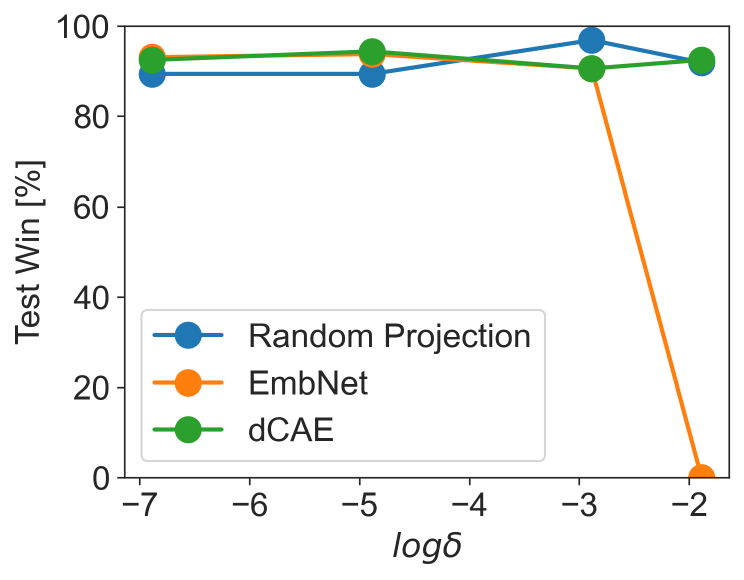} }}%
		\vspace{-0.1in}
	\end{center}
	\caption{Final win-rate according to $\delta$ and various design choices for $f_{\phi}$ on SMAC maps.}%
	\label{fig:winrate_param}%
\end{minipage}
\vspace {-0.1in}
\end{figure} 
Figure \ref{fig:delta_param} and Figure \ref{fig:winrate_param} show $\bar \mu _w$ values and test win-rate at the end of training time according to different $\delta$, presented in log-scale. To see the effect of design choice for $f_{\phi} $ distinctly, we conduct experiments with the conventional episodic control. More data of $\bar \mu _w$ is presented in Tables \ref{Param_study} and \ref{Param_study2} in Appendix \ref{additional_results}. Figure \ref{fig:delta_param} illustrates that dCAE structure shows the best training efficiency throughout various $\delta$ while achieving the optimal policy as other design choices as presented in Figure \ref{fig:winrate_param}. 
\begin{wrapfigure}{R}{0.50\linewidth} 
\vspace{-0.2in}
\begin{minipage}{\linewidth} 
	\begin{center} 
		\begin{subfigure}{0.5\linewidth}
			\includegraphics[width=\linewidth]{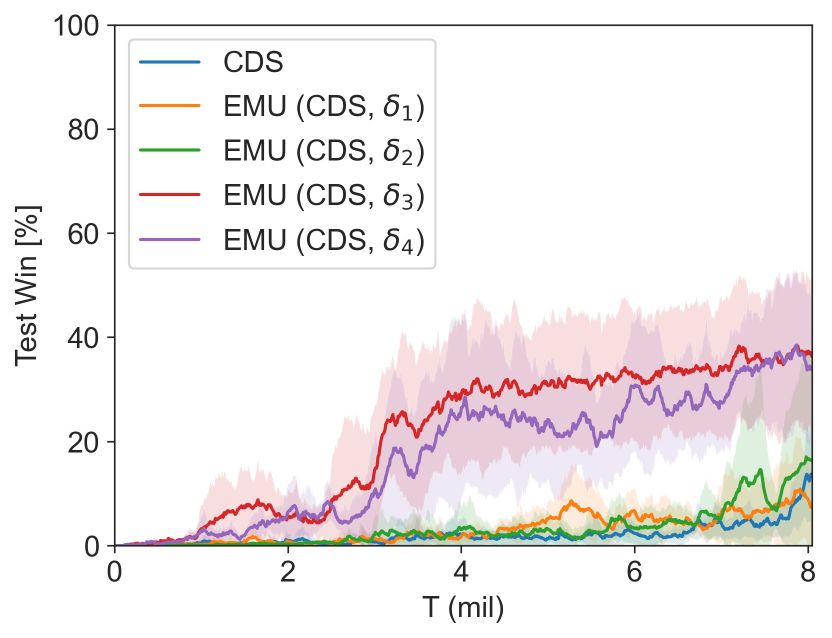}
			\vspace{-0.15in}
			\caption{\texttt{CA\_hard} (GRF)}
		\end{subfigure}\hfill
		\begin{subfigure}{0.5\linewidth}
			\includegraphics[width=\linewidth]{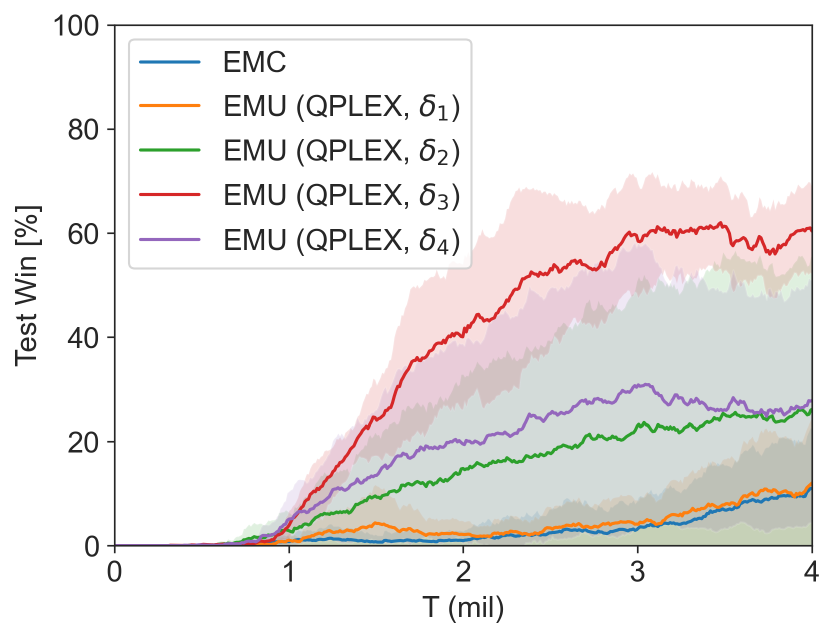}
			\vspace{-0.15in}
			\caption{\texttt{6h\_vs\_8z} (SMAC)}
		\end{subfigure}
	\end{center}	
	\vspace{-0.125in}            
	\caption{Effect of varying $\delta$ on complex MARL tasks.}
	\label{fig:effect_delta}
\end{minipage}
\vspace{-0.15in}
\end{wrapfigure}

Interestingly, dCAE structure works well with a wider range of $\delta$ than EmbNet. We conjecture that EmbNet can select very different states as exploration if those states have similar return $H$ during training.
This excessive memory recall adversely affects learning and fails to find an optimal policy as a result. See Appendix \ref{additional_results} for detailed analysis and Appendix \ref{embedding_loss_ablation} for an ablation study on the loss function of dCAE.

Even though a wide range of $\delta$ works well as in Figures \ref{fig:delta_param} and \ref{fig:winrate_param}, choosing a proper value of $\delta$ in more difficult MARL tasks significantly improves the overall learning performance. Figure \ref{fig:effect_delta} shows the learning curve of EMU according to $\delta_1=1.3e^{-7}$, $\delta_2=1.3e^{-5}$, $\delta_3=1.3e^{-3}$, and $\delta_4=1.3e^{-2}$. In super hard MARL tasks such as \texttt{6h\_vs\_8z} in SMAC and \texttt{CA\_hard} in GRF, $\delta_3$ shows the best performance compared to other $\delta$ values. This is consistent with the value suggested in Appendix \ref{memory_utilization}, where $\delta$ is determined in a memory-efficient way. Further parametric study on $\delta$ and $\lambda_{rcon}$ are presented in Appendix \ref{additional_param} and \ref{parametric_lambda_rcon}, respectively.

\subsection{Q3. Further Ablation Study} 
\vskip -0.05in
In this section, we carry out further ablation studies to see the effect of episodic incentive $r^p$ presented in Section \ref{Episodic_incentive}. From EMU (QPLEX) and EMU (CDS), we ablate the episodic incentive and denote them with {\bf (No-EI)}. We additionally ablate embedding network $f_{\phi}$ from EMU and denote them with {\bf (No-SE)}. In addition, we ablate both parts, yielding EMC (QPLEX-original) and CDS (QPLEX-original). We evaluate the performance of each model on super hard SMAC maps. Additional ablation studies on GRF maps are presented in Appendix \ref{additional_ablation}. Note that EMC (QPLEX-original) utilizes the conventional episodic control presented in \citet{zheng2021episodic}.

\begin{figure}[!h] 
\begin{center}
	\vspace{-0.075in}
	\begin{minipage}[!h]{0.9\linewidth} 
		\subfloat[\centering \texttt{6h\_vs\_8z} SMAC ]{{\includegraphics[width=0.33\linewidth]{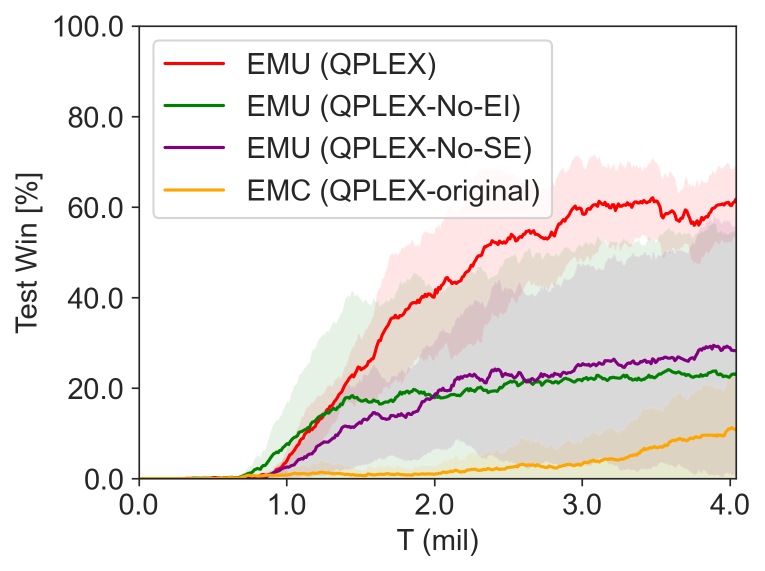} }}
		\subfloat[\centering \texttt{3s5z\_vs\_3s6z} SMAC ]{{\includegraphics[width=0.33\linewidth]{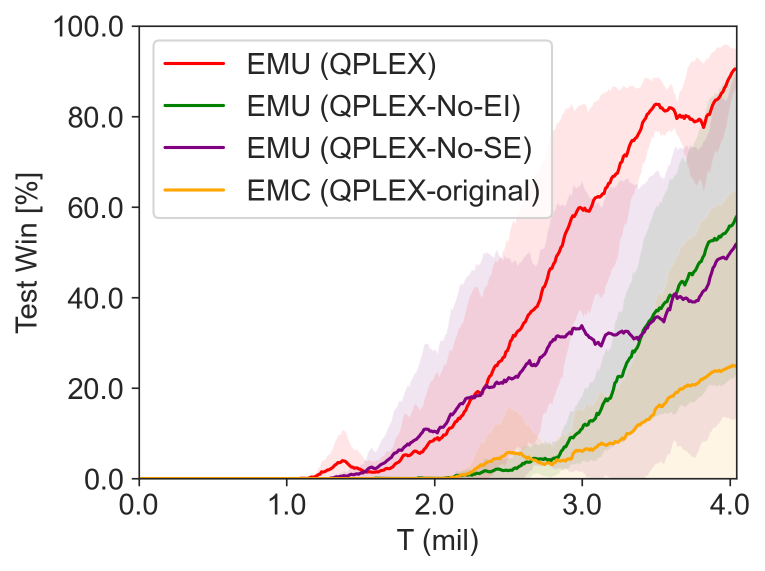} }}
		\subfloat[\centering \texttt{3s5z\_vs\_3s6z} SMAC ]{{\includegraphics[width=0.33\linewidth]{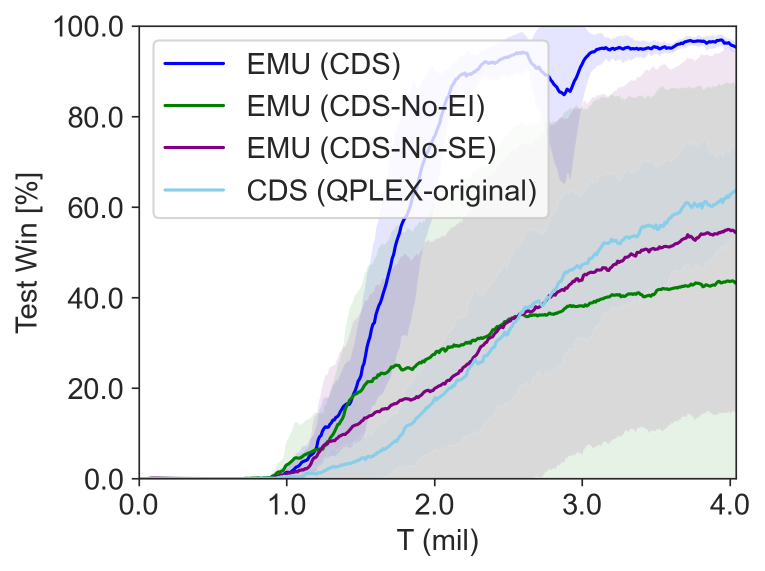} }}
	\end{minipage}
\end{center}
\vspace{-0.1in}
\caption{Ablation studies on episodic incentive via complex MARL tasks.}%
\label{fig:ablation_example}%
\vspace{-0.15in}
\end{figure}

Figure \ref{fig:ablation_example} illustrates that the episodic incentive largely affects learning performance. Especially, EMU (QPLEX-No-EI) and EMU (CDS-No-EI) utilizing the conventional episodic control show a large performance variation according to different seeds. This demonstrates that a naive utilization of episodic control could be detrimental to learning an optimal policy. On the other hand, the episodic incentive selectively encourages transition considering desirability and thus prevents such a local convergence. Appendix \ref{additional_SE_ablation} and \ref{additional_rc_ablation} present an additional ablation  study on semantic embedding and $r^c$, respectively. In addition, Appendix \ref{additional_incentive_ablation} presents a comparison with an alternative incentive \citep{henaff2022exploration} presented in a single-agent setting.

\subsection{Qualitative Analysis and Visualization}
\begin{wrapfigure}{tL}{0.6\textwidth}
\begin{minipage}[t]{0.6\columnwidth}
	\subfloat[\centering Desirable trajectory on \texttt{5m\_vs\_6m} SMAC map ]{{\includegraphics[width=\linewidth]{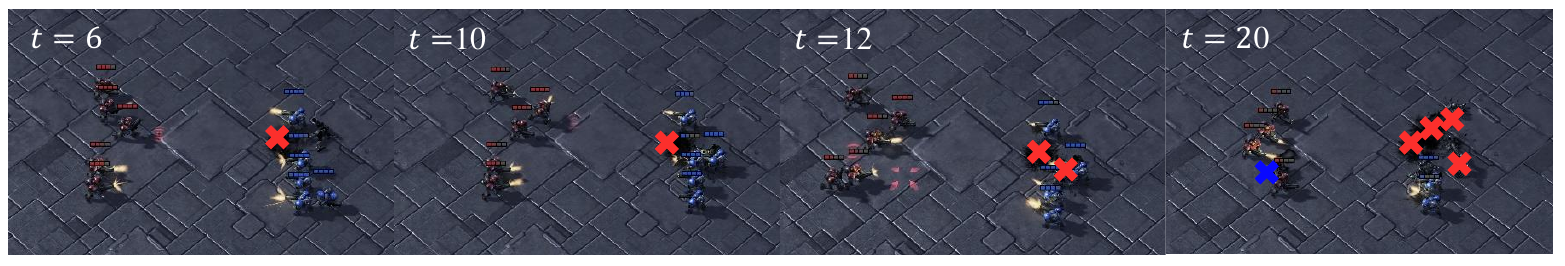} }} \\
	\subfloat[\centering Undesirable trajectory on \texttt{5m\_vs\_6m} SMAC map]{{\includegraphics[width=\linewidth]{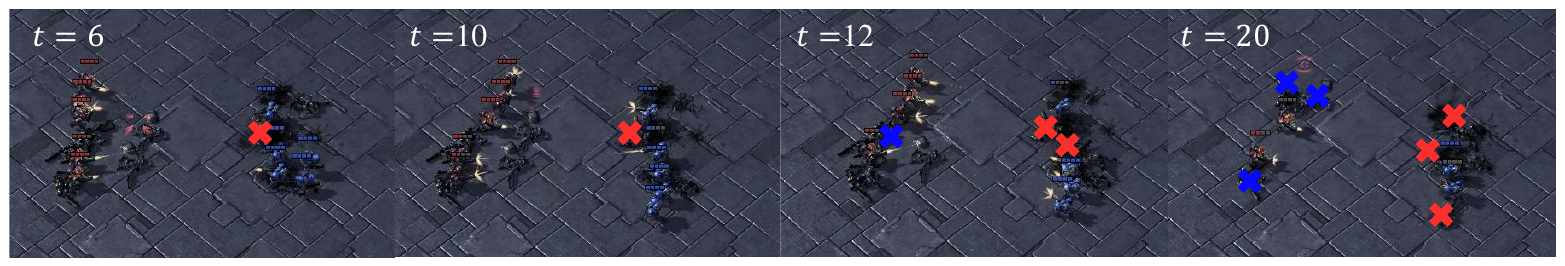} }}		
	\caption{Visualization of test episodes.}
	\label{fig:qualitative}
\end{minipage}
\end{wrapfigure}

\begin{wrapfigure}{tR}{0.5\textwidth}
\vspace{-0.2in}
\begin{minipage}{0.5\textwidth}
	\begin{center}
		\begin{subfigure}{0.5\textwidth}
			\includegraphics[width=\linewidth]{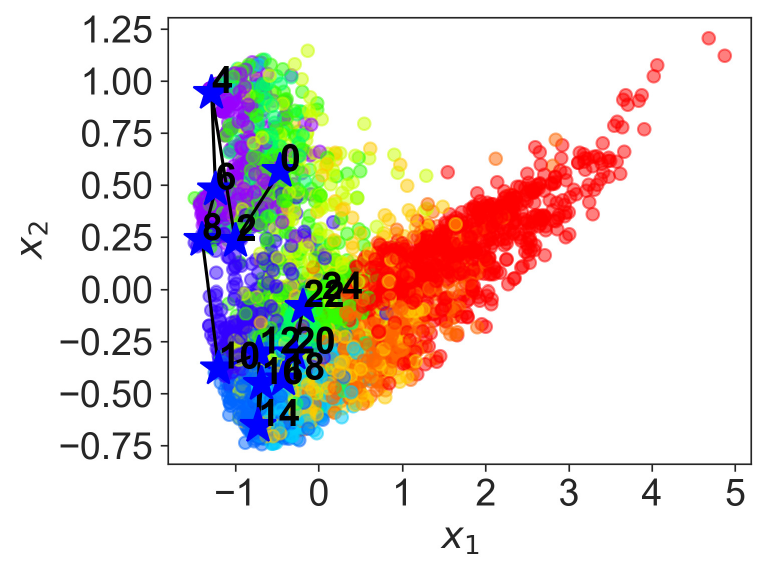}
			\vskip -0.1in
			\caption{Desirable trajectory}
		\end{subfigure}
		\begin{subfigure}{0.5\textwidth}
			\includegraphics[width=\linewidth]{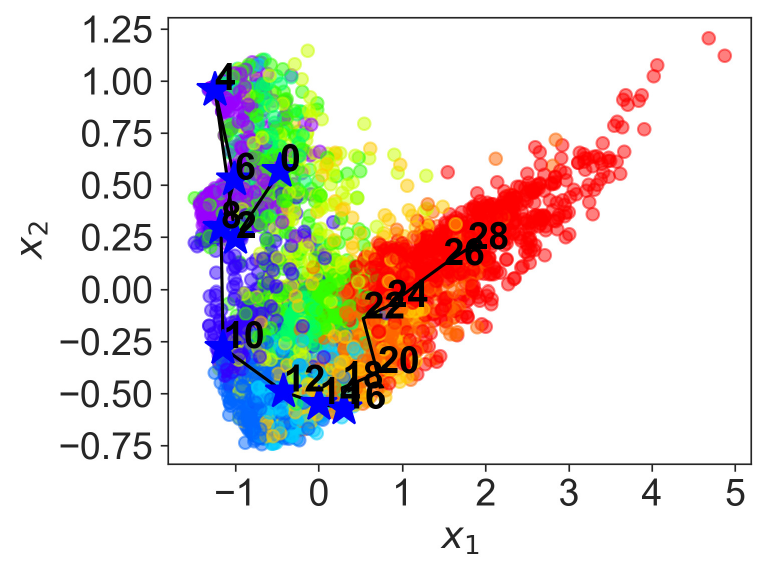}
			\vskip -0.1in
			\caption{Undesirable trajectory}
		\end{subfigure}		
		\vskip -0.05in
		\caption{Test trajectories on embedded space of $\gD_{E}$.}
		\label{fig:qualitative_embdding}
	\end{center}
\end{minipage}
\vspace{-0.4in}
\end{wrapfigure}

\vspace{-0.05in}
In this section, we conduct analysis with visualization to check how the desirability $\xi$ is memorized in $\gD_{E}$ and whether it conveys correct information. Figure \ref{fig:qualitative} illustrates two test scenarios with different seeds, and each snapshot is denoted with a corresponding timestep. In Figure \ref{fig:qualitative_embdding}, the trajectory of each episode is projected onto the embedded space of $\gD_{E}$. 
In Figure \ref{fig:qualitative}, case (a) successfully defeated all enemies, whereas case (b) lost the engagement. Both cases went through a similar, desirable trajectory at the beginning. For example, until $t=10$ agents in both cases focused on killing one enemy and kept all ally agents alive at the same time. However, at $t=12$, case (b) lost one agent, and two trajectories of case (a) and (b) in embedded space began to bifurcate. Case (b) still had a chance to win around $t=14 \sim 16$. However, the states became \textit{undesirable} (denoted without star marker) after losing three ally agents around $t=20$, and case (b) lost the battle as a result. These sequences and characteristics of trajectories are well captured by desirability $\xi$ in $\gD_{E}$ as illustrated in Figure \ref{fig:qualitative_embdding}.        

Furthermore, the desirable state denoted with $\xi=1$ encourages exploration around it though it is not directly retrieved during batch sampling. This occurs through the propagation of its desirability to states currently distinguished as undesirable during memory construction, using Algorithm \ref{alg:memory_construction} in Appendix \ref{memory_contruction}. Consequently, when the state's desirability is precisely memorized in $\gD_{E}$, it can encourage desirable transitions through the episodic incentive $r^p$.
\section{Conclusion}
\label{conclusion}
\vskip -0.1in
This paper presents EMU, a new framework to efficiently utilize episodic memory for cooperative MARL. 
EMU introduces two major components: 1) a trainable semantic embedding and 2) an episodic incentive utilizing desirability of state. Semantic memory embedding allows us to safely utilize similar memory in a wide area, expediting learning via exploratory memory recall. 
The proposed episodic incentive selectively encourages desirable transitions and reduces the risk of local convergence by leveraging the desirability of the state. As a result, there is no need for manual hyperparameter tuning according to the complexity of tasks, unlike conventional episodic control. Experiments and ablation studies validate the effectiveness of each component of EMU. 


\newpage

\section*{Acknowledgements}
\label{acknowledgements}
This research was supported by AI Technology Development for Commonsense Extraction, Reasoning, and Inference from Heterogeneous Data(IITP) funded by the Ministry of Science and ICT(2022-0-00077).

\bibliography{emu_reference}
\bibliographystyle{iclr2024_conference}

\newpage
\appendix
\newpage	

\section{Related Works}\label{related_works}
This section presents the related works regarding incentive generation for exploration, episodic control, and the characteristics of cooperative MARL.

\subsection{Incentive for Multi-agent Exploration}
Balancing between exploration and exploitation in policy learning is a paramount issue in reinforcement learning. To encourage exploration, modified count-based methods \citep{bellemare2016unifying, ostrovski2017count, tang2017exploration}, prediction error-based methods \citep{stadie2015incentivizing, pathak2017curiosity, burda2018exploration, kim2018emi}, and information gain-based methods \citep{mohamed2015variational, houthooft2016vime} have been proposed for a single agent reinforcement learning. In most cases, an incentive for exploration is introduced as an additional reward to a TD target in Q-learning; or such an incentive is added as a regularizer for overall loss functions. Recently, various aforementioned methods to encourage exploration have been adopted to the multi-agent setting \citep{mahajan2019maven, wang2019influence, jaques2019social, mguni2021ligs} and have shown their effectiveness. MAVEN \citep{mahajan2019maven} introduces a regularizer maximizing the mutual information between trajectories and latent variables to learn a diverse set of behaviors. LIIR \citep{du2019liir} learns a parameterized individual intrinsic reward function by maximizing a centralized critic. CDS \citep{chenghao2021celebrating} proposes a novel information-theoretical objective to maximize the mutual information between agents’ identities and trajectories to encourage diverse individualized behaviors. EMC \citep{zheng2021episodic} proposes a curiosity-driven exploration by predicting individual Q-values. This individual-based Q-value prediction can capture the influence among agents as well as the novelty of states.	

\subsection{Episodic Control}
Episodic control \citep{lengyel2007hippocampal} was well adopted on model-free setting \citep{blundell2016model} by storing the highest return of a given state, to efficiently estimate its values or Q-values. Given that the sample generation is often limited by simulation executions or real-world observations, its sample efficiency helps to find an accurate estimation of Q-value \citep{blundell2016model, pritzel2017neural, lin2018episodic}. NEC \citep{pritzel2017neural} uses a differentiable neural dictionary as an episodic memory to estimate the action value by the weighted sum of the values in the memory. EMDQN \citep{lin2018episodic} utilizes a fixed random matrix to generate a state representation, which is used as a key to link between the state representation and the highest return of the state in the episodic memory. ERLAM \citep{zhu2020episodic} learns associative memories by building a graphical representation of states in memory, and GEM \citep{hu2021generalizable} develops state-action values of episodic memory in a generalizable manner. Recently, EMC \citep{zheng2021episodic} extended the approach of EMDQN to a deep MARL with curiosity-driven exploration incentives. EMC utilizes episodic memory to regularize policy learning and shows performance improvement in cooperative MARL tasks. However, EMC requires a hyperparameter tuning to determine the level of importance of the one-step TD memory-based target during training, according to the difficulties of tasks. In this paper, we interpret this regularization as an additional transition reward. Then, we present a novel form of reward, called episodic incentive, to selectively encourage the transition toward desired states, i.e., states toward a common goal in cooperative multi-agent tasks.

\subsection{Cooperative Multi-agent Reinforcement Learning (MARL) task}
In general, there is a common goal in cooperative MARL tasks, which guarantees the maximum return that can be obtained from the environment. Thus, there could be many local optima with high returns but not the maximum, which means the agents failed to achieve the common goal in the end. In other words, there is a distinct difference between the objective of cooperative MARL tasks and that of a single-agent task, which aims to maximize the return as much as possible without any boundary determining success or failure. Our desirability definition presented in Definition \ref{def: desirable trajectory} in MARL setting becomes well justified from this view. Under this characteristic of MARL tasks, learning optimal policy often takes a long time and even fails, yielding a local convergence. EMU was designed to alleviate these issues in MARL. 

\section{Mathematical Proof}\label{mathe_proof}
In this section, we present the omitted proofs of Theorem \ref{thm1} and Theorem \ref{thm2} as follows.
\subsection{Proof of Theorem \ref{thm1}}
\label{proof_thm1}
\begin{proof}
The loss function of a conventional episodic control, $	L_\theta ^{EC}$, can be expressed as the weighted sum of one-step inference TD error $\Delta {\varepsilon _{TD}} = r(s,a) + \gamma {\max _{a'}}{Q_{{\theta ^ - }}}(s',a') - {Q_\theta }(s,a)$ and MC inference error $\Delta {\varepsilon _{EC}} = {Q_{EC}}(s,a) - {Q_\theta }(s,a)$.
\begin{equation}\label{Eq:Loss_EC}
	L_\theta ^{EC} = {(r(s,\bm{a}) + \gamma {\max _{\bm{a}'}}{Q_{{\theta^{-} }}}(s',\bm{a}') - {Q_\theta }(s,\bm{a}))^2} + \lambda {({Q_{EC}}(s,\bm{a}) - {Q_\theta }(s,\bm{a}))^2},
\end{equation}  
where ${Q_{EC}}(s,\bm{a}) = r(s,\bm{a}) + \gamma H(s')$ and ${Q_{{\theta^{-} }}}$ is the target network parameterized by $\theta ^-$. Then, the gradient of $L_\theta ^{EC}$ can be derived as 
\begin{equation}\label{Eq:grad_Loss_EC}
	\begin{aligned}
		{\nabla _\theta }L_\theta ^{EC} &=  - 2{\nabla _\theta }{Q_\theta }(s,\bm{a})[(r(s,\bm{a}) + \gamma {\max _{a'}}{Q_{{\theta ^ - }}}(s',\bm{a}') - {Q_\theta }(s,\bm{a})) + \lambda ({Q_{EC}}(s,\bm{a}) - {Q_\theta }(s,\bm{a}))] \\
		&=  - 2{\nabla _\theta }{Q_\theta }(s,\bm{a})(\Delta {\varepsilon _{TD}} + \lambda \Delta {\varepsilon _{EC}}).
	\end{aligned}
\end{equation} 

Now, we consider an additional reward $r^{EC}$ for the transition to a conventional Q-learning objective, the modified loss function ${L_\theta }$ can be expressed as
\begin{equation}\label{Eq:Loss_with_add_r}
	{L_\theta } = {(r(s,\bm{a}) + {r^{EC}}(s,\bm{a},s') + \gamma {\max _{\bm{a}'}}{Q_{{\theta ^ - }}}(s',\bm{a}') - {Q_\theta }(s,\bm{a}))^2}.
\end{equation} 
Then, the gradient of ${L_\theta }$ presented in Eq. \ref{Eq:Loss_with_add_r} is computed as 
\begin{equation}\label{Eq:grad_Loss_with_add_r}
	{\nabla _\theta }{L_\theta } =  - 2{\nabla _\theta }{Q_\theta }(s,\bm{a})(\Delta {\varepsilon _{TD}} + {r^{EC}}).
\end{equation}  
Comparing Eq. \ref{Eq:grad_Loss_EC} and Eq. \ref{Eq:grad_Loss_with_add_r}, if we set the additional transition reward as ${r^{EC}}(s,\bm{a},s') = \lambda (r(s,\bm{a}) + \gamma H(s') - {Q_\theta }(s,\bm{a}))$, then ${\nabla _\theta }{L_\theta } = {\nabla _\theta }L_\theta ^{EC}$ holds. 
\end{proof}

\subsection{Proof of Theorem \ref{thm2}}
\label{proof_thm2}
\begin{proof}
From Eq. \ref{Eq:exp_eta}, the value of $\hat \eta(s')$ can be expressed as 
\begin{equation}\label{Eq:expected_eta}
	\hat \eta(s')  = {\mathbb{E}_{{\pi _\theta }}}[\eta(s') ] \simeq \frac{{{N_\xi }(s')}}{{{N_{call}}(s')}}\bigl(H(f_{\phi}(s')) - {\max _{a'}}{Q_{{\theta ^ {-} }}}(s',a')\bigr)].
\end{equation}
When the joint actions from the current time follow the optimal policy, $\bm{a}\sim \pi_{\theta}^{*}$, the cumulative reward from $s'$ converges to ${V^{*}}(s')$, i.e., $H(f_{\phi}(s')) \to {V^{*}}(s')$. Then, every recall of $\hat x' = \operatorname{NN}({f_\phi }(s')) \in {\gD_E}$ guarantees the desirable transition, i.e., $\xi(s')=1$, where $\operatorname{NN}(\cdot)$ represents a function for selecting the nearest neighbor. As a result, as $N_{call}(s') \to \infty$, $\frac{{{N_\xi }(s')}}{{{N_{call}}(s')}} \to 1$, yielding $\hat \eta(s') \simeq \frac{{{N_\xi }(s')}}{{{N_{call}}(s')}}\bigl(H(f_{\phi}(s')) - {\max _{a'}}{Q_{{\theta ^ {-} }}}(s',a')\bigr) \to V^*(s') - {\max _{\bm{a}'}}{Q_{{\theta ^{-} }}}(s',\bm{a}')$. Then, the gradient signal with the episodic incentive ${\nabla _\theta }L_\theta ^p$ becomes
\begin{equation}\label{Eq:grad_Lp_derived}
	\begin{aligned}
		{\nabla _\theta }L_\theta ^p & =- 2{\nabla _\theta }{Q_\theta }(s,\bm{a})[\Delta {\varepsilon _{TD}} + r^p ]\\
		&=  - 2{\nabla _\theta }{Q_\theta }(s,\bm{a})[\Delta {\varepsilon _{TD}} + \gamma \hat \eta(s') ]\\
		&\simeq  - 2{\nabla _\theta }{Q_\theta }(s,\bm{a})[\Delta {\varepsilon _{TD}} + \gamma ({V^*}(s') - {\max _{\bm{a}'}}{Q_{{\theta ^ - }}}(s',\bm{a}'))]\\
		&=  - 2{\nabla _\theta }{Q_\theta }(s,\bm{a})[r(s,\bm{a}) + \gamma {\max _{\bm{a}'}}{Q_{{\theta ^ - }}}(s',\bm{a}') - {Q_\theta }(s,\bm{a}) + \gamma ({V^*}(s') - {\max _{\bm{a}'}}{Q_{{\theta ^ - }}}(s',\bm{a}'))]\\
		&= - 2{\nabla _\theta }{Q_\theta }(s,\bm{a})[r(s,\bm{a}) + \gamma {V^*}(s') - {Q_\theta }(s,\bm{a})]\\
		&= {\nabla _\theta }L_\theta ^*,
	\end{aligned}
\end{equation}
which completes the proof.	
\end{proof}
In addition, when ${\max _{\bm{a}'}}{Q_{{\theta ^{-} }}}(s',\bm{a}')$ accurately estimates $V^*(s')$, the original TD-target is preserved as the episodic incentive becomes zero, i.e., $r^p \rightarrow 0$. Then with the properly annealed intrinsic reward $r^c$, the learning objective presented in Eq. \ref{Eq:action_policy_loss} degenerates to the original Bellman optimality equation \citep{sutton2018reinforcement}. On the other hand, even though the assumption of $H(s') \to {V^{*}}(s')$ yields $\Delta {\varepsilon _{EC}} \to \Delta {\varepsilon _{TD}^{*}}$, ${\nabla _\theta }L_\theta ^{EC}$ has an additional bias $\Delta {\varepsilon _{TD}}$ due to weighted sum structure presented in Eq. \ref{Eq:Episodic_loss}. Thus, ${\nabla _\theta }L_\theta ^{EC}$ can converge to ${\nabla _\theta }L_\theta ^{*}$ only when ${\max _{\bm{a}'}}{Q_{{\theta ^{-} }}}(s',\bm{a}') \to V^*(s')$ and $\lambda \to 0$ at the same time.

\section{Implementation and Experiment Details}\label{Implement_experiment_details}
\subsection{Details of Implementation}\label{implementation_details}
\textbf{Encoder and Decoder Structure}\\    
As illustrated in Figure \ref{fig:overview}(c), we have an encoder and decoder structure. For an encoder $f_{\phi}$, we evaluate two types of structure, \textbf{EmbNet} and \textbf{dCAE}. For \textbf{EmbNet} with the learning objective presented in Eq. \ref{Eq:Loss_embedder}, two fully connected layers with 64-dimensional hidden state are used with ReLU activation function between them, followed by layer normalization block at the head. On the other hand, for \textbf{dCAE} with the learning objective presented in Eq. \ref{Eq:Loss_embedder_dCAE}, we utilize a deeper encoder structure which contains three fully connected layers with ReLU activation function. In addition, \textbf{dCAE} takes a timestep $t$ as an input as well as a global state $s_t$. We set episodic latent dimension $\dim(x)=4$ as \citet{zheng2021episodic}. 

\begin{figure}[!htbp] 
\begin{center}
	\begin{minipage}[b]{0.9\linewidth}		
		\subfloat[\centering EmbNet]{{\includegraphics[width=0.45\linewidth]{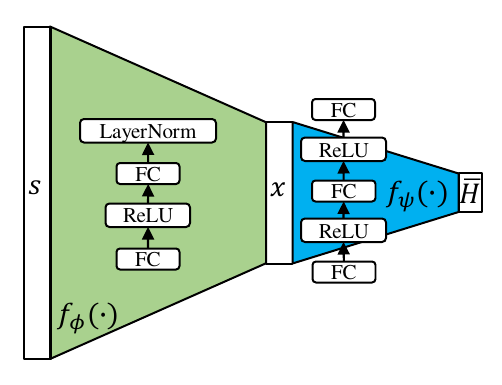} }}
		\qquad
		\subfloat[\centering dCAE ]{{\includegraphics[width=0.45\linewidth]{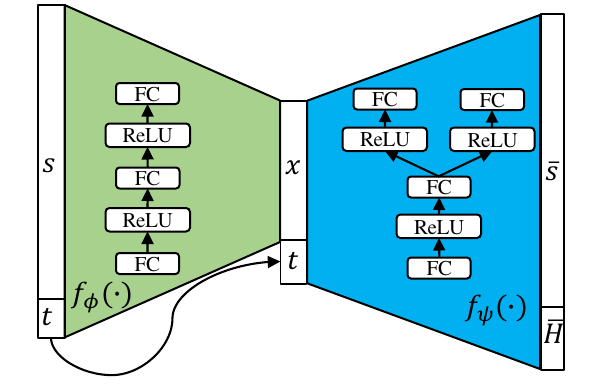} }}%
	\end{minipage}
\end{center}
\caption{Illustration of network structures.}%
\label{fig:encoder_structure}%
\end{figure}

For a decoder $f_{\psi}$, both \textbf{EmbNet} and \textbf{dCAE} utilize a three-fully connected layer with ReLU activation functions. Differences are that \textbf{EmbNet} takes only $x_t$ as input and utilizes the 128-dimensional hidden state while \textbf{dCAE} takes $x_t$ and $t$ as inputs and adopts the 64-dimensional hidden state. As illustrated in Figure \ref{fig:overview}(c), to reconstruct global state $s_t$, \textbf{dCAE} has two separate heads while sharing lower parts of networks; $f_{\phi}^s$ to reconstruct $s_t$ and $f_{\phi}^H$ to predict the return of $s_t$, denoted as $H_t$. Figure \ref{fig:encoder_structure} illustrates network structures of \textbf{EmbNet} and \textbf{dCAE}. The concept of supervised VAE similar to EMU can be found in \citep{le2018supervised}. 

The reason behind avoiding probabilistic autoencoders such as variational autoencoder (VAE) \citep{kingma2013auto,sohn2015learning} is that the stochastic embedding and the prior distribution could have an adverse impact on preserving a pair of $x_t$ and $H_t$ for given a $s_t$. In particular, with stochastic embedding, a fixed $s_t$ can generate diverse $x_t$. As a result, it breaks the pair of $x_t$ and $H_t$ for given $s_t$, which makes it difficult to select a valid memory from $\gD_E$.

For training, we periodically update $f_{\phi}$ and $f_{\psi}$ with an update interval of $t_{emb}$ in parallel to MARL training. At each training phase, we use $M_{emb}$ samples out of the current capacity of $\gD_{E}$, whose maximum capacity is 1 million (1M), and batch size of $m_{emb}$ is used for each training step. After updating $f_{\phi}$, every $x \in \gD_{E}$ needs to be updated with updated $f_{\phi}$. Algorithm \ref{alg:embedder_training} shows the details of learning framework for $f_{\phi}$ and $f_{\psi}$.
Details of the training procedure for $f_{\phi}$ and $f_{\psi}$ along with MARL training are presented in Appendix \ref{overall_training_algorithm}. 

\textbf{Other Network Structure and Hyperparameters}\\
For a mixer structure, we adopt QPLEX \citep{wang2020qplex} in both EMU (QPLEX) and EMU (CDS) and follow the same hyperparameter settings used in their source codes. Common hyperparameters related to individual Q-network and MARL training are adopted by the default settings of PyMARL \citep{samvelyan2019starcraft}.

\begin{minipage}{\linewidth} 
\begin{algorithm}[H]
	\begin{center}
		\begin{algorithmic}[1]				
			\State \textbf{Parameter:} learning rate $\alpha$, number of training dataset $N$, batch size $B$
			\State Sample Training dataset ${(s^{(i)}, H^{(i)}, t^{(i)})}_{i=1}^N \sim \gD_{E}$, 
			\State Initialize weights $\bm{\phi}, \bm{\psi} \leftarrow \mathbf{0}$
			\For{$i=1$ to $\lfloor N/B \rfloor$}
			\State Compute $(x^{(j)} = f_{\bm{\phi}}(s^{(j)}|t^{(j)})_{j=(i-1)B+1}^{iB}$
			\State Predict return $(\bar{H}^{(j)} = f_{\bm{\psi}}^H(x^{(j)}| t^{(i)}) )_{j=(i-1)B+1}^{iB}$
			\State Reconstruct state $(\bar{s}^{(j)} = f_{\bm{\psi}}^s(x^{(j)})| t^{(i)} )_{j=(i-1)B+1}^{iB}$
			\State Compute Loss $\mathcal{L}(\bm{\phi},\bm{\psi})$ via Eq. \ref{Eq:Loss_embedder_dCAE}
			\State Update $\bm{\phi}\leftarrow \bm{\phi}-\alpha\frac{\partial\mathcal{L}}{\partial\bm{\phi}}$, $\bm{\psi}\leftarrow \bm{\psi}-\alpha\frac{\partial\mathcal{L}}{\partial\bm{\psi}}$
			\EndFor	
			\caption{Training Algorithm for State Embedding}
			\label{alg:embedder_training}		
		\end{algorithmic}
	\end{center}
\end{algorithm}
\end{minipage}

\subsection{Experiment Details}\label{experiment_details}
We utilize PyMARL \citep{samvelyan2019starcraft} to execute all of the baseline algorithms with their open-source codes, and the same hyperparameters are used for experiments if they are presented either in uploaded codes or in their manuscripts.

For a general performance evaluation, we test our methods on various maps, which require a different level of coordination according to the map's difficulties. Win-rate is computed with 160 samples: 32 episodes for each training random seed, and 5 different random seeds unless denoted otherwise.

Both the mean and the variance of the performance are presented for all the figures to show their overall performance according to different seeds. Especially for a fair comparison, we set $n_\textrm{circle}$, the number of training per a sampled batch of 32 episodes during training, as 1 for all baselines since some of the baselines increase $n_\textrm{circle}=2$ as a default setting in their codes. 

For performance comparison with baseline methods, we use their codes with fine-tuned algorithm configuration for hyperparameter settings according to their codes and original paper if available. For experiments on SMAC, we use the version of \texttt{starcraft.py} presented in RODE \citep{wang2020rode} adopting some modification for compatibility with QPLEX \citep{wang2020qplex}. All SMAC experiments were conducted on StarCraft II version 4.10.0 in a Linux environment.	

For Google research football task, we use the environmental code provided by \citep{kurach2020google}. In the experiments, we consider three official scenarios such as academy\_3\_vs\_1\_with\_keeper (3\_vs\_1WK), academy\_counterattack\_easy (CA\_easy), and academy\_counterattack\_hard (CA\_hard).

In addition, for controlling $r^c$ in Eq. \ref{Eq:action_policy_loss}, the same hyperparameters related to curiosity-based \citep{zheng2021episodic} or diversity-based exploration \cite{chenghao2021celebrating} are adopted for EMU (QPLEX) and EMU (CDS) as well as for baselines EMC and CDS. After further experiments, we found that the curiosity-based $r^c$ from \citep{zheng2021episodic} adversely influenced super hard SMAC task, with the exception of \texttt{corridor} scenario. Furthermore, the diversity-based exploration from \cite{chenghao2021celebrating} led to a decrease in performance on both easy and hard SMAC maps. Thus, we decided to exclude the use of $r^c$ for EMU (QPLEX) on the super hard SMAC task and for EMU (CDS) on the easy/hard SMAC maps.
EMU set task-dependent $\delta$ values as presented in Table \ref{tab:task_dependent_hyper_EMU}. For other hyperparameters introduced by EMU, the same values presented in Table \ref{tab:hyperparam_emb} are used throughout all tasks. For EMU (QPLEX) in \texttt{corridor}, $\delta=1.3e-5$ is used instead of $\delta=1.3e-3$.

\begin{table}[htbp]
\caption{Task-dependent hyperparameter of EMU.}
\vspace{0.3cm}
\label{tab:task_dependent_hyper_EMU}	
\centering
\begin{tabular}{cc}
	\toprule
	Category                   & $\delta$    \\ 
	\midrule
	easy/hard SMAC maps        & $1.3e^{-5}$ \\
	super hard SMAC maps       & $1.3e^{-3}$ \\
	GRF & $1.3e^{-3}$ \\ 
	\bottomrule
\end{tabular}
\end{table}

\subsection{Details of MARL tasks}\label{details_marl_tasks}
In this section, we specify the dimension of the state space, the action space, the episodic length, and the reward of SMAC \citep{samvelyan2019starcraft} and GRF \citep{kurach2020google}.

In SMAC, the global state of each task of SMAC includes the information of the coordinates of all agents, and features of both allied and enemy units. The action space of each agent consists of the moving actions and attacking enemies, and thus it increases according to the number of enemies. The dimensions of the state and action space and the episodic length vary according to the tasks as shown in Table \ref{tab:details of SMAC}. For reward structure, we used the \textit{shaped reward}, i.e., the default reward settings of SMAC, for all scenarios. The reward is given when dealing damage to the enemies and get bonuses for winning the scenario. The reward is scaled so that the maximum cumulative reward, $R_{max}$, that can be obtained from the episode, becomes around 20 \citep{samvelyan2019starcraft}.

\begin{table}[htbp]
\centering
\caption{Dimension of the state space and the action space and the episodic length of SMAC}
\vspace{0.3cm}
\label{tab:details of SMAC}
\begin{tabular}{cccc}
	\toprule
	Task & Dimension of state space & Dimension of action space & Episodic length\\
	\midrule 
	1c3s5z & 270 & 15 & 180\\
	3s5z & 216 & 14 & 150\\
	3s\_vs\_5z & 68 & 11 & 250\\
	5m\_vs\_6m & 98 & 12 & 70\\
	MMM2 & 322 & 18 & 180\\
	6h\_vs\_8z & 140 & 14 & 150\\
	3s5z\_vs\_3s6z & 230 & 15 & 170\\
	corridor & 282 & 30 & 400\\
	\bottomrule
\end{tabular}\label{tab:details of SMAC}
\end{table}

In GRF, the state of each task includes information on coordinates, ball possession, and the direction of players, etc. The dimension of the state space differs among the tasks as in Table \ref{tab:details of GRF}. The action of each agent consists of moving directions, different ways to kick the ball, sprinting, intercepting the ball and dribbling. The dimensions of the action spaces for each task are the same. Table \ref{tab:details of GRF} summarizes the dimension of the action space and the episodic length. In GRF, there can be two reward modes: one is "sparse reward" and the other is "dense reward." In sparse reward mode, agents get a positive reward +1 when scoring a goal and get -1 when conceding one to the opponents. In dense reward mode, agents can get positive rewards when they approach to opponent's goal, but the maximum cumulative reward is up to +1. In our experiments, we adopt sparse reward mode, and thus the maximum reward, $R_{max}$ is +1 for GRF.

\begin{table}[htbp]
\centering
\caption{Dimension of the state space and the action space and the episodic length of GRF}
\vspace{0.3cm}
\label{tab:details of GRF}
\begin{tabular}{cccc}
	\toprule
	Task & Dimension of state space & Dimension of action space & Episodic length\\
	\midrule
	3\_vs\_1WK & 26 & 19 & 150\\
	CA\_easy & 30 & 19 & 150\\
	CA\_hard & 34 & 19 & 150\\
	\bottomrule
\end{tabular}\label{tab:details of GRF}
\end{table} 

\subsection{Infrastructure}\label{Infrastructure}

Experiments for SMAC \citep{samvelyan2019starcraft} are mainly carried out on NVIDIA GeForce RTX 3090 GPU, and training for the longest experiment such as \texttt{corridor} via EMU (CDS) took less than 18 hours. Note that when training is conducted with $n_\textrm{circle}=2$, it takes more than one and a half times longer. Training encoder/decoder structure and updating $\gD_{E}$ with updated $f_{\phi}$ together only took less than 2 seconds at most in \texttt{corridor} task. As we update $f_{\phi}$ and $f_{\psi}$ periodically with $t_{emb}$, the additional time required for a trainable embedder is certainly negligible compared to MARL training.

\section{Further Experiment Results}\label{further_results}

\subsection{New Performance Index}\label{performance_index}

In this section, we present the details of a new performance index called \textit{overall win-rate}, ${\bar \mu _w}$. For example, let $f_{w}^{i}(s)$ be the test win-rate at training time $s$ of $i$th seed run and $\mu_{w}^{i}(t)$ represents the time integration of $f_{w}^{i}(s)$ until $t$. Then, a normalized overall win-rate, $\bar \mu_{w}$, can be expressed as 
\begin{equation}\label{Eq:overall_winrate}
{\bar \mu _w}(t) = \frac{1}{{{\mu _{\max }}}}\frac{1}{n}\sum\nolimits_{i = 1}^n {\mu _w^i} (t) = \frac{1}{{{\mu _{\max }}}}\frac{1}{n}\sum\nolimits_{i = 1}^n {\int_0^t {f_w^i(s)ds} },
\end{equation}
where $\mu _{\max }=t$ and $\bar \mu _w \in [0,1]$.\\

\begin{figure}[!h]
\begin{center}
	\begin{subfigure}{0.3\linewidth}
		\includegraphics[width=\linewidth]{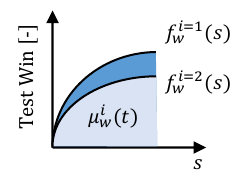}
	\end{subfigure}		
	\vspace{-0.1in}
\end{center}
\caption{Illustration of $\mu_{w}^{i}(t)$.}
\label{fig:new_performance_index}
\vspace{-0.1in}
\end{figure}

Figure \ref{fig:new_performance_index} illustrates the concept of time integration of win-rate, i.e., $\mu_{w}^{i}(t)$, to construct the overall win-rate, $\bar \mu_{w}$. By considering the integration of win-rate of each seed case, the performance variance can be considered, and thus $\bar \mu_{w}$ shows the training efficiency (speed) as well as the training quality (win-rate).

\subsection{Additional Experiment Results}\label{additional_results}
In Section \ref{parametric_ablation_study}, we present the summary of parametric studies on $\delta$ with respect to various choices of $f_{\phi}$. To see the training efficiency and performance at the same time, Table \ref{Param_study} and \ref{Param_study2} present the overall win-rate $\bar \mu_{w}$ according to training time. We conduct the experiments for 5 different seed cases and at each test phase 32 samples were used to evaluate the win-rate $[\%]$. Note that we discard the component of episodic incentive $r^p$ to see the performance variations according to $\delta$ and types of $f_{\phi}$ more clearly.

\begin{table}[!htbp]
\caption{$\bar \mu_{w}$ according to $\delta$ and design choice of embedding function on {\bf hard} SMAC map,  \texttt{3s\_vs\_5z}.}
\vskip -0.05in 
\label{Param_study}
\adjustbox{max width=\textwidth}{
	\begin{tabular}{c ccc@{\hspace{7mm}}ccc@{\hspace{7mm}}ccc}
		\toprule
		\begin{tabular}[c]{@{}c@{}}Training time \\ {[}mil{]}\end{tabular} & \multicolumn{3}{c}{0.69}        & \multicolumn{3}{c}{1.37}        & \multicolumn{3}{c}{2.00}         \\ 
		\midrule
		$\delta$                                                           & random & EmbNet & dCAE           & random & EmbNet & dCAE           & random & EmbNet & dCAE           \\ 
		\midrule
		1.3e-7                                                             & 0.033  & 0.051  & \textbf{0.075} & 0.245  & 0.279  & \textbf{0.343} & 0.413  & 0.443  & \textbf{0.514} \\
		1.3e-5                                                             & 0.010  & 0.044  & \textbf{0.063} & 0.171  & 0.270  & \textbf{0.325} & 0.320  & 0.441  & \textbf{0.491} \\
		1.3e-3                                                             & 0.034  & 0.043  & \textbf{0.078} & 0.226  & 0.270  & \textbf{0.357} & 0.381  & 0.439  & \textbf{0.525} \\
		1.3e-2                                                             & 0.019  & 0.005  & \textbf{0.079} & 0.205  & 0.059  & \textbf{0.346} & 0.348  & 0.101  & \textbf{0.518} \\ 
		\bottomrule
	\end{tabular}
}
\vspace{-0.05in}
\end{table}
\begin{table}[!htbp]
\caption{$\bar \mu_{w}$ according to $\delta$ and design choice of embedding function on {\bf hard} SMAC map,  \texttt{5m\_vs\_6m}.}
\label{Param_study2}
\vskip -0.05in
\adjustbox{max width=\textwidth}{
	\begin{tabular}{c ccc@{\hspace{7mm}}ccc@{\hspace{7mm}}ccc}
		\toprule
		\begin{tabular}[c]{@{}c@{}}Training time \\ {[}mil{]}\end{tabular} & \multicolumn{3}{c}{0.69}  & \multicolumn{3}{c}{1.37} & \multicolumn{3}{c}{2.00} \\ 
		\midrule
		$\delta$                                                           & random & EmbNet & dCAE           & random & EmbNet & dCAE           & random & EmbNet & dCAE           \\ 
		\midrule
		1.3e-7                                                             & 0.040                      & \textbf{0.117}             & 0.110                     & 0.287                      & 0.397                      & \textbf{0.397}            & 0.577                      & 0.690                      & \textbf{0.701}           \\
		1.3e-5                                                             & 0.064                      & 0.107                      & \textbf{0.131}            & 0.334                      & 0.402                      & \textbf{0.436}            & 0.634                      & 0.714                      & \textbf{0.749}           \\
		1.3e-3                                                             & 0.040                      & \textbf{0.080}             & 0.064                     & 0.333                      & \textbf{0.377}             & 0.363                     & 0.646                      & \textbf{0.687}             & 0.677                    \\
		1.3e-2                                                             & 0.038                      & 0.000                      & \textbf{0.048}            & 0.288                      & 0.001                      & \textbf{0.332}            & 0.584                      & 0.001                      & \textbf{0.643}           \\ 
		\bottomrule
	\end{tabular}
}
\vspace{-0.1in}
\end{table}

\begin{figure}[!htbp]
\begin{center}
	\begin{subfigure}{0.25\linewidth}
		\includegraphics[width=\linewidth]{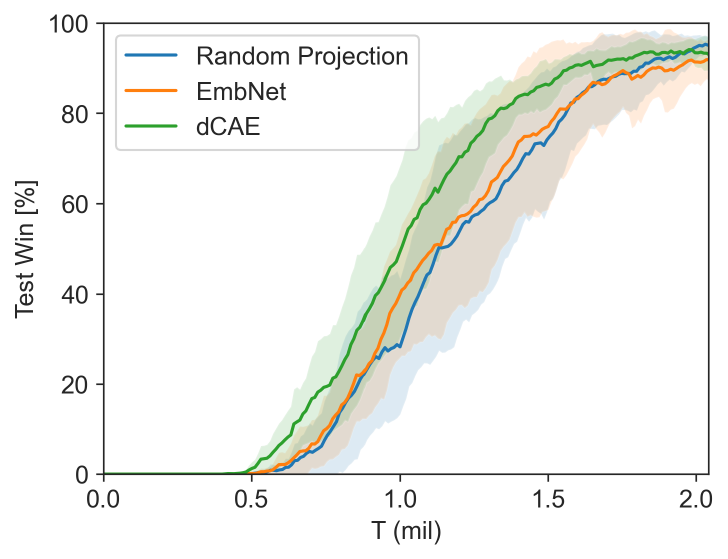}
		\caption{$\delta=1.3e^{-7}$}
	\end{subfigure}\hfill
	\begin{subfigure}{0.25\linewidth}
		\includegraphics[width=\linewidth]{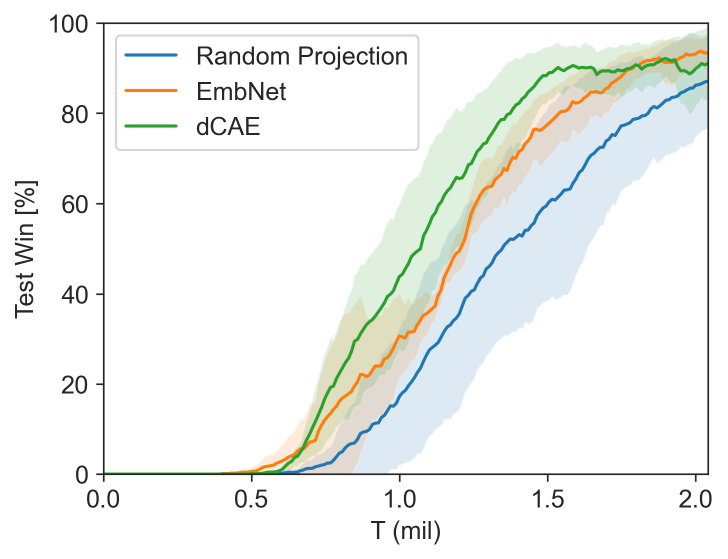}
		\caption{$\delta=1.3e^{-5}$}
	\end{subfigure}\hfill 
	\begin{subfigure}{0.25\linewidth}
		\includegraphics[width=\linewidth]{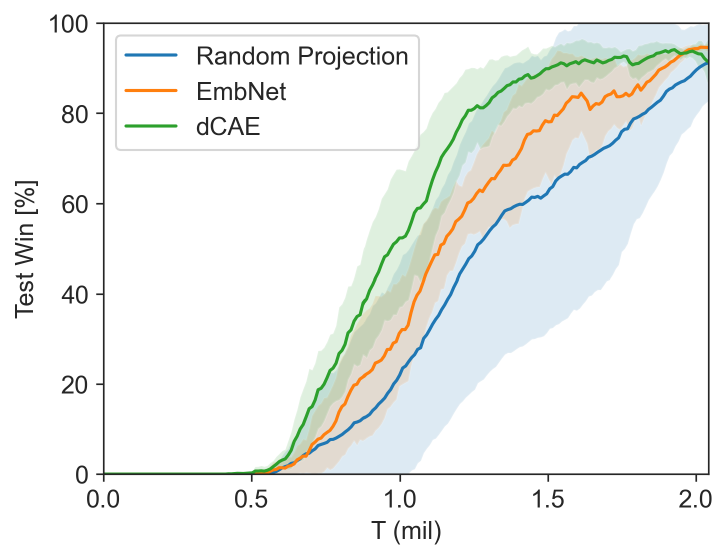}
		\caption{$\delta=1.3e^{-3}$} 
	\end{subfigure}\hfill
	\begin{subfigure}{0.25\linewidth}
		\includegraphics[width=\linewidth]{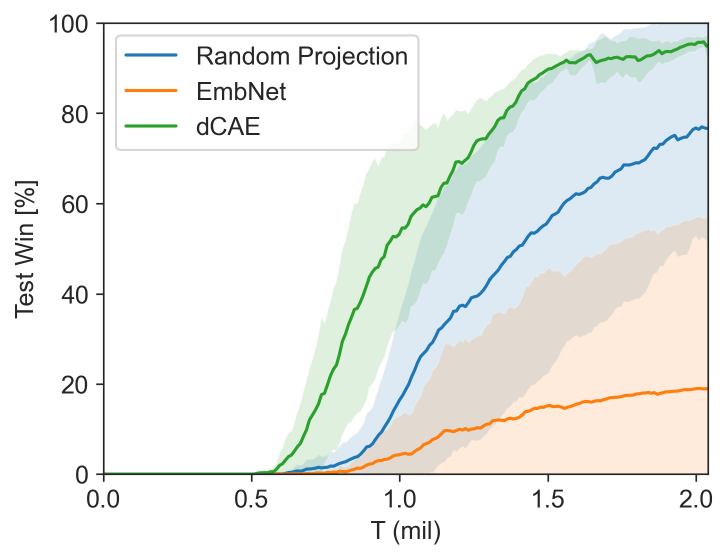}
		\caption{$\delta=1.3e^{-2}$}
	\end{subfigure}
\end{center}
\caption{Parametric studies for $\delta$ on \texttt{3s\_vs\_5z} SMAC map.}
\label{fig:delta_param_3s_vs_5z}
\vspace{-0.15in}
\end{figure}

\begin{figure}[!htbp]
\begin{center}
	\begin{subfigure}{0.25\linewidth}
		\includegraphics[width=\linewidth]{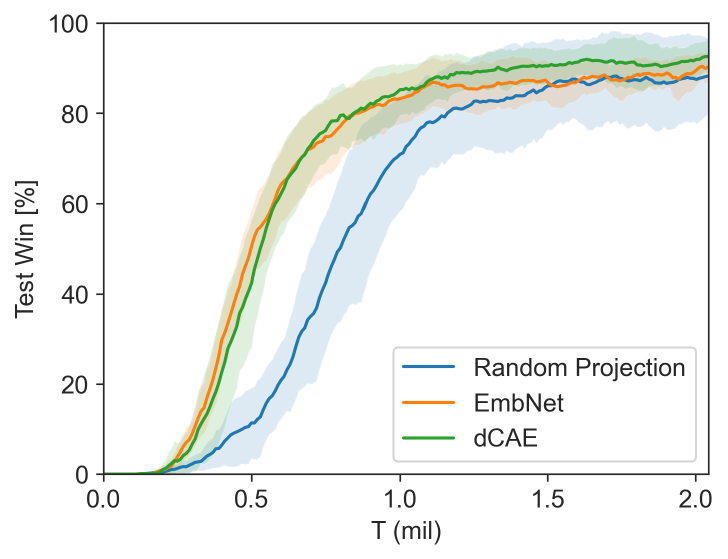}
		\caption{$\delta=1.3e^{-7}$}
	\end{subfigure}\hfill
	\begin{subfigure}{0.25\linewidth}
		\includegraphics[width=\linewidth]{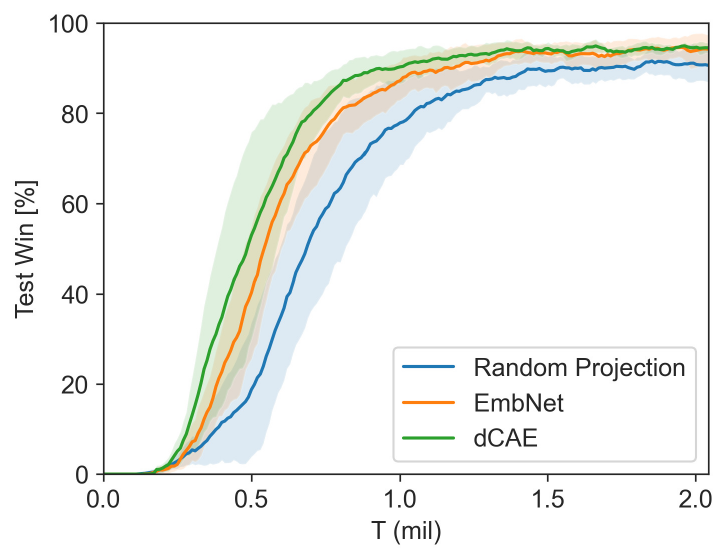}
		\caption{$\delta=1.3e^{-5}$}
	\end{subfigure}\hfill
	\begin{subfigure}{0.25\linewidth}
		\includegraphics[width=\linewidth]{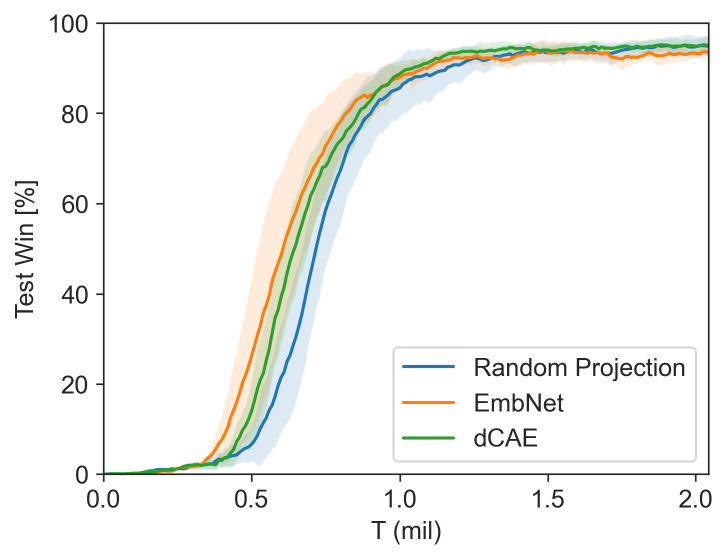}
		\caption{$\delta=1.3e^{-3}$} 
	\end{subfigure}\hfill
	\begin{subfigure}{0.25\linewidth}
		\includegraphics[width=\linewidth]{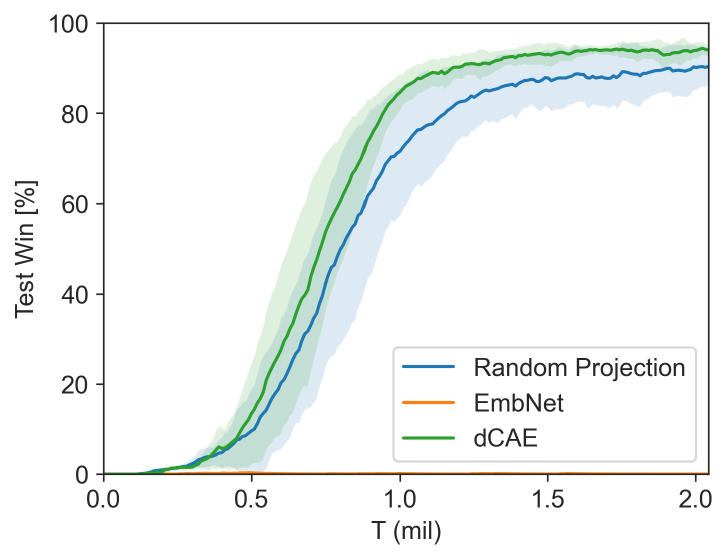}
		\caption{$\delta=1.3e^{-2}$}
	\end{subfigure}
\end{center}
\caption{Parametric studies for $\delta$ on \texttt{5m\_vs\_6m} SMAC map.}
\label{fig:delta_param_5m_vs_6m}
\vspace{-0.15in}
\end{figure}

As Table \ref{Param_study} and \ref{Param_study2} illustrate that dCAE structure for $f_{\phi}$, which considers the reconstruction loss of global state $s$ as in Eq. \ref{Eq:Loss_embedder_dCAE}, shows the best training efficiency in most cases. For \texttt{5m\_vs\_6m} task with $\delta=1.3e^{-3}$, EmbNet achieves the highest value among $f_{\phi}$ choices in terms of $\bar \mu_{w}$ but fails to find optimal policy at $\delta=1.3e^{-2}$ unlike other design choices. This implies that the reconstruction loss of dCAE facilitates the construction of a smoother embedding space for $\gD_{E}$, enabling the retrieval of memories within a broader range of $\delta$ values from the key state.	
\begin{wrapfigure}{!hR}{0.5\linewidth} 
\vspace{-0.2in}
\begin{center}
	\begin{minipage}{\linewidth} 
	\includegraphics[width=1.0\linewidth]{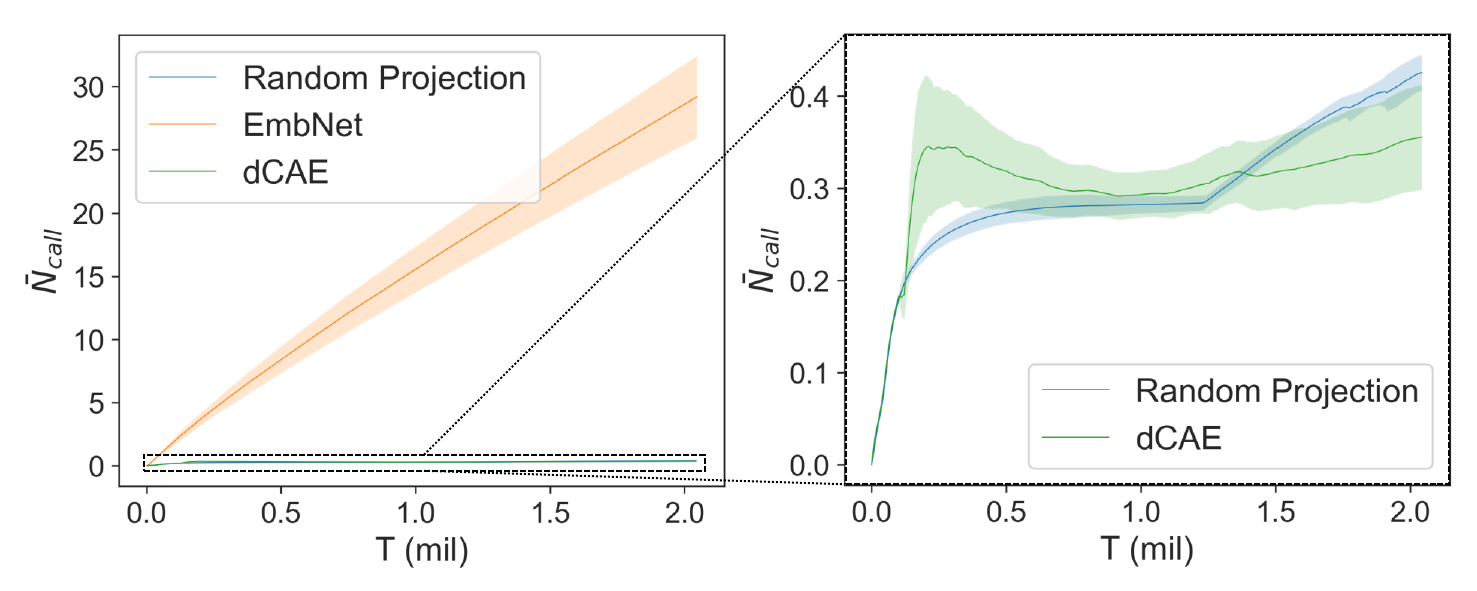} 
	\end{minipage}
\end{center}
\caption{$\bar N_{call}$ of all memories in $\gD_{E}$ when $\delta=0.013$ according to design choice for $f_{\phi}$.}
\label{fig:recall_freq}
\vspace{-0.15in}
\end{wrapfigure}
Figure \ref{fig:delta_param_3s_vs_5z} and \ref{fig:delta_param_5m_vs_6m} show the corresponding learning curves of each encoder structure for different $\delta$ values. A large $\delta$ value results in a higher performance variance than the cases with smaller $\delta$, according to different seed cases. 

This is because a high value of $\delta$ encourages exploratory memory recall. In other words, by adjusting $\delta$, we can control the level of exploration since it controls whether to recall its own MC return or the highest value of other similar states within $\delta$.  
Thus, without constructing smoother embedding space as in dCAE, learning with exploratory memory recall within large $\delta$ can converge to sub-optimality as illustrated by the case of EmbNet in Figure \ref{fig:delta_param_5m_vs_6m}(d). 

In Figure \ref{fig:recall_freq} which shows the averaged number of memory recall ($\bar N_{call}$) of all memories in $\gD_{E}$, $\bar N_{call}$ of EmbNet significantly increases as training proceeds. On the other hand, dCAE was able to prevent this problem and recalled the proper memories in the early learning phase, achieving good training efficiency. Thus, embedding with dCAE can leverage a wide area of memory in $\gD_{E}$ and becomes robust to hyperparameter $\delta$.


\newpage
\subsection{Comparative Evaluation on Additional Starcraft II Maps}\label{additional_results_smac2}
\begin{figure*}[!h]
\begin{center}
	\centerline{\includegraphics[width=0.95\linewidth]{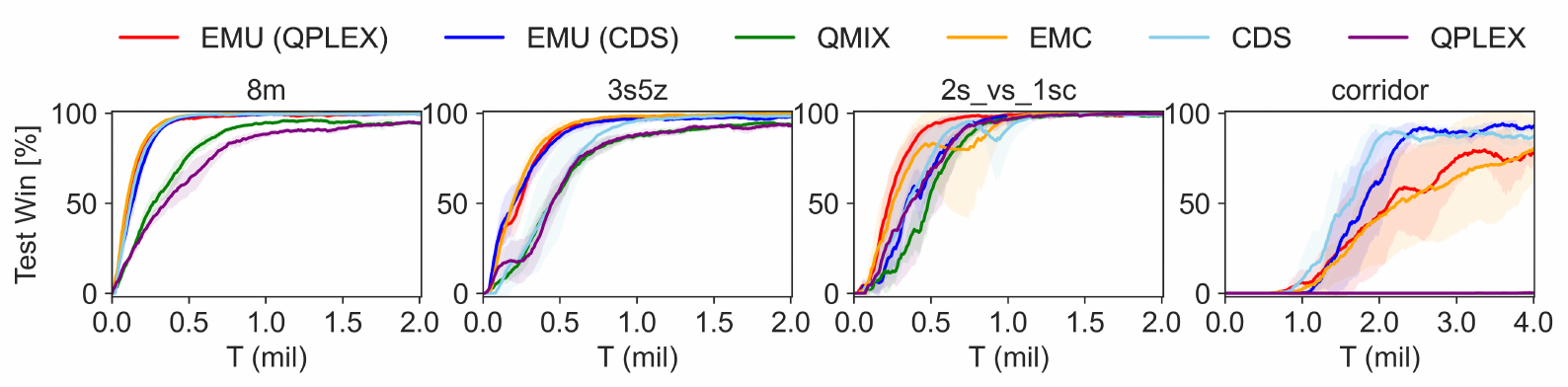}}
\end{center}
\vskip -0.2in
\caption{Performance comparison of EMU against baseline algorithms on additional SMAC maps.}
\label{fig:smac_performance_2}
\vskip -0.2in
\end{figure*}

Figure \ref{fig:smac_performance_2} presents a comparative evaluation of EMU with baseline algorithms on additional SMAC maps. Adopting EMU shows performance gain in various tasks.

\subsection{Comparison of EMU with MAPPO on SMAC}
\begin{figure}[!h]
\vspace{-0.0in}
\begin{center}			
	\centerline{\includegraphics[width=0.9\linewidth]{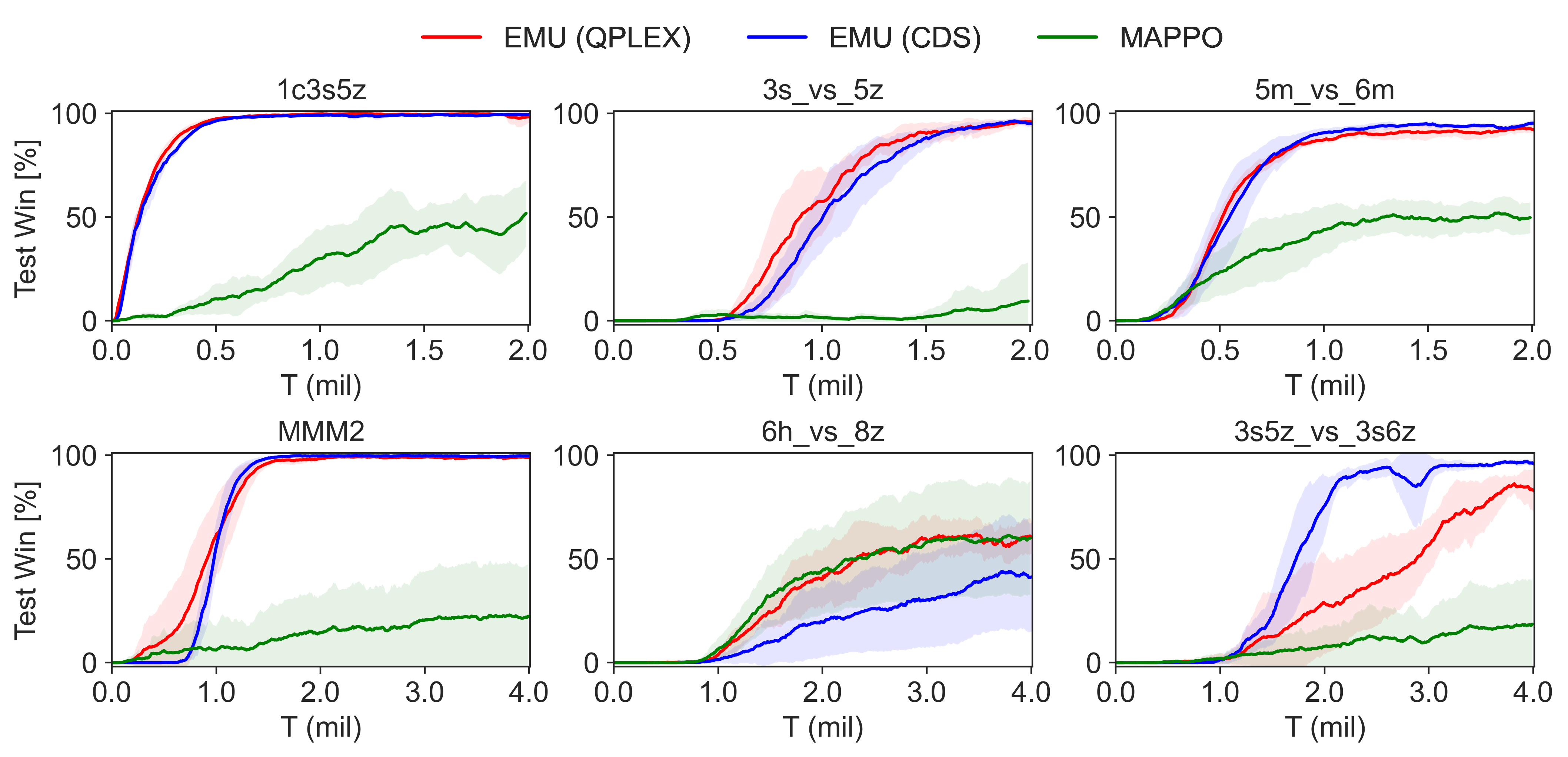}}
\end{center}
\vspace{-0.25in}
\caption{Performance comparison with MAPPO on selected SMAC maps.}
\label{fig:performance_comp_mappo}
\vspace{-0.10in}
\end{figure}

In this subsection, we compare the EMU with MAPPO \citep{yu2022surprising} on selected SMAC maps. Figure \ref{fig:performance_comp_mappo} shows the performance in six SMAC maps: \texttt{1c3s5z}, \texttt{3s\_vs\_5z}, \texttt{5m\_vs\_6m}, \texttt{MMM2}, \texttt{6h\_vs\_8z} and \texttt{3s5z\_vs\_3s6z}. Similar to the previous performance evaluation in Figure \ref{fig:smac_performance}, Win-rate is computed with 160 samples: 32 episodes for each training random seed and 5 different random seeds. Also, for MAPPO, scenario-dependent hyperparameters are adopted from their original settings in the uploaded source code.

From Figure \ref{fig:performance_comp_mappo}, we can see that EMU performs better than MAPPO with an evident gap. Although after extensive training MAPPO showed a comparable performance against off-policy algorithm in its original paper \citep{yu2022surprising}, within the same training timestep used for our experiments, we found that MAPPO suffers from local convergence in super hard SMAC tasks such as \texttt{MMM2} and \texttt{3s5z\_vs\_3s6z} as shown in Figure \ref{fig:performance_comp_mappo}. Only in \texttt{6h\_vs\_8z}, MAPPO shows comparable performance to EMU (QPLEX) with higher performance variance across different seeds.

\newpage 
\subsection{Additional Parametric Study}\label{additional_param}
In this subsection, we conduct an additional parametric study to see the effect of key hyperparameter $\delta$. Unlike the previous parametric study on Appendix \ref{additional_results}, we adopt both dCAE embedding network for $f_{\phi}$ and episodic reward. For evaluation, we consider three GRF tasks such as \texttt{academy\_3\_vs\_1\_with\_keeper} (\texttt{3\_vs\_1WK}), \texttt{academy\_counterattack\_easy} (\texttt{CA-easy}), and \texttt{academy\_counterattack\_hard} (\texttt{CA-hard}); and one {\bf super hard} SMAC map such as \texttt{6h\_vs\_8z}. For each task to evaluate EMU, four $\delta$ values, such as $\delta_1=1.3e^{-7}$, $\delta_2=1.3e^{-5}$, $\delta_3=1.3e^{-3}$, and $\delta_4=1.3e^{-2}$, are considred. Here, to compute the win-rate, 160 samples (32 episodes for each training random seed and 5 different random seeds) are used for \texttt{3\_vs\_1WK} and \texttt{6h\_vs\_8z} while 100 samples (20 episodes for each training random seed and 5 different random seeds) are used for \texttt{CA-easy} and \texttt{CA-hard}. Note that CDS and EMU (CDS) utilize the same hyperparameters, and EMC and EMU (QPLEX) use the same hyperparameters without a curiosity incentive presented in \citet{zheng2021episodic} as the model without it showed the better performance when utilizing episodic control.

\begin{figure}[!htbp]
\begin{center}
	\begin{subfigure}{0.25\linewidth}
		\includegraphics[width=\linewidth]{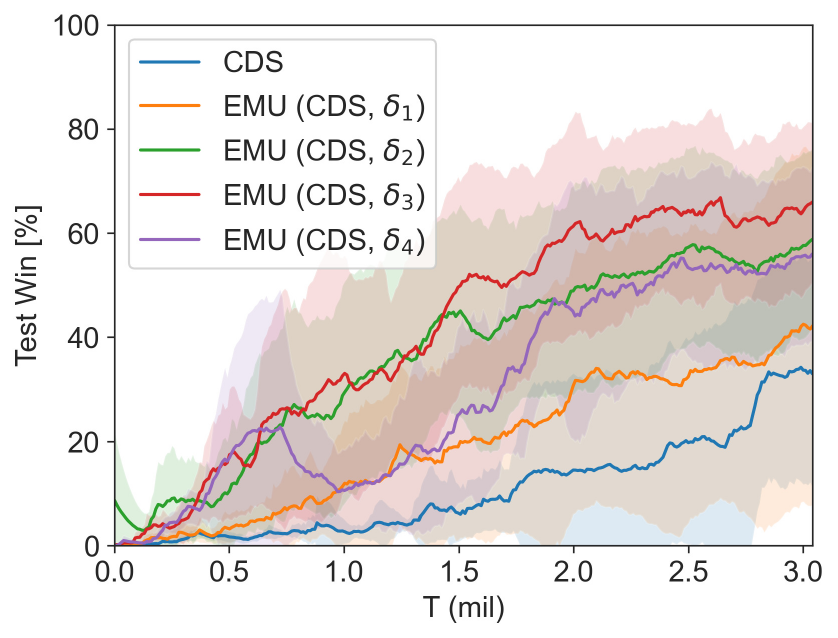}
		\caption{\texttt{3\_vs\_1WK} (GRF)}
	\end{subfigure}\hfill
	\begin{subfigure}{0.25\linewidth}
		\includegraphics[width=\linewidth]{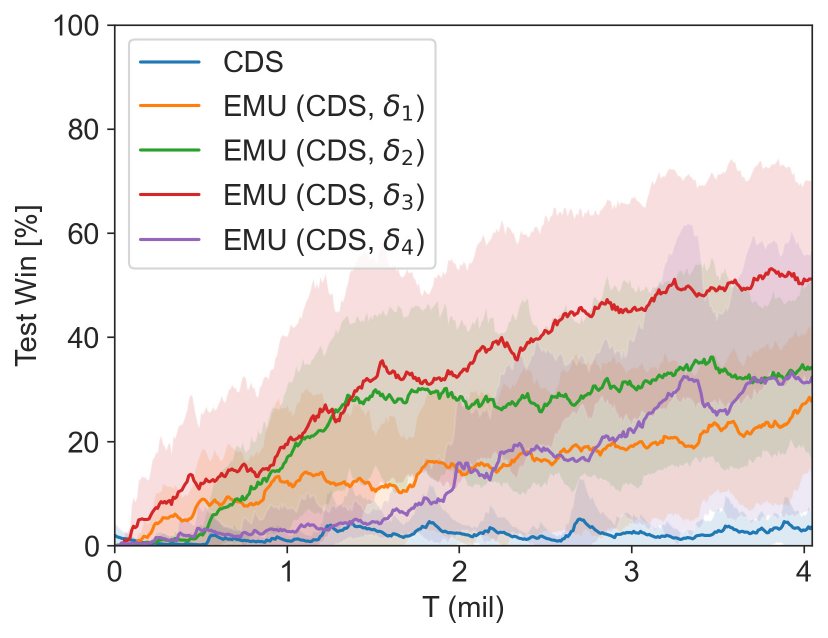}
		\caption{\texttt{CA-easy} (GRF)}
	\end{subfigure}\hfill 
	\begin{subfigure}{0.25\linewidth}
		\includegraphics[width=\linewidth]{GRF_param_CA_hard-eps-converted-to.pdf}
		\caption{\texttt{CA-hard} (GRF)}
	\end{subfigure}\hfill
	\begin{subfigure}{0.25\linewidth}
		\includegraphics[width=\linewidth]{smac_param_6h_vs_8z-eps-converted-to.pdf}
		\caption{\texttt{6h\_vs\_8z} (SMAC)}
	\end{subfigure}
\end{center}
\caption{Parametric studies for $\delta$ on various GRF maps and {\bf super hard} SMAC map.}
\label{fig:delta_param_final}
\end{figure} 

In all cases, EMU with $\delta_3=1.3e^{-3}$ shows the best performance. The tasks considered here are all complex multi-agent tasks, and thus adopting a proper value of $\delta$ benefits the overall performance and achieves the balance between exploration and exploitation by recalling the semantically similar memories from episodic memory. The optimal value of $\delta_3$ is consistent with the determination logic on $\delta$ in a memory efficient way presented in Appendix \ref{memory_utilization}.

\subsection{Additional Parametric Study on $\lambda_{rcon}$}\label{parametric_lambda_rcon}
Additionally, we conduct a parametric study for $\lambda_{rcon}$ in Eq. \ref{Eq:Loss_embedder_dCAE}. For each task, EMU with five $\lambda_{rcon}$ values, such as $\lambda_{rcon,0}=0.01$, $\lambda_{rcon,1}=0.1$, $\lambda_{rcon,2}=0.5$, $\lambda_{rcon,3}=1.0$ and $\lambda_{rcon,4}=10$, are evaluated. Here, to compute the win-rate of each case, 160 samples (32 episodes for each training random seed and 5 different random seeds) are used.
\begin{figure}[!hbtp] 
\begin{center}
	\centering	
	\begin{minipage}[!ht]{0.65\linewidth}
		\subfloat[\centering \texttt{3s5z}
		(SMAC)]{{\includegraphics[width=0.45\linewidth]{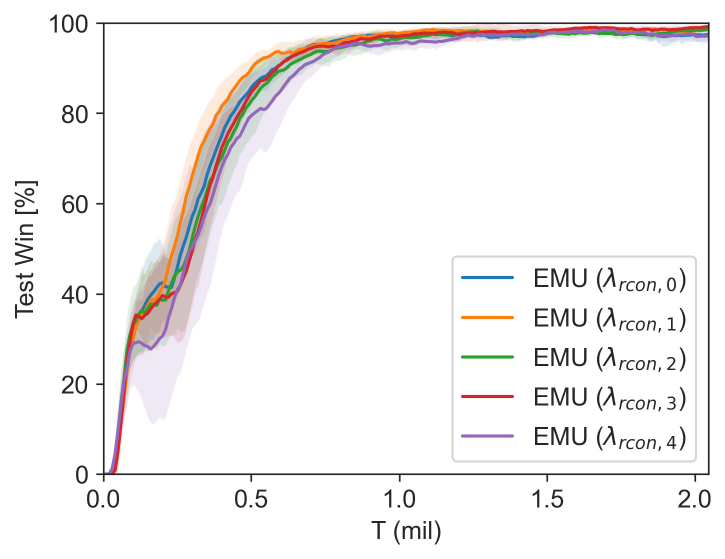} }}%
		\subfloat[\centering \texttt{3s\_vs\_5z} (SMAC)]{{\includegraphics[width=0.45\linewidth]{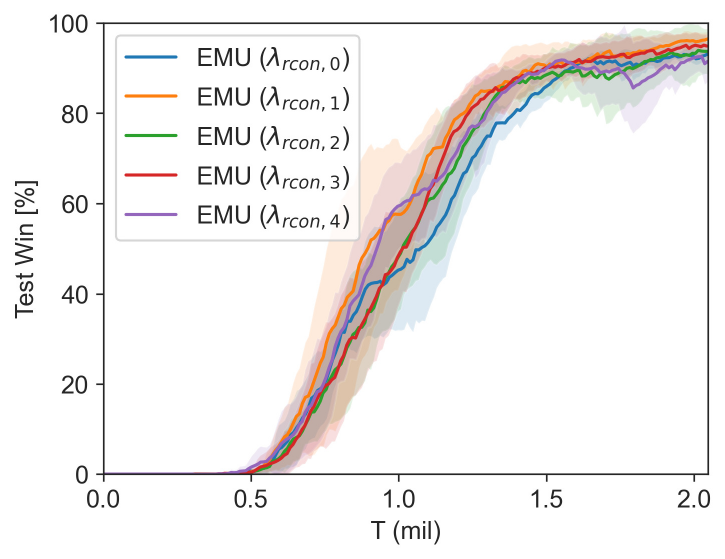} }}%
		\caption{Parametric study for $\lambda_{rcon}$.}%
		\label{fig:performance_parametric_lambda_rcon}%
		\vspace{-0.2in}
	\end{minipage}
\end{center}
\end{figure}
From Figure \ref{fig:performance_parametric_lambda_rcon}, we can see that broad range of $\lambda_{rcon} \in \bigl\{ 0.1, 0.5, 1.0 \bigr\}$ work well in general. However, with large $\lambda_{rcon}$ as $\lambda_{rcon,4}=10$, we can observe that some performance degradation at the early learning phase in \texttt{3s5z} task. This result is in line with the learning trends of Case 1 and Case 2 of \texttt{3s5z} in Figure \ref{fig:performance_embedding_loss}, which do not consider prediction loss and only take into account the reconstruction loss. 
Thus, considering both prediction loss and reconstruction loss as Case 4 in Eq. \ref{Eq:Loss_embedder_dCAE} with proper $\lambda_{rcon}$ is essential to optimize the overall learning performance.

\subsection{Additional Ablation Study in GRF}\label{additional_ablation}
\begin{figure}[!hbtp] 
\begin{center}
	\centering	
	\begin{minipage}[!hbtp]{0.7\linewidth}
		\subfloat[\centering \texttt{3\_vs\_1WK} (GRF)]{{\includegraphics[width=0.45\linewidth]{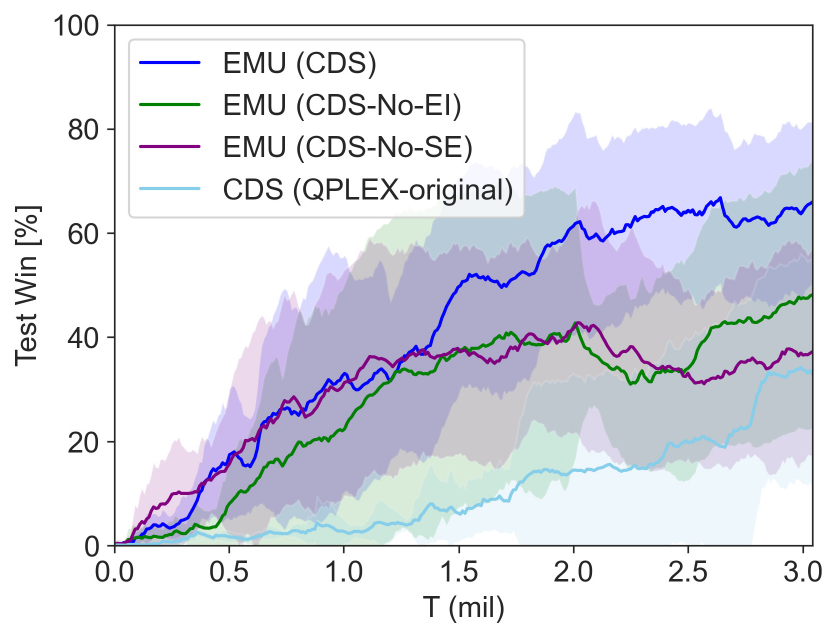} }}%
		\subfloat[\centering \texttt{CA-easy} (GRF) ]{{\includegraphics[width=0.45\linewidth]{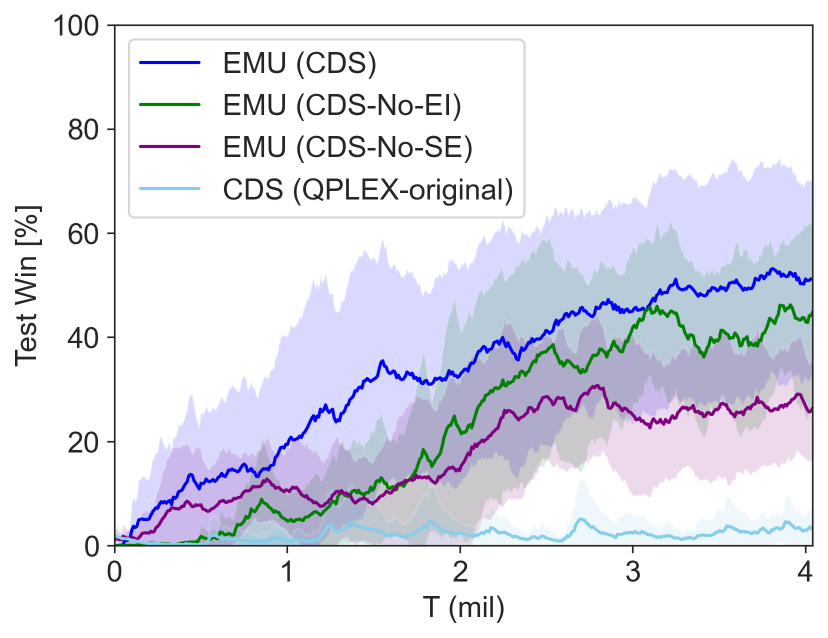} }}%
		\caption{Ablation studies on episodic incentive on GRF tasks.}%
		\label{fig:ablation_grf_example}%
		\vspace{-0.2in}
	\end{minipage}
\end{center}
\end{figure}

In this subsection, we conduct additional ablation studies via GRF tasks to see the effect of episodic incentive. Again, EMU (CDS-No-EI) ablates episodic incentive from EMU (CDS) and utilizes the conventional episodic control presented in Eq. \ref{Eq:Episodic_loss} instead. Again, EMU (CDS-No-SE) ablates semantic embedding by dCAE and adopts random projection with episodic incentive $r^p$. In both tasks, utilizing episodic memory with the proposed embedding function improves the overall performance compared to the original CDS algorithm. By adopting episodic incentives instead of conventional episodic control, EMU (CDS) achieves better learning efficiency and rapidly converges to optimal policies compared to EMU (CDS-No-EI).

\subsection{Additional Ablation Study on Embedding Loss}\label{embedding_loss_ablation}
In our case, the autoencoder uses the reconstruction loss to enforce the embedded representation $x$ to contain the full information of the original feature, $s$. 
We are adding $(H_t-f^H_\psi(f_\phi(s_t|t)|t))^2$ to guide the embedded representation to be consistent to $H_t$, as well, which works as a regularizer to the autoencoder. 
Therefore, $f_\psi^H$ is used in Eq. \ref{Eq:Loss_embedder_dCAE} to predict the observed $H_t$ from $D_E$ as a part of the semantic regularization effort. 

Because $H_t$ is different from $f_\psi^H(x_t)$, the effort of minimizing their difference becomes the regularizer creating a gradient signal to learn $\psi$ and $\phi$. The update of $\phi$ results in the updated $x$ influenced by the regularization. Note that we update $\phi$ through the backpropagation of $\psi$.

The case of $L(\phi,\psi)=||s_t-f_{\psi}^s(f_{\phi}(s_t|t)|t)||_2^2$ occurs when $\lambda_{rcon}$ becomes relatively much higher than $1$, which makes $(H_t-f_{\psi}^H(f_{\phi}(s_t|t)|t))^2$ becomes ineffective. In other words, when $\lambda_{rcon}$ in Eq. \ref{Eq:Loss_embedder_dCAE} becomes relatively much higher than $1$, $(H_t-f_{\psi}^H(f_{\phi}(s_t|t)|t))^2$ becomes ineffective.

\begin{figure}[!hbt]
\centering
\begin{subfigure}{0.25\textwidth}
	\includegraphics[width=\linewidth]{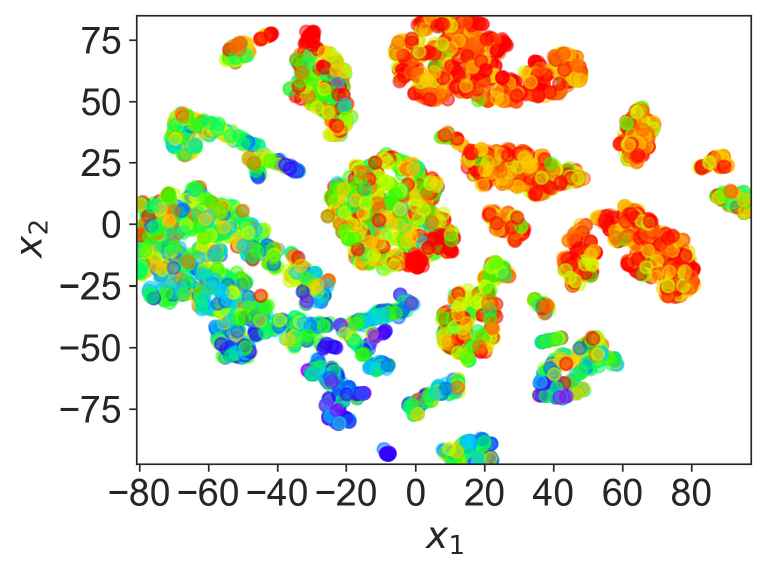}
	\caption{Loss (case 1)}
\end{subfigure}\hfill
\begin{subfigure}{0.25\textwidth}
	\includegraphics[width=\linewidth]{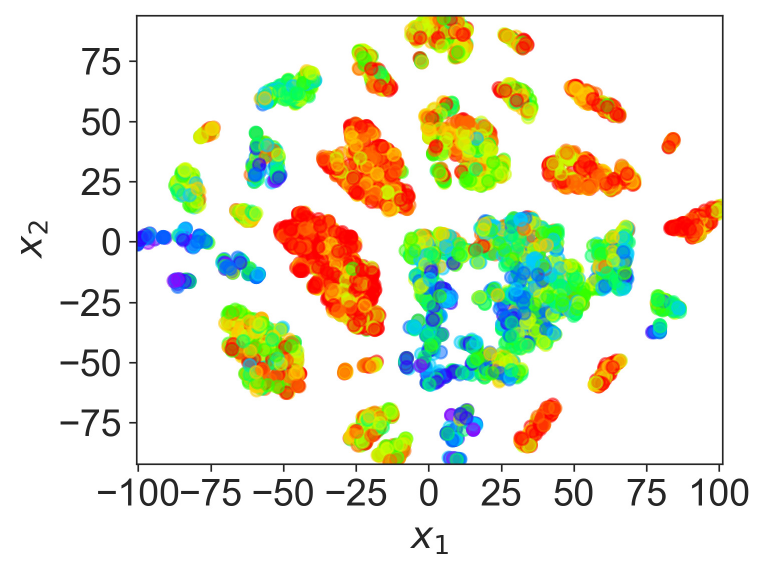}
	\caption{Loss (case 2)}
\end{subfigure}\hfill 
\begin{subfigure}{0.25\textwidth}
	\includegraphics[width=\linewidth]{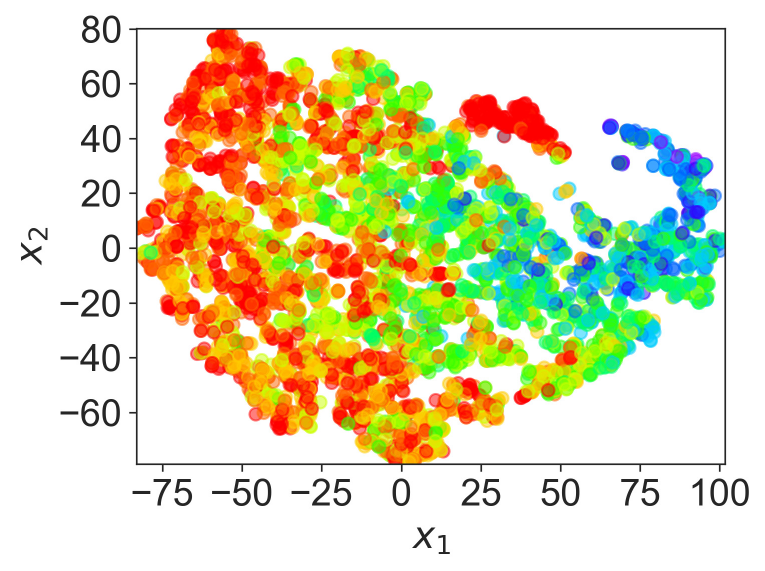}
	\caption{Loss (case 3)}
\end{subfigure}\hfill
\begin{subfigure}{0.25\textwidth}
	\includegraphics[width=\linewidth]{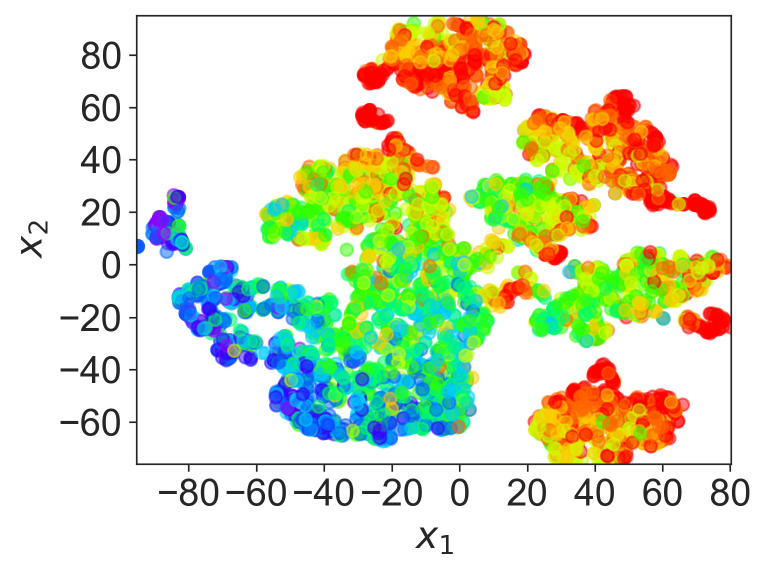}
	\caption{Loss (case 4)}
\end{subfigure}
\caption{t-SNE of sampled embedding $x \in D_E$ trained by dCAE with various loss functions in \texttt{3s\_vs\_5z} SMAC map. Colors from red to purple represent from low return to high return.}
\label{fig:t-SNE_loss}
\end{figure}

\begin{figure}[!hbtp] 
\begin{center}
	\centering	
	\begin{minipage}[!h]{0.60\linewidth}
		\subfloat[\centering \texttt{3s5z}
		(SMAC)]{{\includegraphics[width=0.45\linewidth]{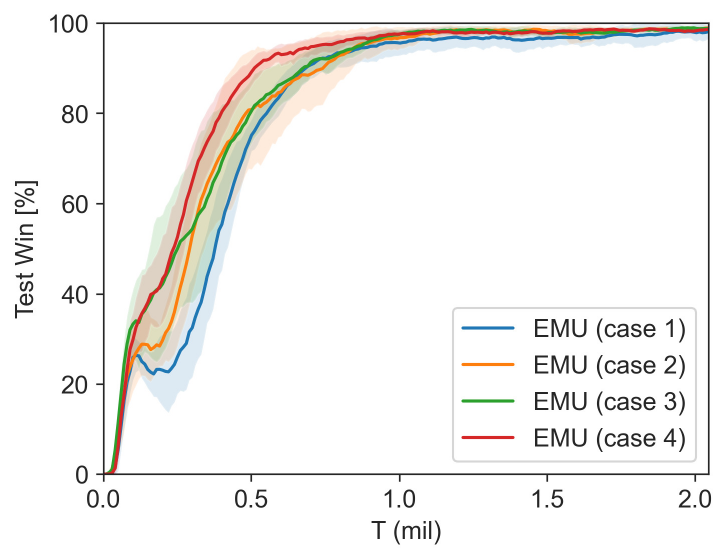} }}%
		\subfloat[\centering \texttt{3s\_vs\_5z} (SMAC)]{{\includegraphics[width=0.45\linewidth]{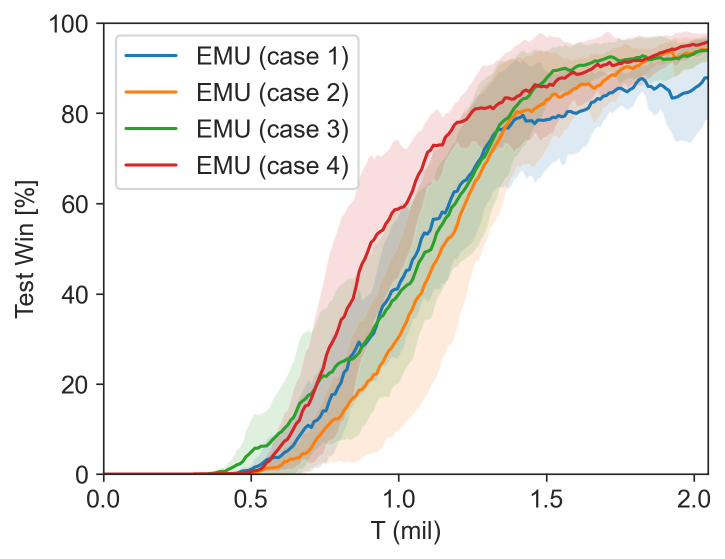} }}%
		\caption{Performance comparison of various loss functions for dCAE.}%
		\label{fig:performance_embedding_loss}%
		\vspace{-0.2in}
	\end{minipage}
\end{center}
\end{figure}

The case of $L(\phi,\psi)=(H_t-f_{\psi}^H(f_{\phi}(s_t|t)|t))^2$ occurs when the scale factor $\lambda_{rcon}$ becomes relatively much smaller than $1$, which makes $(H_t-f_{\psi}^H(f_{\phi}(s_t|t)|t))^2$ become a dominant factor. We conduct ablation studies considering four cases as follows:    
\begin{itemize}
\item \textbf{Case 1:} $L(\phi,\psi)=||s_t-f_\psi^s(f_\phi(s_t))||_2^2$, presented in Figure \ref{fig:t-SNE_loss}(a)	
\item \textbf{Case 2:} $L(\phi,\psi)=||s_t-f_\psi^s(f_\phi(s_t|t)|t)||_2^2$, presented in Figure \ref{fig:t-SNE_loss}(b)	
\item \textbf{Case 3:} $L(\phi,\psi)=(H_t-f^H_\psi(f_\phi(s_t|t)|t))^2$, presented in Figure \ref{fig:t-SNE_loss}(c)	
\item \textbf{Case 4:} $L(\phi,\psi)=( H_t-f_{\psi}^H(f_{\phi}(s_t|t)|t))^2 + \lambda_{rcon} ||s_t-f_{\psi}^s(f_{\phi}(s_t|t)|t)||_2^2$, i.e., Eq. \ref{Eq:Loss_embedder_dCAE}, presented in Figure \ref{fig:t-SNE_loss}(d)	
\end{itemize}
We visualize the result of t-SNE of 50K samples $x \in D_E$ out of 1M memory data trained by various loss functions: The task was 3s\_vs\_5z of SMAC as in Figure \ref{fig:t-SNE} and the training for all models proceeds for 1.5mil training steps. 
Case 1 and Case 2 showed irregular return distribution across the embedding space. In those two cases, there was no consistent pattern in the reward distribution. Case 3 with only return prediction in the loss showed better patterns compared to Case 1 and 2 but some features are not clustered well. We suspect that the consistent state representation also contributes to the return prediction. Case 4 of our suggested loss showed the most regular pattern in the return distribution arranging the low-return states as a cluster and the states with desirable returns as another cluster. In Figure \ref{fig:performance_embedding_loss}, Case 4 shows the best performance in terms of both learning efficiency and terminal win-rate.

\newpage

\subsection{Additional Ablation Study on Semantic Embedding}
\label{additional_SE_ablation}
To further understand the role of semantic embedding, we conduct additional ablation studies and present them with the general performance of other baseline methods. Again, EMU (CDS-No-SE) ablates semantic embedding by dCAE and adopts random projection instead, along with episodic incentive $r^p$. 

\begin{figure}[!h]
\vskip -0.1in
\begin{center}
	\centerline{\includegraphics[width=0.9\linewidth]{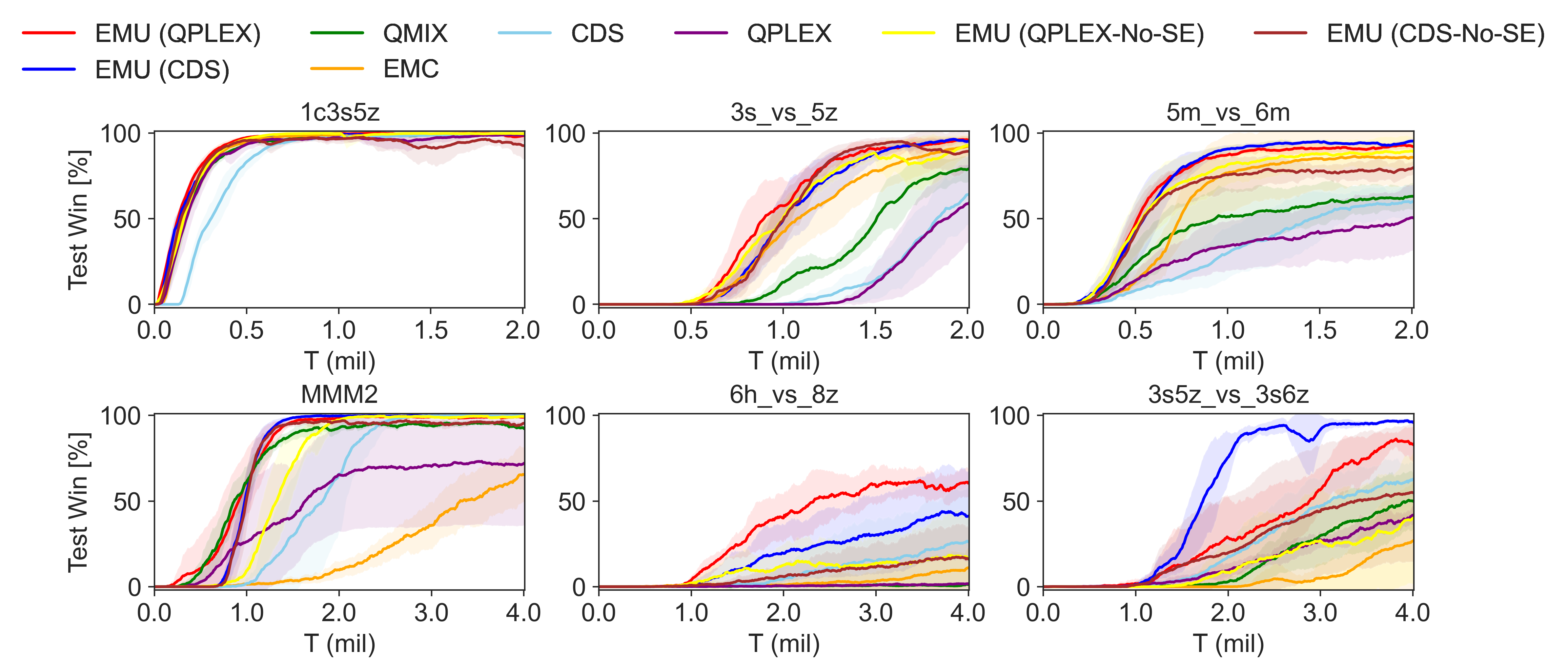}}
	\vskip -0.3in
\end{center}
\caption{Performance comparison of EMU against baseline algorithms on three {\bf easy and hard} SMAC maps: \texttt{1c3s5z}, \texttt{3s\_vs\_5z}, and \texttt{5m\_vs\_6m}, and three {\bf super hard} SMAC maps: \texttt{MMM2}, \texttt{6h\_vs\_8z}, and \texttt{3s5z\_vs\_3s6z}.}
\label{fig:smac_performance_with_ablation}
\vskip -0.1in
\end{figure}	

\begin{figure}[!h]
\vspace{-0.0in}
\begin{center}			
	\centerline{\includegraphics[width=0.7\linewidth]{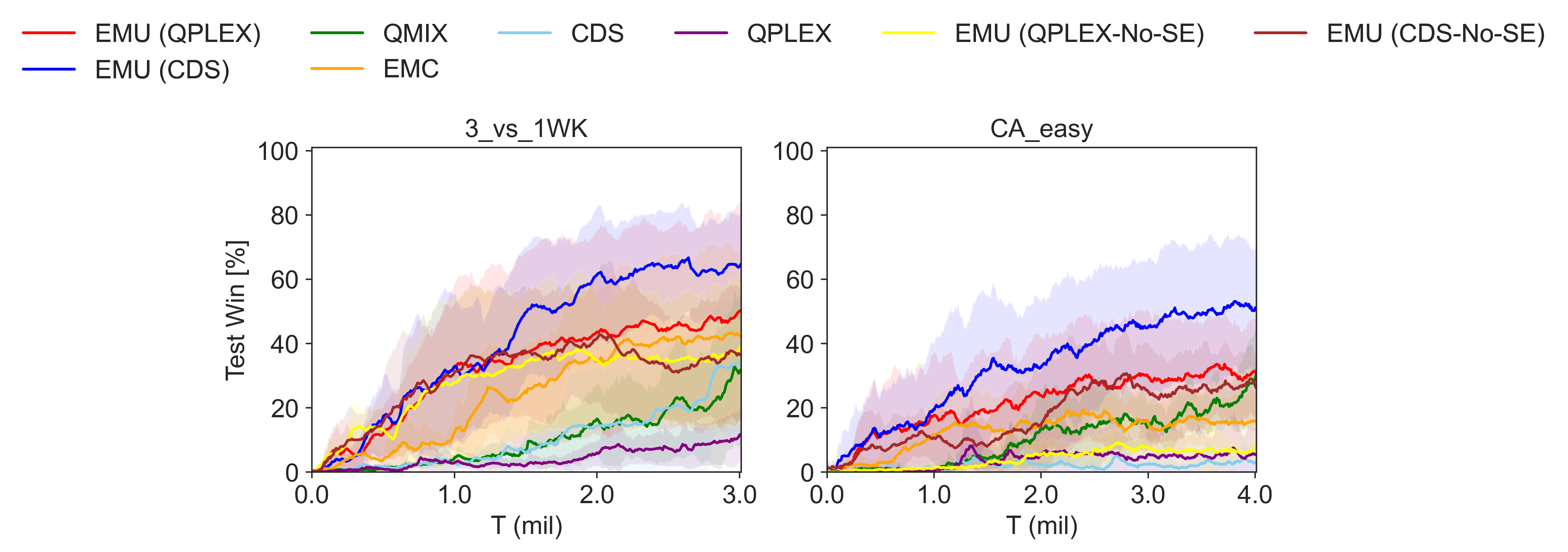}}
\end{center}
\vspace{-0.25in}
\caption{Performance comparison of EMU against baseline algorithms on Google Research Football.}
\label{fig:grf_performance_with_ablation}
\vspace{-0.10in}
\end{figure}

For relatively easy tasks, EMU (QPLEX-No-SE) and EMU (CDS-No-SE) show comparable performance at first but they converge on sub-optimal policy in most tasks. Especially, this characteristic is well observed in the case of EMU (CDS-No-SE). As large size of memories are stored in an episodic buffer as training goes on, the probability of recalling similar memories increases. However, with random projection, semantically incoherent memories can be recalled and thus it can adversely affect the value estimation. We deem this is the reason for the convergence on suboptimal policy in the case of EMU (No-SE). 
Thus we can conclude that recalling semantically coherent memory is an essential component of EMU.
\newpage

\subsection{Additional Ablation on $r^c$}
\label{additional_rc_ablation}
\begin{figure}[!hbt] 
\begin{center}
	\begin{minipage}[!htb]{0.8\linewidth} 
		\subfloat[\centering \texttt{3s\_vs\_5z} ]{{\includegraphics[width=0.33\linewidth]{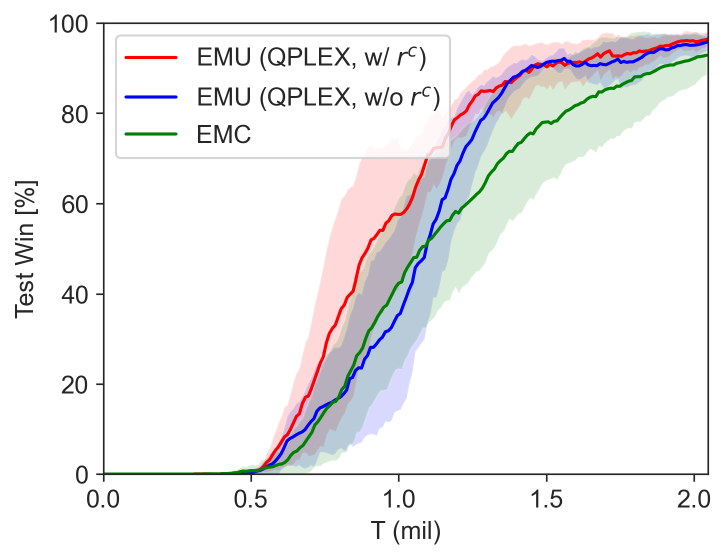} }}
		\subfloat[\centering \texttt{5m\_vs\_6m} ]{{\includegraphics[width=0.33\linewidth]{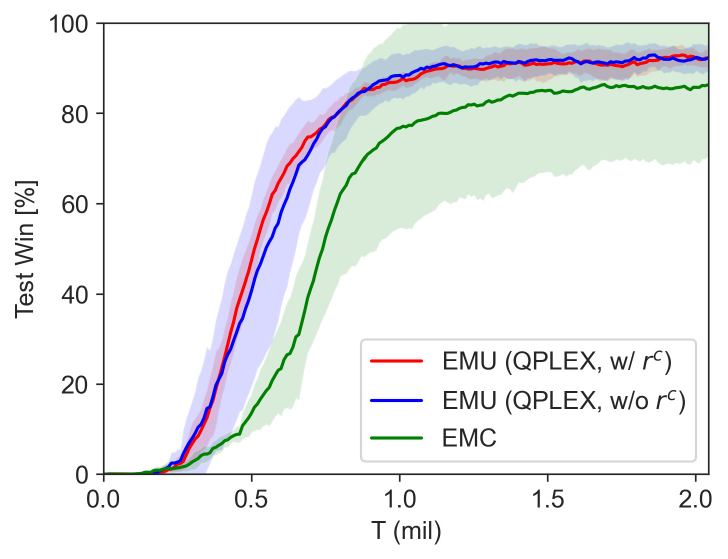} }}
		\subfloat[\centering \texttt{3s5z\_vs\_3s6z} ]{{\includegraphics[width=0.33\linewidth]{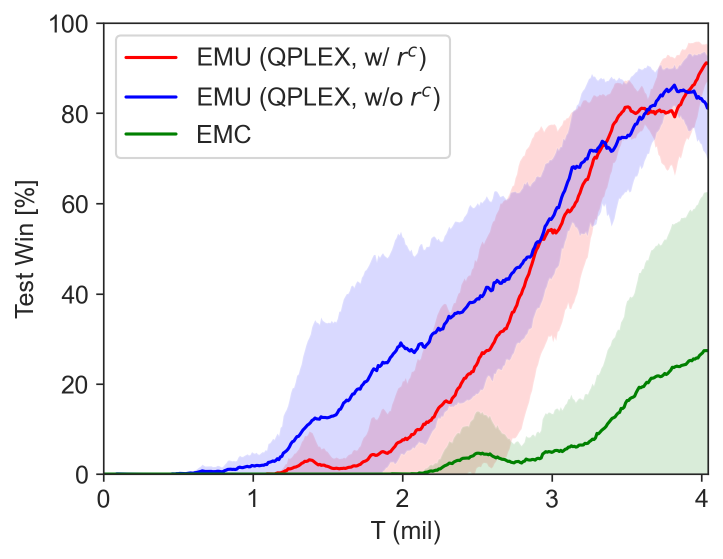} }}
		\vspace{-0.1in}
	\end{minipage}
\end{center}
\caption{Ablation studies on $r^c$ in SMAC tasks.}%
\label{fig:ablation_rc_1}%
\vspace{-0.1in}
\end{figure}

\begin{figure}[!hbtp] 
\begin{center}
	\begin{minipage}[!htb]{0.8\linewidth} 
		\subfloat[\centering \texttt{3s\_vs\_5z} ]{{\includegraphics[width=0.33\linewidth]{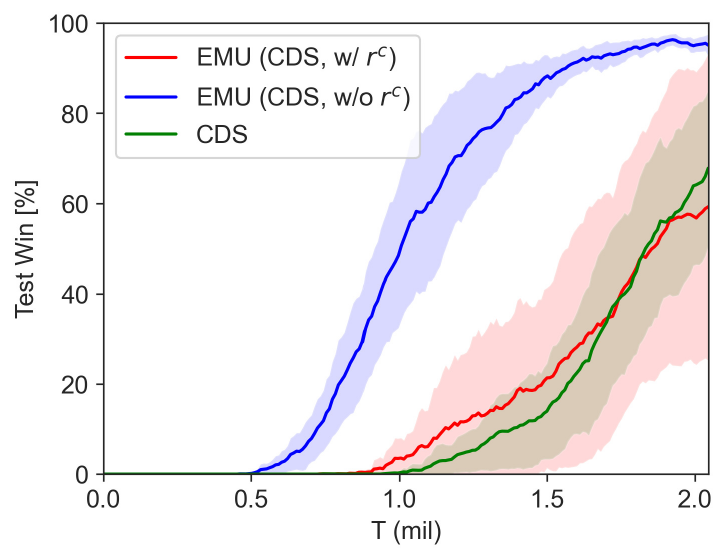} }}
		\subfloat[\centering \texttt{5m\_vs\_6m} ]{{\includegraphics[width=0.33\linewidth]{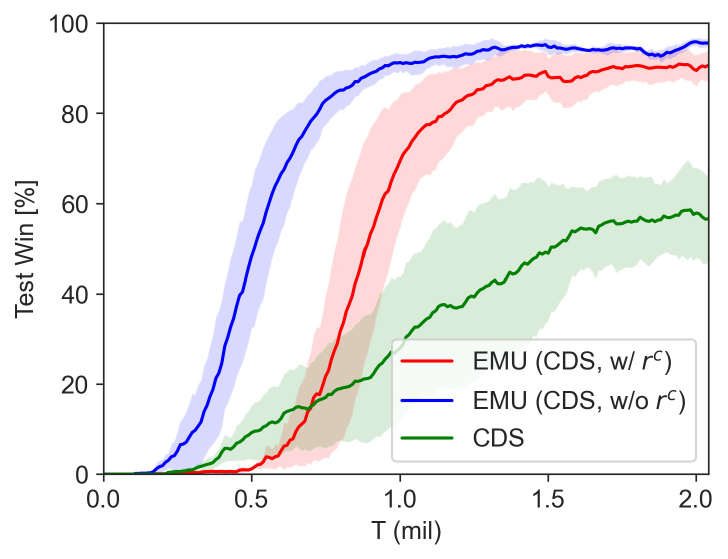} }}
		\subfloat[\centering \texttt{3s5z\_vs\_3s6z} ]{{\includegraphics[width=0.33\linewidth]{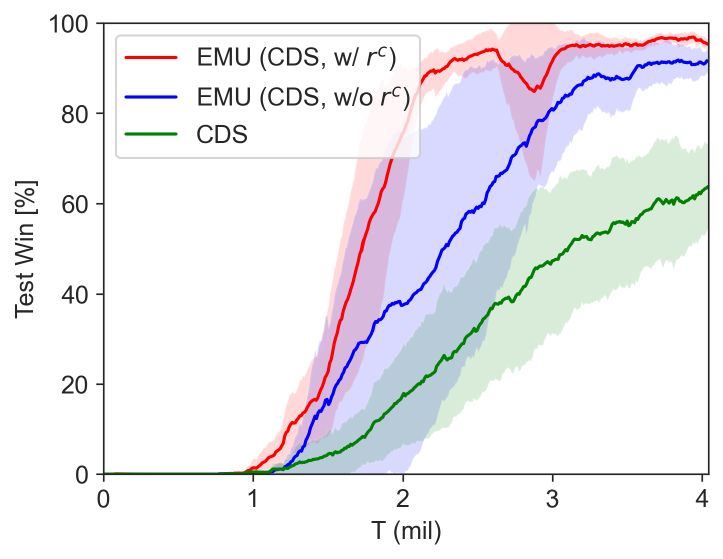} }}
		\vspace{-0.1in}
	\end{minipage}
\end{center}
\caption{Ablation studies on $r^c$ in SMAC tasks.}%
\label{fig:ablation_rc_2}%
\vspace{-0.1in}
\end{figure}

In Eq.\ref{Eq:action_policy_loss}, we introduce $r^c$ as an additional reward which may encourage exploratory behavior or coordination. The reason we introduce $r^c$ is to show that EMU can be used in conjunction with any form of incentive encouraging further exploration. Our method may not be strongly effective until some desired states are found, although it has exploratory behavior via the proposed semantic embeddings, controlled by $\delta$. Until then, such incentives could be beneficial to find desired or goal states. Figures \ref{fig:ablation_rc_1}-\ref{fig:ablation_rc_2} show the ablation study of with and without $r^c$, and the contribution of $r^c$ is limited compared to $r^p$.			

\subsection{Comparison of Episodic Incentive with Exploratory Incentive}
\label{additional_incentive_ablation}
In this subsection, we replace the episodic incentive with another exploratory incentive, introduced by \citep{henaff2022exploration}. In \citep{henaff2022exploration}, the authors extend the count-based episodic bonuses to continuous spaces by introducing episodic elliptical bonuses for exploration. In this concept, a high reward is given when the state projected in the embedding space is different from the previous states within the same episode. In detail, with a given feature encoder $\phi$, the elliptical bonus $b_t$ at timestep $t$ is computed as follows:
\begin{equation}\label{Eq:elliptical_incentive}
\begin{aligned}
	b_t=\phi(s_t)^T C_t^{-1} \phi(s_t)
\end{aligned}
\end{equation}
where $C_t^{-1}$ is an inverse covariance matrix with an initial value of $C_{t=0}^{-1}=1/\lambda_{e3b} \bm{I}$. Here, $\lambda_{e3b}$ is a covariance regularizer. For update inverse covariance, the authors suggested a computationally efficient update as
\begin{equation}\label{Eq:inverse_cov_update}
\begin{aligned}
	C_{t+1}^{-1}=C_{t}^{-1}-\frac{1}{1+b_{t+1}}uu^T
\end{aligned}
\end{equation}
where $u=C_t^{-1}\phi(s_{t+1})$. Then, the final reward $\bar{r}_t$ with episodic elliptical bonuses $b_t$ is expressed as
\begin{equation}\label{Eq:inverse_cov_update}
\begin{aligned}
	\bar{r}_t = r_t + \beta_{e3b} b_t
\end{aligned}
\end{equation}
where $\beta_{e3b}$ and $r_t$ are a corresponding scale factor and external reward given by the environment, respectively.

\begin{figure}[!h]
\vspace{-0.0in}
\begin{center}			
\centerline{\includegraphics[width=0.65\linewidth]{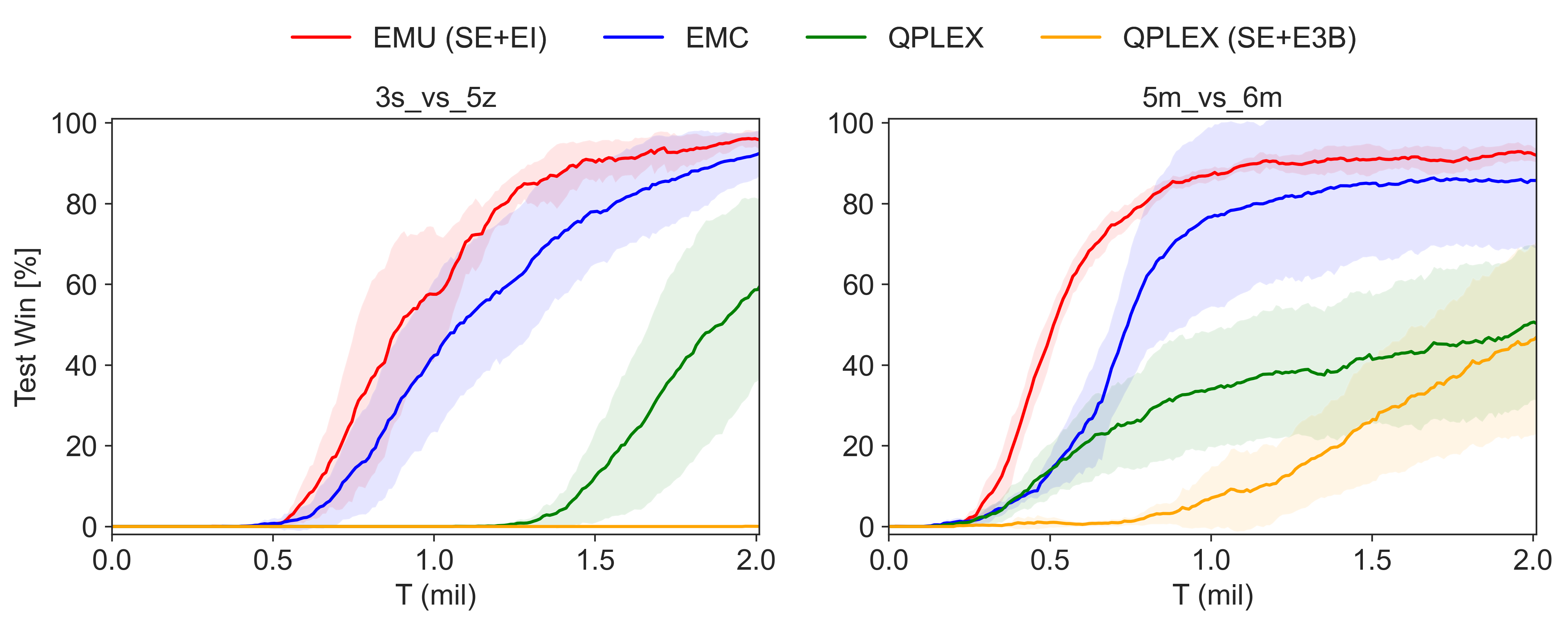}}
\end{center}
\vspace{-0.25in}
\caption{Performance comparison with elliptical incentive on selected SMAC maps.}
\label{fig:performance_comp_e3b}
\vspace{-0.10in}
\end{figure}

For this comparison, we utilize the dCAE structure as a state embedding function $\phi$. For a mixer, QPLEX \citep{wang2020qplex} is adopted for all cases, and we denote the case with an elliptical incentive instead of the proposed episodic incentive as QPLEX (SE+E3B). Figure \ref{fig:performance_comp_e3b} illustrates the performance of adopting an elliptical incentive for exploration instead of the proposed episodic incentive. QPLEX (SE+E3B) uses the same hyperparameters with EMU (SE+EI) and we set $\lambda_{e3b}=0.1$ according to \cite{henaff2022exploration}.

As illustrated by Figure \ref{fig:performance_comp_e3b}, adopting an elliptical incentive presented by \citep{henaff2022exploration} instead of an episodic incentive does not give any performance gain and even adversely influences the performance compared to QPLEX. It seems that adding excessive surprise-based incentives can be a disturbance in MARL tasks since finding a new state itself does not guarantee better coordination among agents. In MARL, agents need to find the proper combination of joint action in a given similar observations when finding an optimal policy. On the other hand, in high-dimensional pixel-based single-agent tasks such as Habitat \citep{ramakrishnan2021habitat}, finding a new state itself can be beneficial in policy optimization. From this, we can note that adopting a certain algorithm from a single-agent RL case to MARL case may require a modification or adjustment with domain knowledge.

\begin{figure}[!h]
\vspace{-0.0in}
\begin{center}			
\centerline{\includegraphics[width=0.65\linewidth]{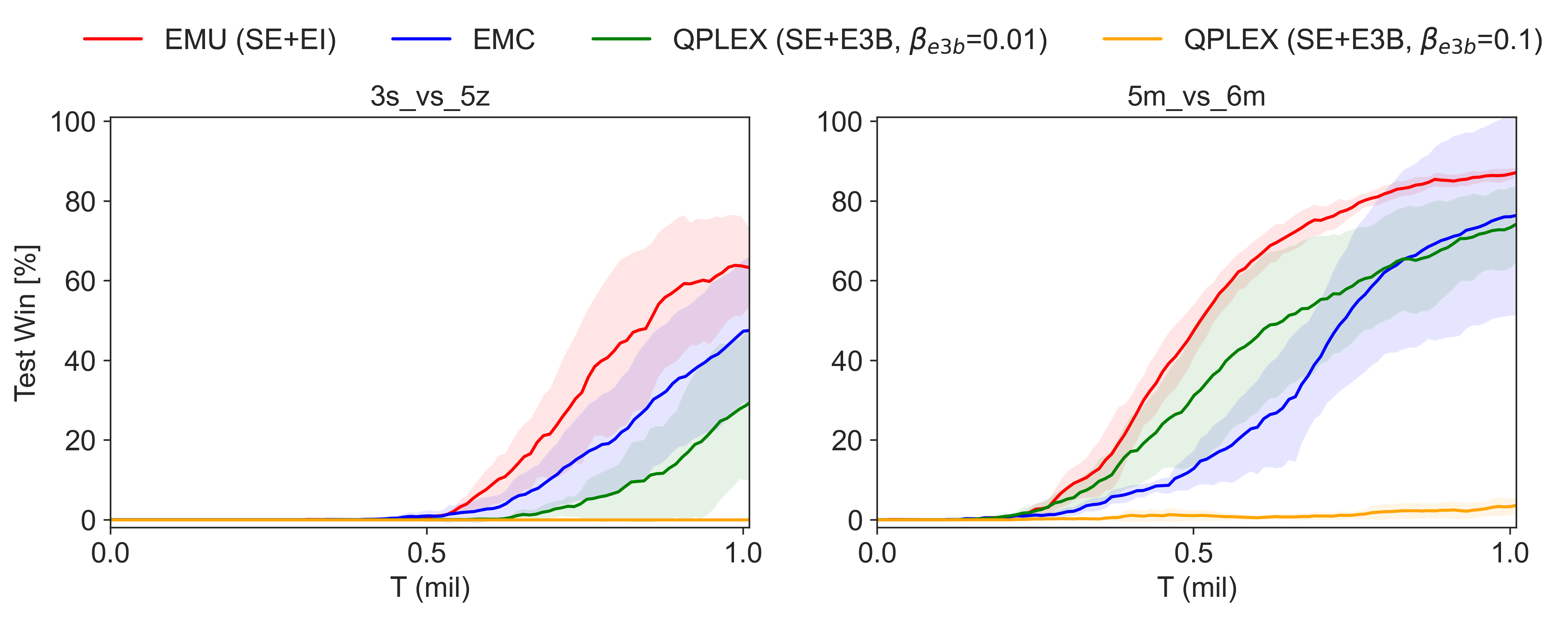}}
\end{center}
\vspace{-0.25in}
\caption{Performance comparison with an elliptical incentive on selected SMAC maps.}
\label{fig:performance_comp_e3b_param}
\vspace{-0.10in}
\end{figure}

As a simple tuning, we conduct parametric study for $\beta_{e3b}=\left\{0.01, 0.1\right\}$ to adjust magnitude of incentive of E3B. Figure \ref{fig:performance_comp_e3b_param} illustrates the results. In Figure \ref{fig:performance_comp_e3b_param}, QPLEX (SE+E3B) with $\beta_{e3b}=0.01$ shows a better performance than the case with $\beta_{e3b}=0.1$ and comparable performance to EMC in \texttt{5m\_vs\_6m}. However, EMU with the proposed episodic incentive shows the best performance.
From this comparison, we can see that incentives proposed by previous work need to be adjusted according to the type of tasks, as it was done in EMC \citep{zheng2021episodic}. On the other hand, with the proposed episodic incentive we do not need such hyperparameter-scaling, allowing much more flexible application across various tasks.

\newpage
\subsection{Additional Toy Experiment and Applicability Tests}
\label{additional_applicability}
In this section, we conduct additional experiments on the didactic example presented by \citep{zheng2021episodic} to see how the proposed method would behave in a simple but complex coordination task. Additionally, by defining $R_{thr}$ to define the desirability presented in Definition \ref{def: desirable trajectory}, we can extend EMU to a single-agent RL task, where a strict goal is not defined in general. 

\begin{figure}[!hbtp] 
\begin{center}
\centering	
\begin{minipage}[!h]{0.7\linewidth}
	\subfloat[\centering Gridworld]{{\includegraphics[width=0.50\linewidth]{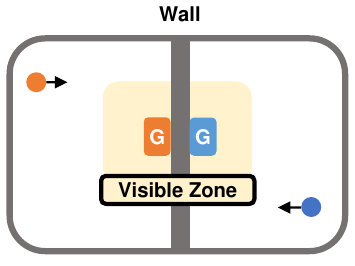} }}%
	\qquad
	\subfloat[\centering Performance evaluation ($p=2$)]{{\includegraphics[width=0.45\linewidth]{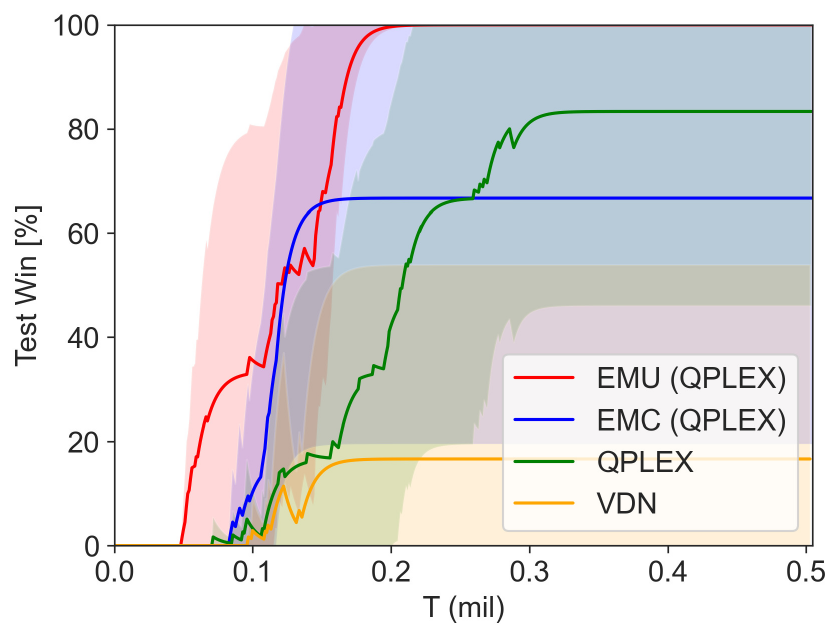} }}%
	\caption{Didactic experiments on gridworld.}%
	\label{fig:didactic_example}%
	\vspace{-0.2in}
\end{minipage}
\end{center}
\end{figure}

\textbf{Didactic experiment on Gridworld} We adopt the didactic example such as gridworld environment from \citep{zheng2021episodic} to demonstrate the motivation and how the proposed method can overcome the existing limitations of the conventional episodic control. In this task, two agents in gridworld (see Figure \ref{fig:didactic_example}(a)) need to reach their goal states at the same time to get a reward $r=10$ and if only one arrives first, they get a penalty with the amount of $-p$. Please refer to \citep{zheng2021episodic} for further details.

To see the sole effect of the episodic control, we discard the curiosity incentive part of EMC, and for a fair comparison, we set the same exploration rate of $\epsilon$-greedy with $T_{\epsilon}=200K$ for all algorithms. We evaluate the win-rate with 180 samples (30 episodes for each training random seed and 6 different random seeds) at each training time. Notably, adopting episodic control with a naive utilization suffers from local convergence (see QPLEX and EMC (QPLEX) in Figure \ref{fig:didactic_example}(b)), even though it expedites learning efficiency at the early training phase. On the other hand, EMU shows more robust performance under different seed cases and achieves the best performance by an efficient and discreet utilization of episodic memories.

\textbf{Applicability test to single agent RL task} We first need to define $R_{thr}$ value to effectively apply EMU to a single-agent task where a goal of an episode is generally not strictly defined, unlike cooperative multi-agent tasks with a shared common goal.

In a single-agent task where the action space is continuous such as MuJoCo \citep{todorov2012mujoco}, the actor-critic method is often adopted. Efficient memory utilization of EMU can be used to train the critic network and thus indirectly influence policy learning, unlike general cooperative MARL tasks where value-based RL is often considered.

We implement EMU on top of TD3 and use the open-source code presented in \citep{fujimoto2018addressing}. We begin to train the model after sufficient data is stored in the replay buffer and conduct \textbf{6 times of training per episode with 256 mini-batches.} Note that this is different from the default settings of RL training, which conducts training at each timestep. Our modified setting aims to see the effect on the sample efficiency of the proposed model. The performance of the trained model is evaluated at every 50k timesteps.

We use the same hyperparameter settings as in MARL task presented in Table \ref{tab:hyperparam_emb} except for the update interval, $t_{emb}=100K$ according to large episodic timestep in single-RL compared to MARL tasks. It is worth mentioning that additional customized parameter settings for single-agent tasks may further improve the performance. In our evaluation, three single-agent tasks such as \texttt{Hopper-v4}, \texttt{Walker2D-v4} and \texttt{Humanoid-v4} are considered, and Figure \ref{fig:single_rl_task} illustrates each task. Here, $\delta_2=1.3e-5$ is used for \texttt{Hopper-v4} and \texttt{Walker2D-v4}, and $\delta_3=1.3e-3$ is used for \texttt{Humanoid-v4} as \texttt{Humanoid-v4} task contains much higher state dimension space as 376-dimension. Please refer to \cite{todorov2012mujoco} for a detailed description of tasks.

\begin{figure}[!htbp]
\begin{center}
\begin{subfigure}{0.25\linewidth}
	\includegraphics[width=\linewidth]{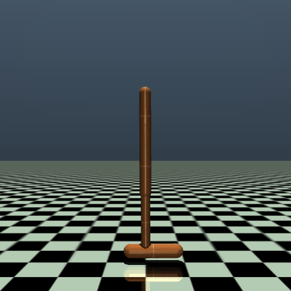}
	\caption{\texttt{Hopper-v4}}
\end{subfigure}\quad  
\begin{subfigure}{0.25\linewidth}
	\includegraphics[width=\linewidth]{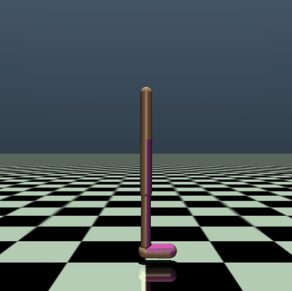}
	\caption{\texttt{Walker2D-v4}}
\end{subfigure}\quad 
\begin{subfigure}{0.25\linewidth}
	\includegraphics[width=\linewidth]{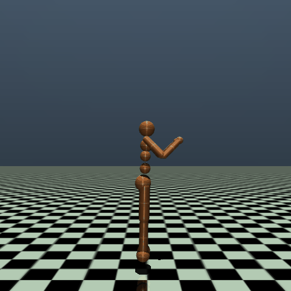}
	\caption{\texttt{Humanoid-v4}}
\end{subfigure}
\end{center}
\caption{Illustration of MuJoCo scenarios.}
\label{fig:single_rl_task}
\end{figure} 

\begin{figure}[!htbp]
\vspace{-0.1in}
\begin{center}
\begin{subfigure}{0.3\linewidth}
	\includegraphics[width=\linewidth]{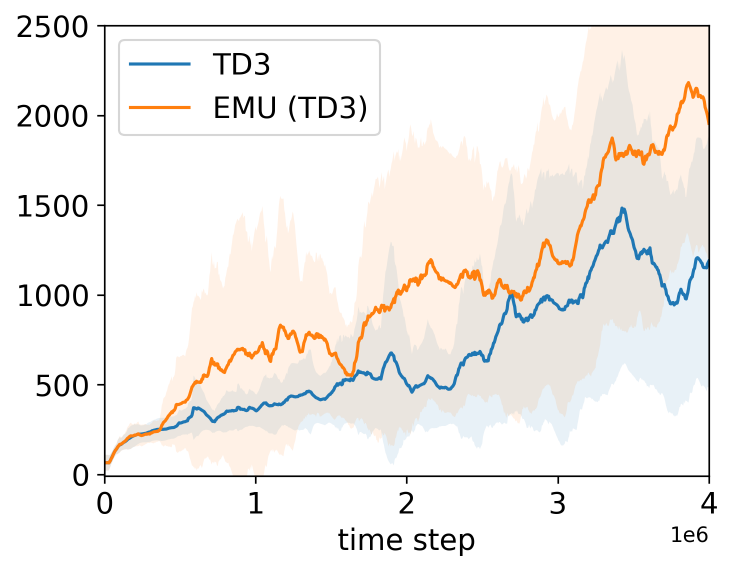}
	\caption{Performance (\texttt{Hopper})}
\end{subfigure}\quad 
\begin{subfigure}{0.3\linewidth}
	\includegraphics[width=\linewidth]{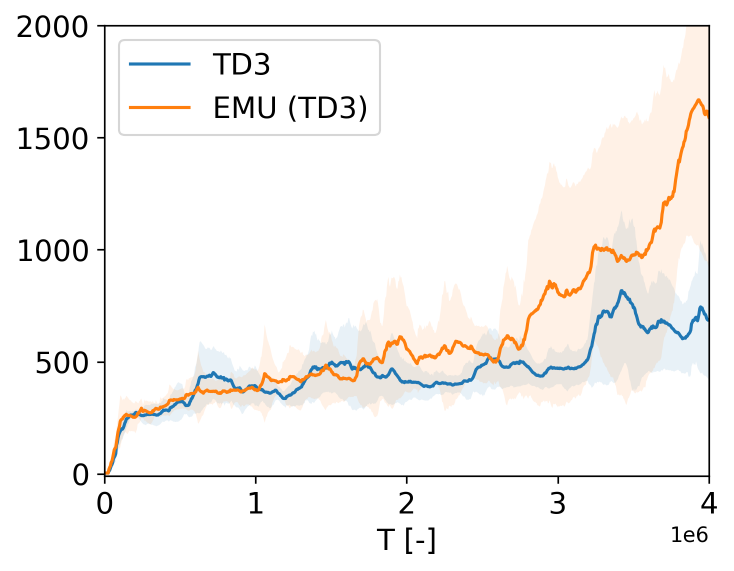}
	\caption{Performance (\texttt{Walker2D})}
\end{subfigure}\quad
\begin{subfigure}{0.3\linewidth}
	\includegraphics[width=\linewidth]{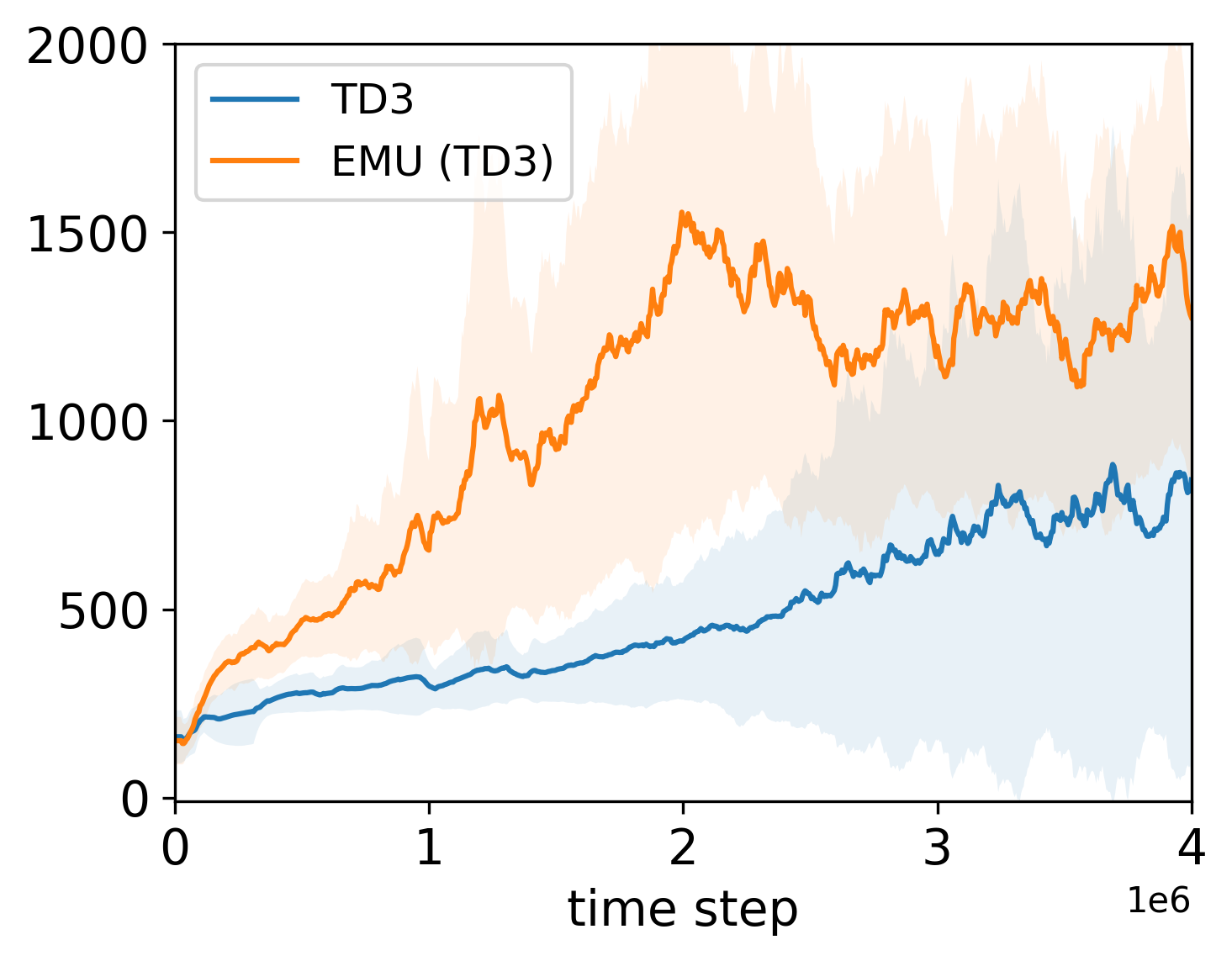}
	\caption{Performance (\texttt{Humanoid)}}
\end{subfigure}
\end{center}
\caption{Applicability test to single agent task ($R_{thr}=500$).}
\label{fig:single_rl_task}
\end{figure} 

In Figure \ref{fig:single_rl_task}, EMU (TD3) shows the performance improvement compared to the original TD3. Thanks to semantically similar memory recall and episodic incentive, states deemed desirable could have high values, and trained policy is encouraged to visit them more frequently. As a result, EMU (TD3) shows the better performance. Interestingly, under state dimension as \texttt{Humanoid-v4} task, TD3 and EMU (TD3) show a distinct performance gap in the early training phase. This is because, in a task with a high-dimensional state space, it is hard for a critic network to capture important features determining the value of a given state. Thus, it takes longer to estimate state value accurately. However, with the help of semantically similar memory recall and error compensation through episodic incentive, a critic network in EMU (TD3) can accurately estimate the value of the state much faster than the original TD3, leading to faster policy optimization.

\begin{figure}[!htbp]
\begin{center}
\begin{subfigure}{0.3\linewidth}
	\includegraphics[width=\linewidth]{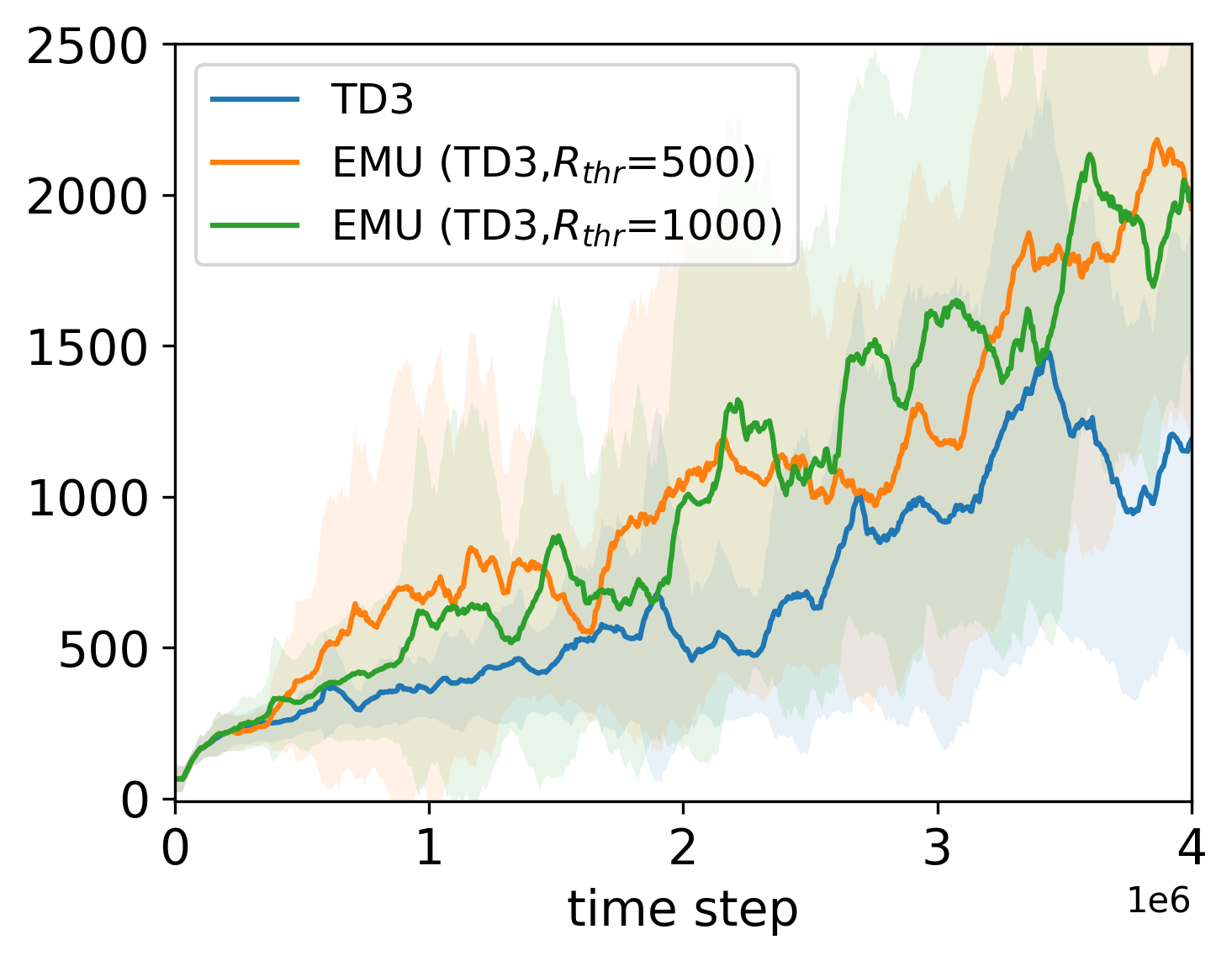}
	\caption{\texttt{\texttt{Hopper-v4}}}
\end{subfigure}		
\begin{subfigure}{0.3\linewidth}
	\includegraphics[width=\linewidth]{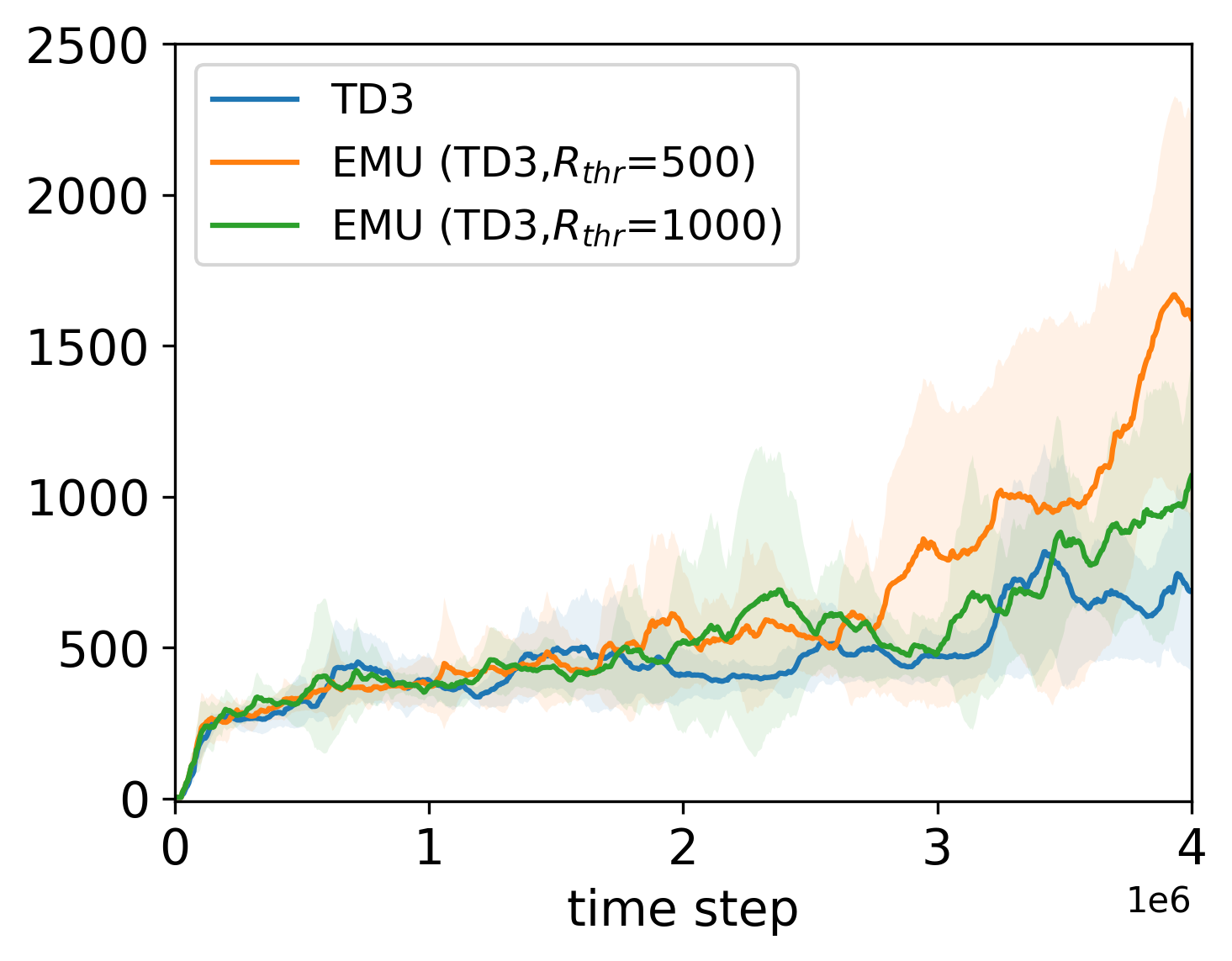}
	\caption{\texttt{\texttt{Walker2d-v4}}}
\end{subfigure}\hfill		
\end{center}
\caption{Parametric study on $R_{thr}$.}
\label{fig:R_thr_study}
\vskip -0.20in
\end{figure}

Unlike cooperative MARL tasks, single-RL tasks normally do not have a desirability threshold. Thus, one may need to determine $R_{thr}$ based on domain knowledge or a preference for the level of return to be deemed successful. Figure \ref{fig:R_thr_study} presents a performance variation according to $R_{thr}$.

When we set $R_{thr}=1000$ in \texttt{Walker2d} task, desirability signal is rarely obtained compared to the case with $R_{thr}=500$ in the early training phase. Thus, EMU with $R_{thr}=500$ shows the better performance. However, both cases of EMU show better performance compared to the original TD3. In \texttt{Hopper} task, both cases of $R_{thr}=500$ and $R_{thr}=1000$ show the similar performance. Thus, when determining $R_{thr}$, it can be beneficial to set a small value rather than a large one that can be hardly obtained.

Although setting a small $R_{thr}$ does not require much domain knowledge, a possible option to detour this is a periodic update of desirability based on the average return value $H(s)$ in all $s \in \gD_{E}$. In this way, a certain state with low return which was originally deemed as \textit{desirable} can be reevaluated as \textit{undesirable} as training proceeds. The episodic incentive is not further given to those undesirable states.

\textbf{Scalability to image-based single-agent RL task}
Although MARL tasks already contain high-dimension state space such as 322-dimension in \texttt{MMM2} and 282-dimension
in \texttt{corridor}, image-based single RL tasks, such as Atari \cite{bellemare2013arcade} game, often accompany higher state spaces such as [210x160x3] for "RGB" and [210x160] for "grayscale". We use the "grayscale" type for the following experiments. For the details of the state space in MARL task, please see Appendix \ref{details_marl_tasks}.

In an image-based task, storing all state values to update all the key values in $\gD_{E}$ as $f_{\phi}$ updates can be memory-inefficient, and a semantic embedding from original states may become overhead compared to the case without it. In such case, one may resort to a pre-trained feature extraction model such as ResNet model provided by torch-vision in a certain amount for dimension reduction only, before passing through the proposed semantic embedding. The feature extraction model above is not an object of training.

\begin{figure}[!htbp]
\begin{center}
\begin{subfigure}{0.225\linewidth}
	\includegraphics[width=\linewidth]{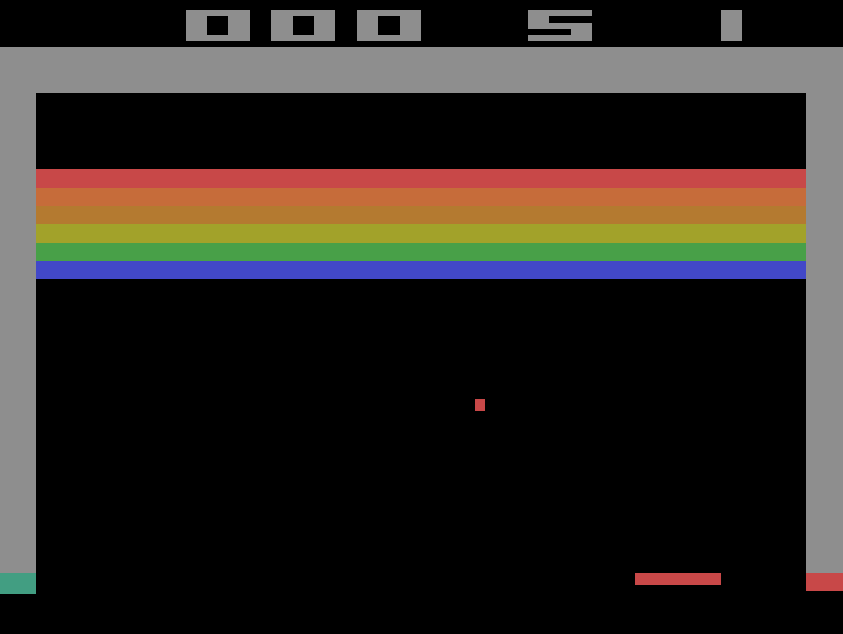}
	\vskip +0.11in
	\caption{\texttt{Breakout}}
\end{subfigure}\hfill
\begin{subfigure}{0.275\linewidth}
	\includegraphics[width=\linewidth]{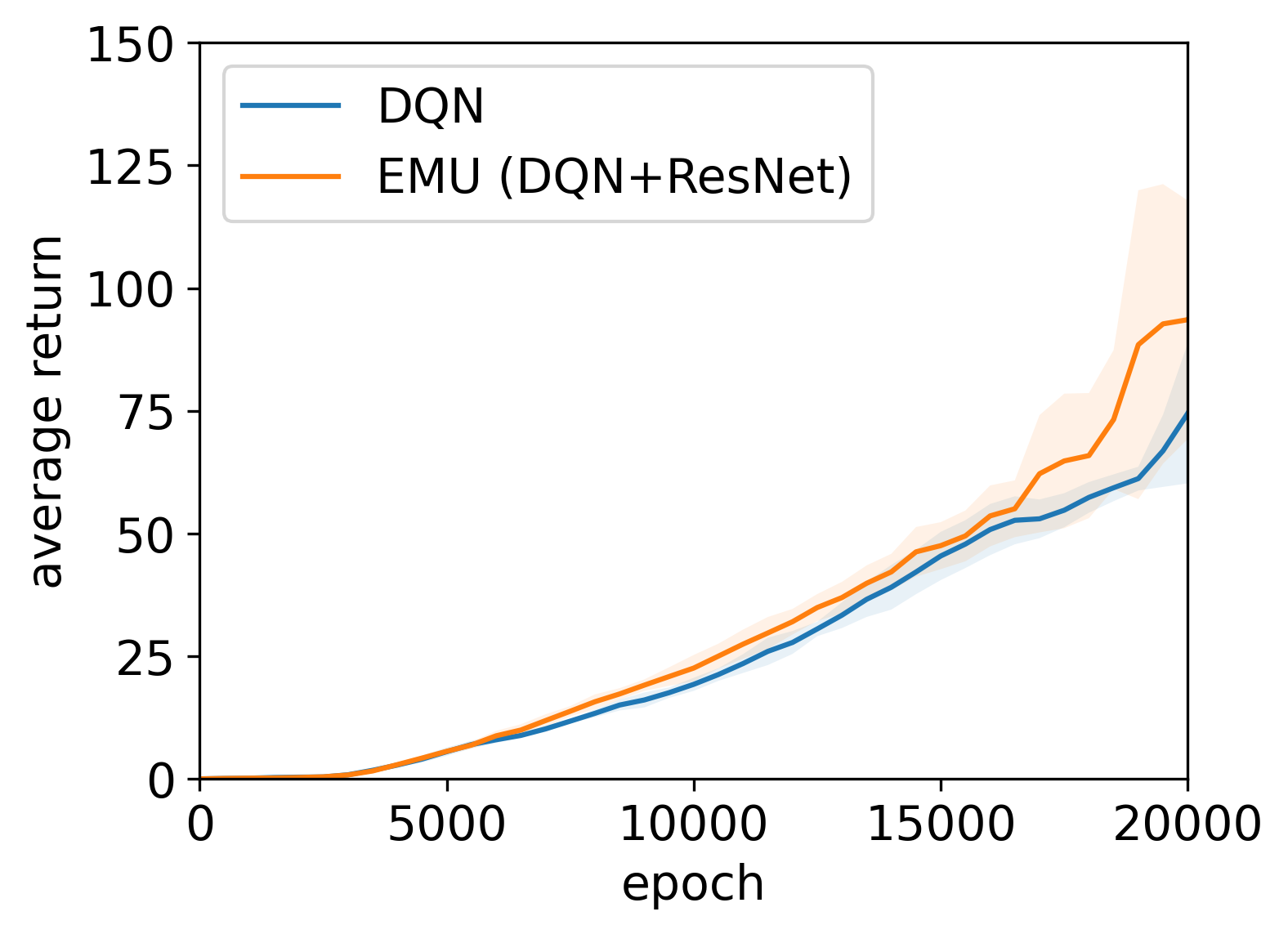}
	\caption{Performance (\texttt{Breakout})}
\end{subfigure}\hfill 
\begin{subfigure}{0.225\linewidth}
	\includegraphics[width=\linewidth]{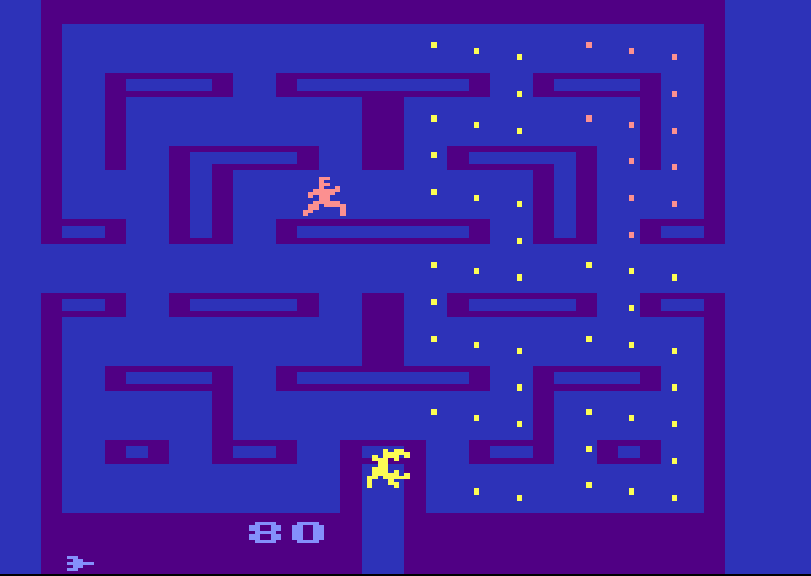}
	\vskip +0.15in
	\caption{\texttt{Alien}}
\end{subfigure}\hfill
\begin{subfigure}{0.275\linewidth}
	\includegraphics[width=\linewidth]{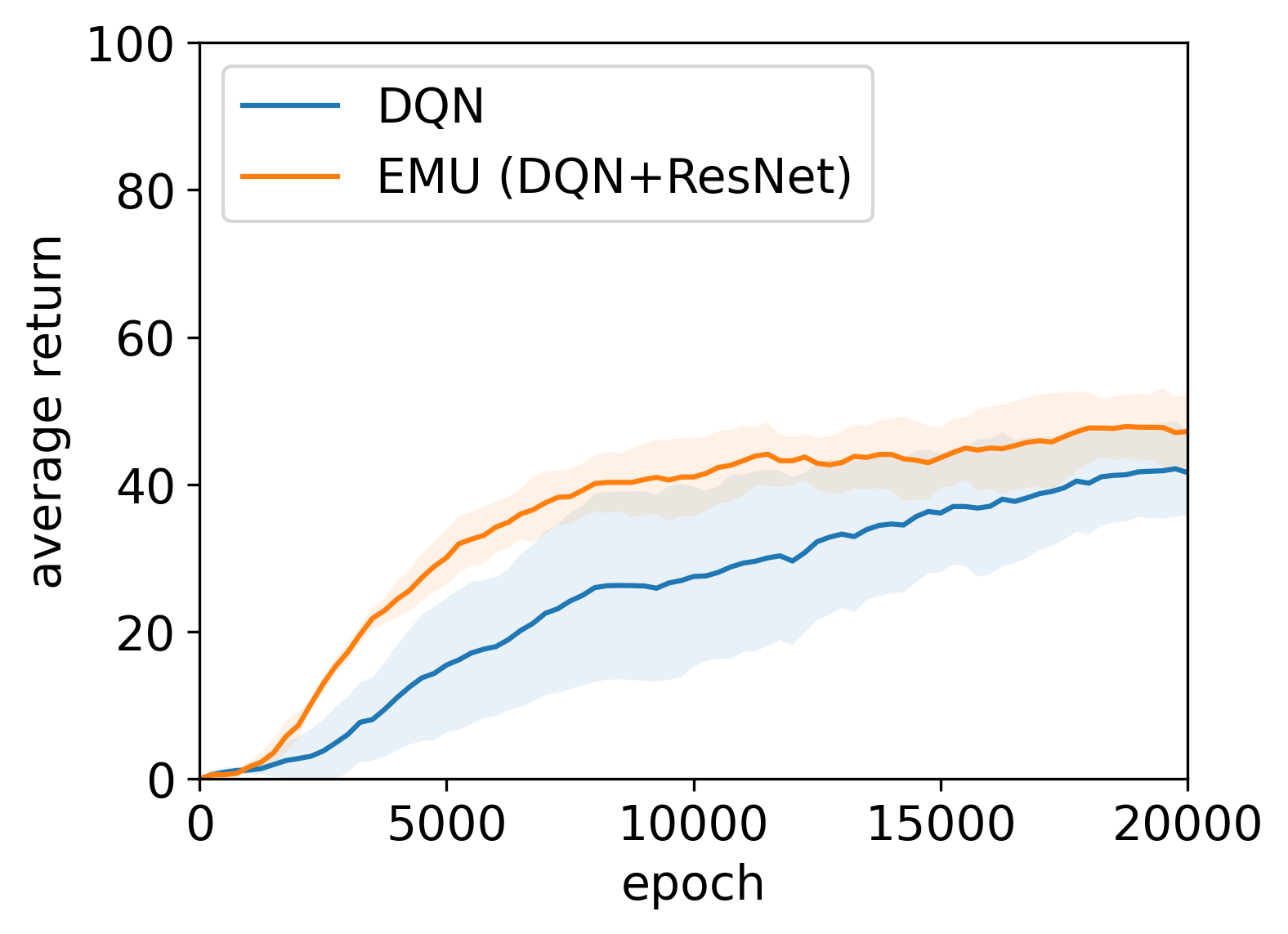}
	\caption{Performance (\texttt{Alien)}}
\end{subfigure}
\end{center}
\caption{Image-based single-RL task example.}
\label{fig:image_single_rl_task}
\vskip -0.20in
\end{figure}

As an example, we implement EMU on the top of DQN model and compare it with the original DQN on Atari task. For the EMU (DQN), we adopt some part of pre-trained ResNet18 presented by torch-vision for dimensionality reduction, before passing an input image to semantic embedding. At each epoch, 320 random samples are used for training in \texttt{Breakout} task, and 640 random samples are used in \texttt{Alien} task. The same mini-batch size of 32 is used for both cases. For $f_{\phi}$ training, the same parameters presented in Table \ref{tab:hyperparam_emb} are adopted except for the $t_{emb}=10K$ considering the timestep of single RL task. We also use the same $\delta_2=1.3e-5$ and set $R_{thr}=50$ for \texttt{Breakout} and $R_{thr}=40$ for \texttt{Alien}, respectively. Please refer to \cite{bellemare2013arcade} and \texttt{https://gymnasium.farama.org/environments/atari} for task details. As in Figure \ref{fig:image_single_rl_task}, we found a performance gain by adopting EMU on high-dimensional image-based tasks.

\newpage

\section{Training Algorithm}\label{Training_algorithm}
\subsection{Memory Construction}\label{memory_contruction}
During the centralized training, we can access the information on whether the episodic return reaches the highest return $R_{\textrm{max}}$ or threshold $R_{thr}$, i.e., defeating all enemies in SMAC or scoring a goal in GRF. 
When storing information to $\gD_{E}$, by the definition presented Definition. \ref{def: desirable trajectory}, we set $\xi(s)=1$ for $\forall s \in \mathcal{T}_{\xi}$. 

For efficient memory construction, we propagate the desirability of the state to a similar state within the threshold $\delta$. With this desirability propagation, similar states have an incentive for a visit. In addition, once a memory is saved in $\gD_{E}$, the memory is preserved until it becomes obsolete (the oldest memory to be recalled). When a desirable state is found near the existing suboptimal memory within $\delta$, we replace the suboptimal memory with the desirable one, which gives the effect of a memory shift to the desirable state. Algorithm \ref{alg:memory_construction} presents the memory construction with the desirability propagation and memory shift.

\begin{algorithm}[H]
\begin{center}
\caption{Episodic memory construction}\label{alg:memory_construction}
\begin{algorithmic}[1]								
	\State $\xi_{\mathcal{T}}$: Optimality of trajectory
	\State $\mathcal{T}=\left\{s_0,\bm{a_0},r_0,s_1,...,s_T\right\}$: Episodic trajectory 
	\State Initialize $R_{t}=0$
	\For{$t=T$ to $0$}
	\State Compute $x_t=f_{\phi}(s_t)$ and $y_t=(x_t-\hat{\mu}_x)/\hat{\sigma}_x$
	\State pick the nearest neighbor $\hat{x}_t \in \gD_{E}$ and get $\hat{y}_t$.
	\If {$||\hat{y}_t-y_t||_2<\delta$}
	\State $N_{call}(\hat{x}_t) \leftarrow N_{call}(\hat{x}_t) + 1$
	\If {$\xi_{\mathcal{T}} == 1$}
	\State $N_{\xi}(\hat{x}_t) \leftarrow N_{\xi}(\hat{x}_t) + 1$
	\EndIf
	\If {$\xi_t == 0$ and $\xi_{\mathcal{T}} == 1$}
	\State $\xi_t \leftarrow \xi_{\mathcal{T}}$ \textcolor{gray}{\Comment{desirability propagation}}
	\State $\hat{x}_t \leftarrow x_t$, $\hat{y}_t \leftarrow y_t$, $\hat{s}_t \leftarrow s_t$ \textcolor{gray}{\Comment{memory shift}}
	\State $\hat{H}_t \leftarrow R_t$				
	\Else{} 
	\If{$\hat{H}_t < R_t $} $\hat{H}_t \leftarrow R_t$			
	\EndIf
	\EndIf
	\Else{}
	\State Add memory $\gD_{E} \leftarrow (x_t, y_t, R_t, s_t, \xi_t)$
	\EndIf
	\EndFor				
\end{algorithmic}
\end{center}
\end{algorithm}

For memory capacity and latent dimension, we used the same values as \citet{zheng2021episodic}, and Table \ref{tab:hyperparam_memory} shows the summary of hyperparameter related to episodic memory.

\begin{table}[htbp]
\centering
\caption{Configuration of Episodic Memory.}
\vspace{0.3cm}
\label{tab:hyperparam_memory}
\begin{tabular}{cc}
\toprule
Configuration                                                                                       & Value \\ 
\midrule 
episodic latent dimension, $\dim(x)$                                                                 & 4     \\
episodic memory capacity                                                                            & 1M    \\
\begin{tabular}[c]{@{}c@{}}a scale factor, $\lambda$\\ (for conventional episodic control only)\end{tabular} & 0.1   \\ 
\bottomrule
\end{tabular}\label{memory_parameter}
\end{table}

\begin{table}[htbp]
\centering
\caption{Additional CPU memory usage to save global states.}
\vspace{0.3cm}
\label{cpu_memory_usage}
\begin{tabular}{cc}
\toprule
SMAC task      & \begin{tabular}[c]{@{}c@{}}CPU memory usage (1M data) \\ (GiB)\end{tabular} \\ 
\midrule
5m\_vs\_6m     & 0.4                                                                               \\
3s5z\_vs\_3s6z & 0.9                                                                               \\
MMM2           & 1.2                                                                               \\ 
\bottomrule
\end{tabular}
\end{table}

The memory construction for EMU seems to require a significantly large memory space, especially for saving global states $s$. However, $\gD_E$ uses CPU memory instead of GPU memory, and the memory required for the proposed embedder structure is minimal compared to the memory usage of original RL training (<1\%). Thus, a memory burden due to a trainable embedding structure is negligible. Table \ref{cpu_memory_usage} presents examples of CPU memory usage to save global states $s \in \gD_{E}$.

\subsection{Overall Training Algorithm}
\label{overall_training_algorithm}
In this section, we present details of the overall MARL training algorithm including training of $f_{\phi}$. Additional hyperparameters related to Algorithm \ref{alg:embedder_training} to update encoder $f_{\phi}$ and decoder $f_{\psi}$ are presented in Table \ref{tab:hyperparam_emb}. Note that variables $N$ and $B$ are consistent with Algorithm \ref{alg:embedder_training}.
\begin{table}[!htbp]
\centering
\caption{EMU Hyperparameters for $f_{\phi}$ and $f_{\psi}$ training.}
\vspace{0.3cm}
\label{tab:hyperparam_emb}
\begin{tabular}{cc}
\toprule
Configuration                                       & Value                      \\ 
\midrule 
\multicolumn{1}{l}{a scale factor of reconstruction loss, $\lambda_{rcon}$}      & 
\multicolumn{1}{c}{0.1}     \\
\multicolumn{1}{l}{update interval, $t_{emb}$}      & \multicolumn{1}{c}{1K}     \\
\multicolumn{1}{l}{training samples, $N$}     & \multicolumn{1}{c}{102.4K} \\
\multicolumn{1}{l}{batch size of training, $B$} & \multicolumn{1}{c}{1024}     \\ 
\bottomrule
\end{tabular}
\end{table}

Algorithm \ref{alg:EMU} presents the pseudo-code of overall training for EMU. In Algorithm \ref{alg:EMU}, network parameters related to a mixer and individual Q-network are denoted as $\theta$, and double Q-learning with target network is adopted as other baseline methods \citep{rashid2018qmix,rashid2020weighted,wang2020qplex,zheng2021episodic,chenghao2021celebrating}.\\
\begin{algorithm}[htbp]
\begin{center}
\begin{algorithmic}[1]		
	\State $\gD$: Replay buffer
	\State $\gD_{E}$: Episodic buffer
	\State $Q_{\theta}^i$: Individual Q-network of $n$ agents
	\State $M$: Batch size of RL training		
	\State Initialize network parameters $\theta$, $\phi$, $\psi$
	
	\While{$t_{env} \leq t_{max}$}
	\State Interact with the environment via $\epsilon$-greedy policy based on $[Q_{\theta}^i]_{i=1}^{n}$ and get a trajectory $\mathcal{T}$. 
	\State Run Algorithm \ref{alg:memory_construction} to update $\gD_{E}$ with $\mathcal{T}$
	\State Append $\mathcal{T}$ to  $\gD$
	
	\For{$k = 1$ \textbf{to} $n_\textrm{circle}$}
	\State Get $M$ sample trajectories $[\mathcal{T}]_{i=1}^{M} \sim \gD$ 
	\State Run MARL training algorithm using $[\mathcal{T}]_{i=1}^{M}$ and $\gD_{E}$, to update $\theta$ with Eq.\ref{Eq:action_policy_loss}
	\EndFor		
	
	\If{$t_{env}$ mod $t_{emb}$ == 0}
	\State Run Algorithm \ref{alg:embedder_training} to update $\phi$, $\psi$
	\State Update all $x \in \gD_{E}$ with updated $f_{\phi}$
	
	\EndIf
	\EndWhile	
\end{algorithmic}
\end{center}
\caption{EMU: Efficient episodic Memory Utilization for MARL}
\label{alg:EMU}
\end{algorithm}

Here, any CTDE training algorithm can be adopted for MARL training algorithm in \texttt{line 12} in Algorithm \ref{alg:EMU}. As we mentioned in Section \ref{Infrastructure}, training of $f_{\phi}$ and $f_{\psi}$ and updating all $x \in \gD_{E}$ only takes less than two seconds at most under the task with largest state dimension such as \texttt{corridor}. Thus, the computation burden for trainable embedder is negligible compared to the original MARL training.

\section{Memory Utilization}\label{memory_utilization}
A remaining issue in utilizing episodic memory is how to determine a proper threshold value $\delta$ in Eq. \ref{Eq:Episodic_update}. Note that this $\delta$ is used for both updating the memory and recalling the memory. One simple option is determining $\delta$ based on prior knowledge or experience, such as hyperparameter tuning. Instead, in this section, we present a more memory-efficient way for $\delta$ selection.
When computing $||{\hat x} - {x}||_2 < \delta$, the similarity is compared elementwisely. However, this similarity measure puts a different weight on each dimension of $x$ since each dimension of $x$ could have a different range of distribution. Thus, instead of $x$, we utilize the normalized value. Let us define a normalized embedding $y$ with the statistical mean ($\mu_x$) and variance ($\sigma_x$) of $x$ as 
\begin{equation}\label{Eq:normalization}
y=(x-\mu_x)/\sigma_x.
\end{equation}  
Here, the normalization is conducted for each dimension of $x$. Then, the similarity measure via $||{\hat y} - {y}||_2 < \delta$ with Eq. \ref{Eq:normalization} puts an equal weight to each dimension, as $y$ has a similar range of distribution in each dimension. In addition, an affine projection of Eq. \ref{Eq:normalization} maintains the closeness of original $x$-distribution, and thus we can safely utilize $y$-distribution instead of $x$-distribution to measure the similarity. 

In addition, $y$ defined in Eq. \ref{Eq:normalization} nearly follows the normal distribution, although it does not strictly follow it. This is due to the fact that the memorized samples $x$ in $\gD_E$ do not originate from the same distribution, nor are they uncorrelated, as they can stem from the same episode. However, we can achieve an approximate coverage of the majority of the distribution, specifically $3\sigma_y$ in both positive and negative directions of $y$, by setting $\delta$ as 
\begin{equation}\label{Eq:delta_determination}
\delta \leq \frac{(2 \times 3\sigma_y)^{\dim(y)}}{M}.
\end{equation}  
For example, when $M=1e^6$ and $\textrm{dim}(y)=4$, if $\sigma_y \approx 1$ then $\delta \leq 0.0013$. This is the reason we select $\delta=0.0013$ for the exploratory memory recall.

\end{document}